\documentclass{article}

\PassOptionsToPackage{table}{xcolor}
\usepackage{iclr2027_conference,times}
\iclrfinalcopy

\usepackage[utf8]{inputenc}
\usepackage[T1]{fontenc}
\usepackage{hyperref}
\usepackage{url}
\usepackage{booktabs}
\usepackage{amsfonts}
\usepackage{amsmath}
\usepackage{amssymb}
\usepackage{nicefrac}
\usepackage{microtype}
\usepackage{xcolor}
\usepackage{graphicx}
\usepackage{adjustbox}
\usepackage{subcaption}
\usepackage{wrapfig}
\usepackage{morefloats}
\extrafloats{200}
\usepackage{placeins}
\usepackage{algorithm}
\usepackage{algpseudocode}
\usepackage{multirow}
\usepackage{xspace}
\usepackage{enumitem}
\usepackage{amsthm}
\usepackage[capitalize,noabbrev]{cleveref}
\usepackage{appendix}
\usepackage{etoc}
\hypersetup{
  colorlinks=true,
  linkcolor={[rgb]{0.20,0.30,0.65}},
  citecolor={[rgb]{0.20,0.55,0.30}},
  urlcolor={[rgb]{0.20,0.30,0.65}}
}

\crefname{proposition}{Proposition}{Propositions}
\Crefname{proposition}{Proposition}{Propositions}
\crefname{corollary}{Corollary}{Corollaries}
\Crefname{corollary}{Corollary}{Corollaries}
\crefname{observation}{Observation}{Observations}
\Crefname{observation}{Observation}{Observations}
\crefname{remark}{Remark}{Remarks}
\Crefname{remark}{Remark}{Remarks}
\crefname{definition}{Definition}{Definitions}
\Crefname{definition}{Definition}{Definitions}
\crefname{algorithm}{Algorithm}{Algorithms}
\Crefname{algorithm}{Algorithm}{Algorithms}

\newtheorem{theorem}{Theorem}

\newtheorem{proposition}{Proposition}

\crefname{assumption}{Assumption}{Assumptions}
\Crefname{assumption}{Assumption}{Assumptions}
\theoremstyle{remark}

\newcommand{\ptbench}{\textbf{PRIMEBench}\xspace}

\newcommand{\ie}{\textit{i.e.}}

\newcommand{\rev}[1]{#1}
\newenvironment{revblock}{}{}

\title{Hierarchical Compression of\\Vision-Language Model Benchmarks}

\author{%
  Hyunjong Ok$^{1}$\thanks{Work done during an internship at Upstage.}
  \quad Seunggu Kang$^{2}$\thanks{Corresponding authors.}
  \quad Jaeho Lee$^{1}$\footnotemark[2] \\
  $^{1}$Pohang University of Science and Technology \quad $^{2}$Upstage AI \\
  \texttt{\{hyunjong.ok, jaeho.lee\}@postech.ac.kr}, \texttt{seunggu@upstage.ai}
}

\begin{document}

\etocdepthtag.toc{mtchapter}

\maketitle
\lhead{Preprint. Under review.}

\begin{abstract}
Thorough evaluation of vision--language models (VLMs) has become prohibitively expensive, as benchmarks span an ever-broader spectrum of capabilities and new models arrive at a relentless pace.
Benchmark compression methods that preserve model rankings at a fraction of the cost are well studied for language models, but for VLMs the question remains under-explored.
We present \ptbench (\textbf{P}runing \textbf{R}edundant \textbf{I}tems for \textbf{M}ultimodal \textbf{E}valuation), a vision-aware hierarchical benchmark compression framework that substantially reduces evaluation cost while preserving model rankings.
This hierarchical framework operates in four stages: data cleaning to remove items answerable without the image and all-correct items, category representative selection to pick one benchmark per capability category, item pruning with Vision-Aware Variance (VAW), and category-count pruning.
VAW combines inter-model variance with a vision-dependence score computed from multimodal embeddings alone, while encouraging coverage of diverse items within each benchmark. On models held out from item selection, it has the highest mean fidelity at the released $5\%$ retention.
The hierarchical design lets practitioners stop at any stage to match their compute budget; the released suite removes over $97\%$ of items while preserving model rankings.
Beyond compression, our analyses show how VLM evaluation behaves as model panels grow and evolve, providing guidance for designing future benchmarks that are more efficient, robust to model turnover, and explicit about the limits of evaluation-side pruning.
\end{abstract}

\section{Introduction}
\label{sec:intro}


A single full-suite evaluation of a modern vision--language model (VLM) now consumes hundreds of GPU-hours. The benchmark landscape has expanded far beyond traditional visual question answering~\citep{goyal2017vqav2,hudson2019gqa} to include GUI and web understanding for vision agents~\citep{cheng2024seeclick,koh2024visualwebarena}, with new categories emerging rapidly alongside advances in agentic and embodied AI. Consequently, every new model release must be evaluated against an ever-growing suite of benchmarks, causing evaluation costs to compound.


In this work, we study the problem of \textit{compressing} VLM benchmarks for efficient evaluation. Such compression is possible because VLM benchmarks exhibit redundancy at two levels. First, \textit{across benchmarks}, the model-benchmark score matrix is markedly low-rank, suggesting that many benchmarks measure overlapping capabilities. Second, \textit{within benchmarks}, many items contribute little discriminative signal, because they are unimportant or redundant with other items. Although several prior works have explored benchmark compression using these redundancies, they focus exclusively on within-benchmark redundancy \citep{polo2024tinybenchmarks,kipnis2025metabench,essencebench2026}. Moreover, only a handful of studies consider compression for VLM benchmarks, and none validates benchmark-level compression at large scale~\citep{uebayashi2026m3irt,joshi2026datbench}.

To this end, we propose \ptbench, a hierarchical benchmark compression framework for VLMs, supported by large-scale evaluations. Our method exploits across- and within-benchmark redundancies. Across benchmarks, benchmarks within a capability category induce highly correlated model rankings~\citep{liang2023helm,polo2024tinybenchmarks}; we replace each category with a single representative benchmark and exhaustively select a compact subset that preserves the full-suite ordering.
Within benchmarks, items solved by all models or failed by all carry no separation signal, while items with high inter-model variance dominate ranking quality; the all-wrong tail has a diagnostic role for future VLM evaluation.
We exploit these redundancies with one VLM-specific consideration: criteria for text-only large language models (LLMs) cannot tell apart items whose answer requires the image from items whose answer is text-driven, so we treat per-item vision-dependence as a design principle.

\begin{figure*}[t]
  \centering
  \includegraphics[width=0.83\textwidth]{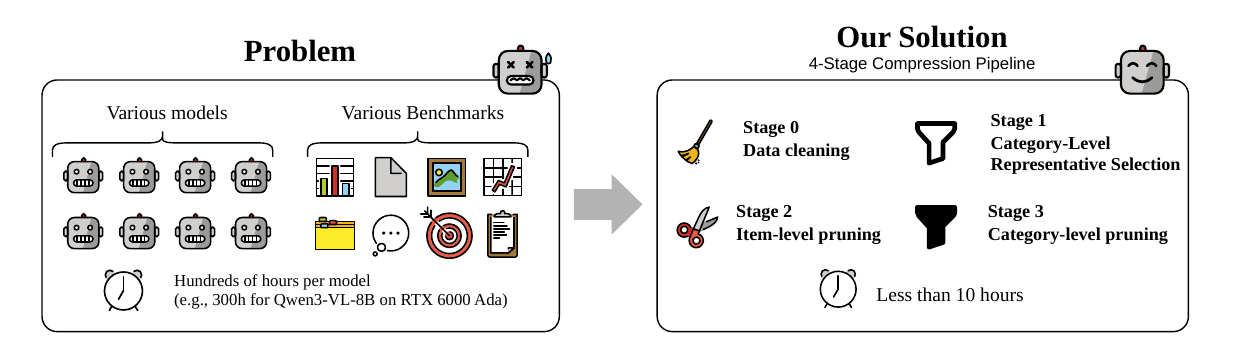}
  \caption{\textbf{Per-model VLM evaluation cost grows to hundreds of GPU-hours; \ptbench compresses the same suite to a compact subset that preserves model rankings.}
  \emph{Left}: a full sweep over the benchmark suite consumes hundreds of GPU-hours per model and compounds with quarterly model releases.
  \emph{Right}: the rankings induced by the compact \ptbench suite track the full-suite rankings closely at a small fraction of the original item budget.}
  \label{fig:teaser}
\end{figure*}


Our four-stage pipeline processes the score matrix end to end.
\emph{Data Cleaning} (Stage~0) removes items solvable without visual input and items answered correctly by every model.
\emph{Category Representative Selection} (Stage~1) picks one representative benchmark per category by correlation against the category mean, the average standardised score over the category's benchmarks.
\emph{Item Pruning with Vision-Aware Variance} (Stage~2) first divides the category budget, the number of items the category keeps, among its benchmarks in proportion to their sizes, which gives each benchmark its share. Within each benchmark it groups the items into \emph{coverage cells} (clusters of items with similar vision-text embeddings, one per item to keep) and keeps the highest-scoring item of each cell. The score is the inter-model variance plus a label-free vision-dependence prior, and the prior's effective weight is annealed with the number of evaluated models so that it fades as variance estimates become reliable.
\emph{Category-Count Pruning} (Stage~3) selects a compact subset of category pools, yielding the released suite.
We validate the pipeline on a panel of $30$ recent VLMs and its benchmark selection externally on a large public leaderboard, and complement it with diagnostic analyses that go beyond ranking and aim to inspire the design of next-generation VLM benchmarks.

Our contributions are threefold.
\begin{enumerate}[leftmargin=*,itemsep=2pt,topsep=2pt]
    \item \textbf{Open four-stage VLM compression pipeline.}
    We release PRIMEBench, to our knowledge the first VLM compression suite that compresses both benchmarks and items, with an end-to-end evaluation of ranking preservation, including on models held out from selection, and a benchmark-level check on an external leaderboard.
    \item \textbf{Vision-Aware Variance (VAW), a variance-based item criterion with a label-free vision-dependence prior and coverage cells.}
    VAW scores each item by its inter-model variance plus a vision-dependence prior computed from embeddings alone, whose effective weight shrinks as the number of models $M$ grows, and keeps one item per cell within each benchmark's share.
    \item \textbf{Diagnostic analyses that guide future VLM evaluation.}
    Beyond compression, our analyses explain where the gain of VAW comes from, test whether a compressed selection stays valid as new models arrive, and mark where compressed evaluation helps and where it stops, pointing to how evaluation suites can stay efficient as the model landscape changes.
\end{enumerate}


\section{Related work}
\label{sec:related}

\paragraph{VLM benchmarks and evaluation.}
VLMs are now evaluated across a broad capability spectrum: general visual question answering~\citep{goyal2017vqav2,hudson2019gqa}, document and chart understanding~\citep{mathew2021docvqa,masry2022chartqa}, and multi-discipline reasoning~\citep{yue2024mmmu,liu2024mmbench,fu2025mme}.
Evaluating a single model now routinely spans 20+ benchmarks, with public leaderboards such as the OpenVLM leaderboard~\citep{duan2024vlmevalkit} aggregating community results at increasing cost.
Discriminative signal diminishes as models improve, since competitive models increasingly solve the same items~\citep{kiela2021dynabench}.

\paragraph{Benchmark compression.}
A growing body of work reduces evaluation cost by selecting item subsets, but almost exclusively for text-only LLMs.
Item-level criteria fall into families: item response theory (IRT) anchor selection (tinyBenchmarks~\citep{polo2024tinybenchmarks}, metabench~\citep{kipnis2025metabench}), inter-model-disagreement ranking (DISCO~\citep{disco2026}), genetic-algorithm search guided by sample attribution (EssenceBench~\citep{essencebench2026}), sparse-anchor optimisation (SparseEval~\citep{sparseeval2026}), rank-correlation prediction (SubLIME~\citep{saranathan2025sublime}), and embedding coverage (LMMs-Eval Lite~\citep{zhang2025lmmseval}).
Like tinyBenchmarks and DISCO, which evaluate a subset on models outside the set it was fitted on, we score every VLM selector on models unseen during selection.
For multimodal evaluation, prior work has proposed multimodal IRT that decomposes item difficulty into image-only, text-only, and cross-modal factors and uses it to select subsets (M3IRT~\citep{uebayashi2026m3irt}), and applied point-biserial item discrimination to VLM benchmarks (DatBench~\citep{joshi2026datbench}); none releases a compressed suite that jointly compresses the benchmark axis (which benchmarks to keep) and the item axis (which items within each benchmark to retain), and DatBench's correctness-only criterion remains vision-blind, ignoring whether an item's answer requires the visual input.
\ptbench unifies the two axes for multimodal evaluation and validates benchmark-level ranking preservation on a large external panel.

\begin{table}[!t]
\caption{\textbf{Selected benchmark compression methods, organised by modality, compression level, and validation panel size.}
  \emph{Mod.}\ denotes the demonstration setting in the original work (T=text-only LLMs, V=vision--language). \emph{Level}: B=benchmark, I=item. \emph{Scale}: largest validation panel.
  \ptbench combines benchmark- and item-level compression, releases a compact suite, and validates item selection on $30$ VLMs and benchmark-level ranking preservation against $231$ VLMs.}
  \label{tab:related_methods}
  \centering
  \footnotesize
  \setlength{\tabcolsep}{25pt}
  \resizebox{\linewidth}{!}{%
\begin{tabular}{l c c l}
    \toprule
    Method & Mod. & Level & Scale \\
    \midrule
    tinyBenchmarks~\citep{polo2024tinybenchmarks} & T & I   & ${\sim}$300 LLMs \\
    metabench~\citep{kipnis2025metabench}          & T & I   & ${\sim}$5000 LLMs \\
    DISCO~\citep{disco2026}                        & T & I   & --- \\
    EssenceBench~\citep{essencebench2026}          & T & I   & --- \\
    SparseEval~\citep{sparseeval2026}              & T & I   & --- \\
    SubLIME~\citep{saranathan2025sublime}          & T & I   & --- \\
    \midrule
    M3IRT~\citep{uebayashi2026m3irt}               & V & I    & $24$ VLMs \\
    DatBench~\citep{joshi2026datbench}             & V & I    & $27$ VLMs \\
    \midrule
    \textbf{\ptbench (ours)} & \textbf{V} & \textbf{B$+$I} & \textbf{30\,/\,231 VLMs} \\
    \bottomrule
  \end{tabular}%
  }
\end{table}

\section{Problem statement}\label{sec:problem}
We formalise the task of benchmark compression as follows. Let $\mathcal{M}$ be a set of $M$ models, and let $\mathcal{B}$ be a catalogue of $B$ benchmarks. Each benchmark $b \in \mathcal{B}$ is a set of $N_b$ items.

The \textit{item score} $x_{m,b,i}$ denotes the prediction quality of the model $m \in \mathcal{M}$ on the $i$-th item of the benchmark $b \in \mathcal{B}$. For closed-option problems, we have $x_{m,b,i} \in \{0,1\}$, indicating whether the model has been correct on the item or not. The item score can be continuous, e.g., when using continuous metrics or LLM judge outputs. The \textit{score} $s_{m,b} = \frac{1}{N_b}\sum_{i=1}^{N_b} x_{m,b,i}$ is the average item score of the model $m$ on benchmark $b$, where $N_b$ is the number of items in $b$. The \textit{mean score} $\bar{s}_m = \frac{1}{B}\sum_b s_{m,b}$ is the average score of a model $m$ over all $B$ benchmarks.

The \textit{full ranking} $\boldsymbol{\pi}$ is an ordering of $M$ models according to the mean score, i.e.,
\begin{align}
\boldsymbol{\pi} = (m_{(1)},m_{(2)},\ldots,m_{(M)}),\qquad \mathrm{such\:that}\quad \bar{s}_{m_{(1)}} \ge \bar{s}_{m_{(2)}} \ge \cdots \ge \bar{s}_{m_{(M)}}
\end{align}
We consider compressing this catalogue along two axes: across-benchmark, and within-benchmark. In other words, our goal is to select the \textit{benchmark subset}
$\mathcal{S} \subseteq \mathcal{B}$ and the \textit{item subsets} $\mathcal{I}'_b \subseteq \mathcal{I}_b := \{1,\ldots,N_b\}$ for each $b \in \mathcal{S}$ in a way that the ranking evaluated on this set is close to the full ranking $\boldsymbol{\pi}$. Precisely, let $\bar{s}'_{m}$ be the average score of the model $m$ over the compressed suite, and $\boldsymbol{\pi}'$ be the corresponding \textit{compressed ranking}. Then, our goal is to solve:
\begin{align}
\max_{\mathcal{S},\{\mathcal{I}'_{b}\}}\rho(\boldsymbol{\pi},\boldsymbol{\pi}'),\qquad \mathrm{subject\:to} \quad |\mathcal{S}| \le k,\:|\mathcal{I}'_b| \le r N_b,
\end{align}
where $k$ is the number of benchmarks (category pools in Stage~3) kept, $r$ is the fraction of each benchmark's items kept (the \emph{retention}; Stage~2 adds a $100$-item floor per category pool), and $\rho(\cdot,\cdot)$ is the Spearman rank correlation.

\section{Method}
\label{sec:method}

Compression of a multi-benchmark VLM evaluation suite has two natural axes: \emph{which benchmarks to keep} and \emph{which items within them to retain}. Before either decision can be made, the item pool itself needs to be cleaned of items that no longer separate models or that admit a non-visual shortcut.
Our method therefore proceeds in four stages---data cleaning (\cref{sec:stage0}), category representative selection (\cref{sec:stage1}), item pruning with VAW (\cref{sec:stage2}), and category-count pruning (\cref{sec:stage3})---each motivated by a distinct property of VLM evaluation.


\subsection{Stage 0: Data cleaning}
\label{sec:stage0}

Stage~0 filters out benchmark items that do not carry useful evaluation signals. For example, the items that all models in $\mathcal{M}$ can answer correctly may carry no useful information for ranking. Also, the items that can be answered correctly from the text alone may fall outside the scope of VLM evaluation---their solvability reflects language priors that are measured by LLM-only benchmarks already, not vision--language capability. This stage removes both populations to obtain a clean pool that measures VLM-specific capability. We use this pool as the reference for end-to-end evaluation.


Concretely, for benchmarks with binary correctness $x_{m,b,i} \in \{0,1\}$, Stage~0 applies two filters, the second where enough image-free scores exist.
\begin{itemize}[leftmargin=*]
\item The \emph{all-correct} filter removes items that all evaluated models in $\mathcal{M}$ solve correctly, i.e., $x_{m,b,i} = 1$ for all $m \in \mathcal{M}$. For items that all models predict incorrectly, we keep those items, since later generations of models may solve them.
\item The \emph{blind-solvability} filter computes $\beta_i$---the fraction of the models run without images that answer item $i$ correctly from the text prompt alone---and removes items with $\beta_i \geq \tau$ for a fixed $\tau{=}2/3$, a text-only filtering strategy motivated by MMStar \citep{chen2024mmstar}.
\end{itemize}

On benchmarks with continuous or judge scores both filters use the scores scaled to $[0,1]$, so an item is blind-solvable when the image-free models reach a mean score of at least $2/3$ (\cref{app:cleaning}).


\subsection{Stage 1: \texorpdfstring{Category representative selection}{Category representative selection}}
\label{sec:stage1}

Stage~1 reduces the number of benchmarks by removing the benchmarks that measure overlapping VLM capabilities. We organise the benchmarks into 18 capability categories---constructed by aggregating the evaluation taxonomies adopted by recent frontier VLM technical reports~\citep{qwen3vl2025,internvl3_5_2025,glm4_1v_2025}---and select one representative benchmark for each category. We provide additional details on the category construction in \cref{app:category_rep}.


More formally, define the score vector $\mathbf{s}_b = (s_{1,b},\ldots,s_{M,b})$ for each benchmark $b$. For each category $C$, we define its \textit{category mean} as an average of the $z$-score-normalised score vectors of the benchmarks belonging to this category, i.e., $\mathbf{s}_C = \tfrac{1}{|C|}\sum_{b \in C} z(\mathbf{s}_b)$, where $z(\cdot)$ denotes the $z$-score normalisation. As the category representative, we select the benchmark whose score vector is best correlated with the category mean:
\begin{equation}
b^*_C = \arg\max_{b \in C}\,\rho(\mathbf{s}_b, \mathbf{s}_C).
\label{eq:rep}
\end{equation}

For categories whose representative correlates weakly with the category mean, we supplement it with items from other benchmarks in the category, using a variance-ranked prefix of those candidate items, to preserve the category-level ranking; details are in \cref{app:multisource}.


\subsection{Stage 2: \texorpdfstring{Item pruning with Vision-Aware Variance}{Item pruning with Vision-Aware Variance}}
\label{sec:stage2}
Stage~2 reduces the number of items in category pools, by keeping only the items that play a critical role in ranking the vision--language capabilities of VLMs. Our decision criterion is similar to that in Stage~0. In particular, we ask two questions: (1) Does this item provide meaningful discriminative signal? (2) Does this item use visual input in a meaningful way?

Precisely, we define the following \textbf{VAW} (\underline{V}ision-\underline{AW}are Variance) score
\begin{align}
q_i^{\mathrm{VAW}} = \widehat{\sigma}_i^2 + \alpha \cdot \delta_i^{\mathrm{VL}},
\label{eq:vaw}
\end{align}
which is a mixture of two scores: $\widehat{\sigma}_i^2$ is the \textit{variance score}, measuring the strength of the discriminative signal provided by the item $i$; $\delta_i^{\mathrm{VL}}$ is the \textit{vision-dependence score}, a proxy score for measuring how much the item relies on visual input; the weight $\alpha\ge 0$ sets their balance.

\textbf{Variance score.} The variance score is a $\sqrt{M}$-scaled variance of item scores across models, i.e.,
\begin{equation}
    \widehat{\sigma}^2_{i} := \frac{1}{\sqrt{M}} \sum_{m=1}^{M} (x_{m,i} - \bar{x}_i)^2,
    \label{eq:vardisc}
\end{equation}
where $\bar{x}_i = \tfrac{1}{M}\sum_{m}x_{m,i}$ is the average item score of the models on item $i$, and item scores are scaled from the range of their metric to $[0,1]$ so that binary and continuous scores share one scale.\footnote{We drop the benchmark index $b$ of $x_{m,b,i}$ to lighten notation.} Whenever the item score is binary, the variance score admits a simpler form $\widehat{\sigma}^2_i = \sqrt{M}\,\bar{x}_i(1-\bar{x}_i)$.\footnote{With binary $x_{m,i}\in\{0,1\}$, $\sum_m(x_{m,i}-\bar{x}_i)^2 = M\,\bar{x}_i(1-\bar{x}_i)$, so dividing by $\sqrt{M}$ gives $\sqrt{M}\,\bar{x}_i(1-\bar{x}_i)$; equivalently $\widehat\sigma^2_i=\sqrt{M}\,\widehat v_i$ where $\widehat v_i=\bar{x}_i(1-\bar{x}_i)$ is the standard plug-in variance.} LLM benchmark compression methods also use similar variance scores \citep{disco2026}.

We use $1/\sqrt{M}$ as the scaling factor instead of the standard $1/M$. This makes the effective weight of the vision-dependence score decrease with the number of evaluated models. Dividing \eqref{eq:vaw} by $\sqrt{M}$ gives $\widehat v_i + (\alpha/\sqrt{M})\,\delta_i^{\mathrm{VL}}$, where $\widehat v_i=\widehat\sigma^2_i/\sqrt{M}$ is the plug-in variance of the item's scores across models and $\alpha_M=\alpha/\sqrt{M}$ is the \emph{tilt}. Under i.i.d.\ model responses, the estimation error of $\widehat v_i$ concentrates at the generic $O(M^{-1/2})$ rate (\cref{app:scaling_law}), while the tilt shrinks at the same rate. Thus, the vision-dependence term has greater relative weight for smaller model panels and fades as the variance estimate becomes more stable. For a common $M$, this rescaling does not change the item ranking; when the number of valid model scores differs across items, we rank using the corresponding divided score. We fix $\alpha = 0.187$ for all experiments, without tuning it on the held-out splits; $\alpha = 0$ gives the variance score alone.


\textbf{Vision-dependence score.} The vision-dependence score measures how adding visual inputs can steer the direction of the text-only embedding. Precisely, the score is defined as the cosine distance
\begin{align}
\delta_i^{\mathrm{VL}} = 1 - \cos(E_i^{\mathrm{L}}, E_i^{\mathrm{VL}}),
\end{align}
where $E_i^{\mathrm{L}}$ denotes the text-only embedding of the item prompt, and $E_i^{\mathrm{VL}}$ denotes the vision-text embedding, both extracted by one VLM encoder, Qwen3-VL-Embedding-8B~\citep{qwen3vl_embedding_2026}.

The vision-dependence score has two advantages. First, the score is easy to compute, as we do not need to generate answers in multiple decoding steps. Second, the score is label-free: it does not use the models' item scores $x_{m,b,i}$.

\textbf{Selection with coverage cells.} At retention $r$, a Stage~1 \emph{category pool} of $n_C$ items keeps $K_C=\max(\lfloor r n_C\rfloor,\min(n_C,100))$ items, which we call the \emph{category budget}. We divide this budget among benchmarks in proportion to their sizes, so benchmark $b$ receives $K_b$ items as its \emph{benchmark share}. Within each benchmark, we group the items into $K_b$ \emph{coverage cells} by $k$-means on the vision-text embeddings $E_i^{\mathrm{VL}}$ and keep the highest-scoring item from each cell. This fills the benchmark share while discouraging near-duplicate items from consuming the budget. Ties are broken in a seeded random order, and the embeddings are reduced to $64$ dimensions by a fixed random projection before $k$-means. As $M$ grows, VAW recovers the variance-only choice within each cell when the competing items are sufficiently separated in variance (\cref{app:scaling_law}).

\subsection{Stage 3: \texorpdfstring{Category-count pruning}{Category-count pruning}}
\label{sec:stage3}

\begin{figure}[!t]
  \centering
  \includegraphics[width=0.66\textwidth]{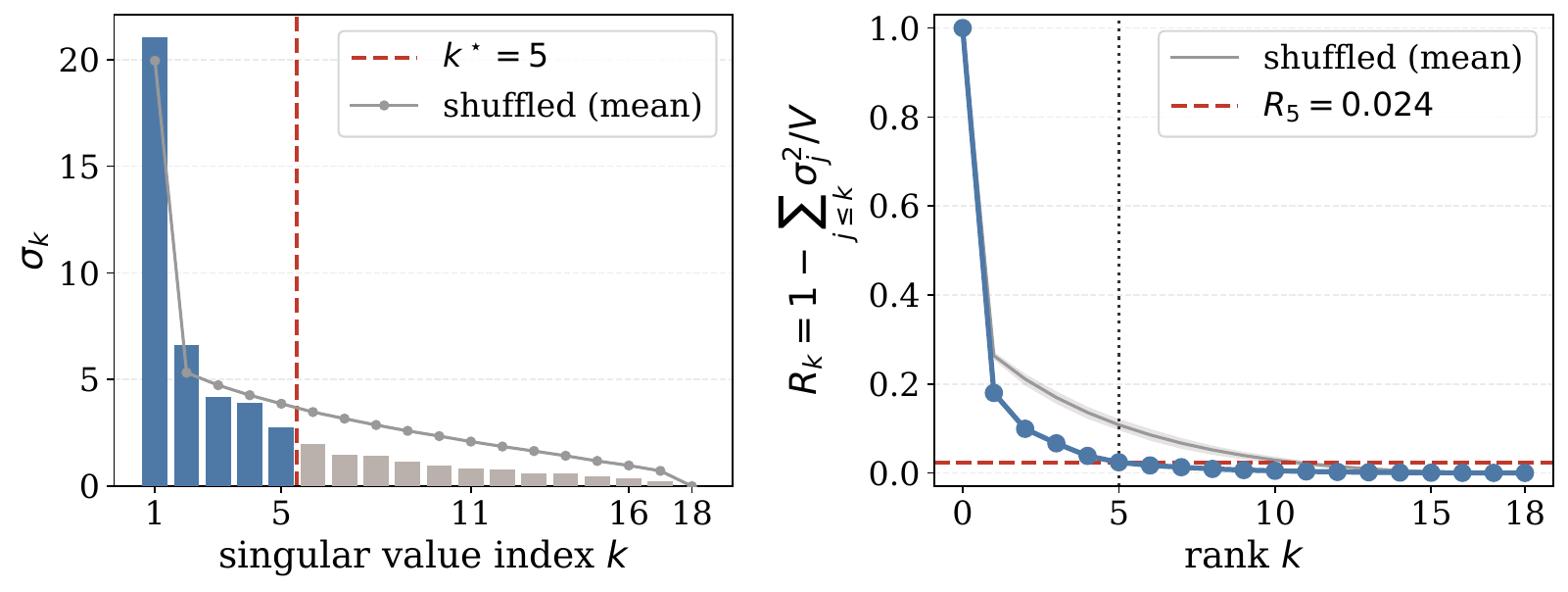}
  \caption{\textbf{The VLM score matrix is markedly low-rank.}
  Singular-value spectrum (\emph{left}) and cumulative residual energy (\emph{right}) of the per-model $z$-normalised category score matrix. The top-$5$ singular values capture $97.6\%$ of the spectral energy, matching the released number of categories $k^\star{=}5$.}
  \label{fig:low_rank_spectrum}
\end{figure}

Stage~3 further compresses the suite by pruning out redundant categories. The key motivation comes from the observation that the model$\times$category score matrix is markedly low-rank. On the $18$-category score matrix, the cumulative spectral energy reaches 95\% within the top-$5$ singular values, where energy is the sum of squared singular values (\cref{fig:low_rank_spectrum}).



We therefore select a subset of $k$ categories from $\tilde{\mathcal{B}}$---the set of category pools constructed by Stage~1.
Let $\tilde{\boldsymbol{\pi}}$ denote the model ranking induced by the mean score over all Stage~2-pruned category pools in $\tilde{\mathcal{B}}$, and let $\tilde{\mathbf{s}}_C$ denote the per-model score vector on category pool $C$ \emph{after Stage~2 item pruning at $r{=}5\%$} (\ie, Stage~3 is fitted on the Stage~2 output, making the four-stage pipeline strictly sequential).
Because the search space $\binom{|\tilde{\mathcal{B}}|}{k}$ is small for the numbers of categories of interest, we exhaustively enumerate every $k$-subset and pick the one that maximises Spearman correlation with $\tilde{\boldsymbol{\pi}}$:
\begin{equation}
\mathcal{S}^\star_k = \arg\max_{\mathcal{S}\subseteq \tilde{\mathcal{B}},\,|\mathcal{S}|=k}\,\rho\!\left(\tilde{\boldsymbol{\pi}},\;\operatorname{rank}\!\left(\frac{1}{|\mathcal{S}|}\sum_{C \in \mathcal{S}} \tilde{\mathbf{s}}_{C}\right)\right).
\label{eq:stage3}
\end{equation}
The $k$-sweep is in \cref{app:ksweep}.

Concretely, the released suite reports only the $k$ selected category pools, each pruned by Stage~2, yielding $\sum_{C\in\mathcal{S}^\star_k} K_C$ items in total.

\section{Experimental setup}
\label{sec:setup}

\textbf{Models.}
We evaluate the proposed compression pipeline on $30$ vision--language and omni models from multiple families, including Qwen, InternVL, Gemma, LLaVA-OneVision, MiniCPM-V, and Kimi-VL; per-model details are in \cref{app:models}. Vision-text embeddings from Qwen3-VL-Embedding-8B~\citep{qwen3vl_embedding_2026}, which is not among the evaluated models, are used for Stage~2; \cref{app:encoder} compares alternative embedding encoders.

\textbf{Benchmarks.}
The catalogue contains $43$ benchmarks organised into $18$ capability categories spanning visual question answering, reasoning, document understanding, and GUI/web understanding; scoring types and item counts are given in \cref{app:benchmarks}.

\textbf{External validation.}
We additionally evaluate on the OpenVLM leaderboard~\citep{duan2024vlmevalkit}, which contains up to $265$ VLMs; details are in \cref{app:external_validation}.

\textbf{Baselines.}
Method comparisons focus on Stage~2, where the item selectors differ. We compare VAW with DISCO~\citep{disco2026}, DatBench~$r_{pb}$~\citep{joshi2026datbench}, IRT~2PL~\citep{kipnis2025metabench}, AnchorPoints~\citep{vivek2024anchor}, and Variance at retentions $r \in \{1,2,5,10,20,50\}\%$. Variance selects the category-wide top-$K$ items ($K{=}K_C$) by \cref{eq:vardisc}, without benchmark shares or coverage cells. DISCO selects items from prediction disagreement on the logged model answers; for a consistent comparison, its selected items are scored by their mean rather than by DISCO's learned performance predictor.

\textbf{Metrics and evaluation.}\label{sec:unseen}
Our primary metric is the Spearman rank correlation~\citep{spearman1904} between compressed- and full-pool model rankings. Each split partitions the $30$ models into build and held-out sets, with Stage~2 selection performed only on the build models. For the Stage~2 comparison, all selectors operate on common category pools, built once on the full panel, and fidelity is measured on held-out models within each category. For end-to-end evaluation, Stages~1--3 are re-fitted on the build models and evaluated against the ranking from all Stage~0-cleaned benchmarks. Category pools retain at least $100$ items, with smaller pools kept in full; the same floor applies to every selector. We report results over $20$ balanced random $15/15$ splits and a chronological split using the $15$ oldest models for selection. Stage~0 cleaning and the released suite both use the full $30$-model panel. Additional protocol details are provided in \cref{app:unseen_protocol}.

\section{Results}
\label{sec:results}

\paragraph{End-to-end compression.}
\label{sec:res_endtoend}

\begin{table}[!t]
\centering
\caption{\textbf{Stage~0--3 pipeline at a glance: the released suite preserves model rankings with a fraction of the original items.}
Change is relative to each row's input, except for the released suite, which is relative to all $43$ benchmarks. $\rho$ is the median over categories for Stage~1, the mean over categories on unseen models for Stage~2, the correlation of the $5$ selected categories against all $18$ pruned categories for Stage~3, and the correlation with the ranking over all Stage~0-cleaned benchmarks for the released suite.}
\label{tab:stage_audit_summary}
\small
\setlength{\tabcolsep}{5pt}
\begin{adjustbox}{max width=\linewidth}
\begin{tabular}{l l r r r c}
\toprule
Stage & Unit & Input & Output & Change & Ranking fidelity $\rho$ \\
\midrule
Stage~0 (cleaning) & items & 386{,}324 & 270{,}408 & $-30.0\%$ & --- \\
Stage~1 (category rep) & benchmarks & 43 & 18 & $-58.1\%$ & 0.951 (median) \\
Stage~2 (item pruning, VAW) & category-pool items & 222{,}376 & 11{,}498 & $-94.8\%$ & 0.925 (mean) \\
Stage~3 ($k{=}5$) & categories & 18 & 5 & $-72.2\%$ & \textbf{0.988} \\
\textbf{Released suite} & items & 386{,}324 & 9{,}377 & $-97.57\%$ & \textbf{0.959} \\
\bottomrule
\end{tabular}
\end{adjustbox}
\end{table}

The full Stage~0$\to$3 pipeline reduces the $386{,}324$ items of the $43$ benchmarks to $9{,}377$ at $k{=}5$, and the released suite achieves $\rho{=}$$0.959$ on the full $30$-VLM panel against the unpruned Stage~0-cleaned reference (\cref{tab:stage_audit_summary}).

\paragraph{Stage~0 (data cleaning).} The \emph{blind-solvability} filter ($\tau{=}2/3$) removes items solvable without the image, and the complementary \emph{all-correct} filter removes items that every model solves. Stage~0 keeps all-wrong items to avoid prematurely discarding items that later models may solve; Stage~2 may subsequently prune them, since their inter-model variance is zero. Together the two filters retain 270{,}408 of the $386{,}324$ items of the $43$ benchmarks; per-filter breakdown is in \cref{app:cleaning}.

\paragraph{Stage~1 (category representative selection).} For each of the $18$ capability categories we pick the benchmark whose per-model score vector best correlates with the category mean (\cref{eq:rep}); after same-category item-level supplementation, the median per-category $\rho$ is $0.951$. The per-category list and supplementation cost are in \cref{app:stage1}.

\paragraph{Stage~2 (item pruning with Vision-Aware Variance).}
\begin{figure}[!t]
\centering
\captionsetup[subfigure]{justification=centering}

  \begin{subfigure}[c]{0.50\linewidth}
    \centering
    \includegraphics[width=\linewidth]{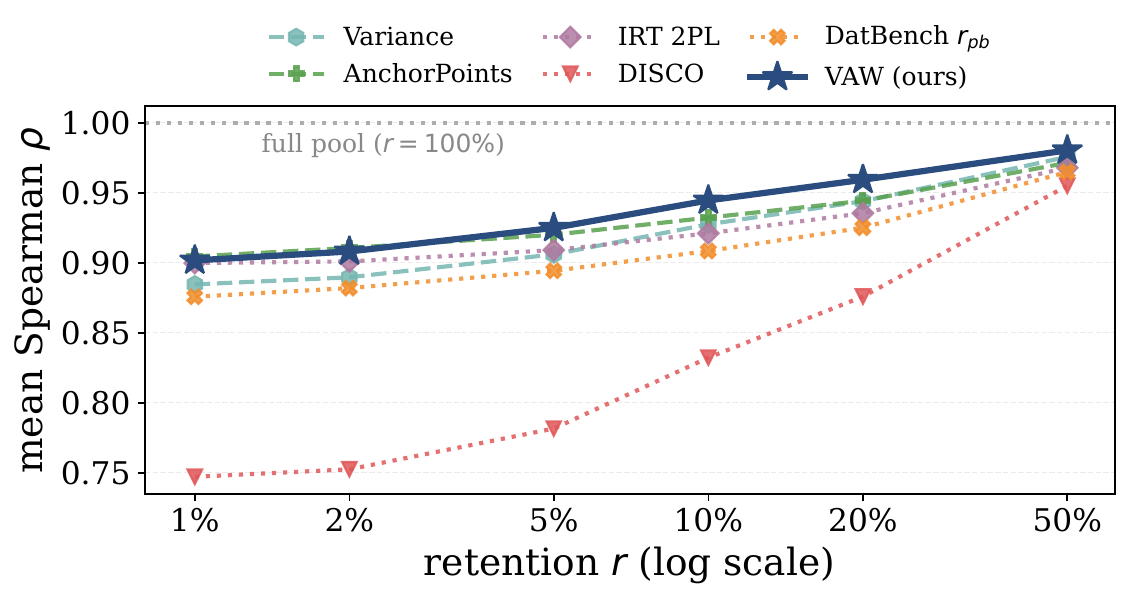}
    \subcaption{Spearman $\rho$ vs.\ retention $r$.}
    \label{fig:pt_method_scaling-curve}
  \end{subfigure}\hfill
  \begin{subfigure}[c]{0.48\linewidth}
    \centering
    \footnotesize
    \setlength{\tabcolsep}{3pt}
    \renewcommand{\arraystretch}{1.25}
    \begin{adjustbox}{max width=\linewidth}
    \begin{tabular}{l cccccc}
      \toprule
      Selector & $1\%$ & $2\%$ & $5\%$ & $10\%$ & $20\%$ & $50\%$ \\
      \midrule
      Variance & 0.884 & 0.890 & 0.906 & 0.927 & \underline{0.944} & \underline{0.976} \\
      AnchorPoints~\citep{vivek2024anchor} & \textbf{0.904} & \textbf{0.911} & \underline{0.920} & \underline{0.932} & \underline{0.944} & 0.971 \\
      IRT 2PL~\citep{kipnis2025metabench} & 0.900 & 0.901 & 0.909 & 0.921 & 0.935 & 0.968 \\
      DISCO~\citep{disco2026} & 0.747 & 0.752 & 0.781 & 0.832 & 0.876 & 0.955 \\
      DatBench~$r_{pb}$~\citep{joshi2026datbench} & 0.876 & 0.882 & 0.894 & 0.908 & 0.925 & 0.965 \\
      \textbf{VAW (ours)} & \underline{0.902} & \underline{0.908} & \textbf{0.925} & \textbf{0.944} & \textbf{0.959} & \textbf{0.980} \\
      \bottomrule
    \end{tabular}
    \end{adjustbox}
    \subcaption{Values at fixed $r$.\\
    (Bold = best per column; underline = second.)}
    \label{fig:pt_method_scaling-table}
  \end{subfigure}

\caption{\textbf{VAW leads at the released $5\%$ retention on models held out from Stage~2 selection.}
Stage~2 item pruning on the $18$ PRIMEBench category pools, with selection on half of the model panel and mean Spearman $\rho$ evaluated on the other half over $20$ random splits.}
\label{fig:pt_method_scaling}
\end{figure}
Within each category pool, Stage~2 keeps a small fraction $r$ of items.
On the 18 PRIMEBench category pools evaluated on held-out models (\cref{fig:pt_method_scaling}), VAW has the highest fidelity at $r=5$--$50\%$, while AnchorPoints leads at $r\le 2\%$. Ablations show that benchmark shares and coverage cells account for most of VAW's gain over plain Variance, and embedding-based cells outperform random partitions (\cref{app:ablation_cells}). Comparisons with selection and scoring on the same OpenVLM models are in \cref{app:openvlm_stage2}.

\paragraph{Stage~3 (category-count pruning).}
Stage~3 selects $k$ of the Stage~2-pruned category pools by exhaustively evaluating all $\binom{18}{k}$ subsets. Its selection objective compares the ranking induced by the selected categories with that induced by all $18$ pruned categories. At the released setting $k=5$, Stage~3 selects \{Academic Knowledge, Chart Understanding, Hallucination Detection, Visual Reasoning, Web/UI Understanding\} and reaches $\rho=0.988$ against this $18$-category pruned reference.

This differs from the end-to-end fidelity of the released suite ($\rho=0.959$ above), which is measured against the ranking from all Stage~0-cleaned benchmarks. We further evaluate generalisation under the unseen-model protocol: Stages~1--3 are fitted on $15$ build models and evaluated on the other $15$. Averaged over $20$ splits, VAW at $r{=}5\%$ followed by Stage~3 reaches $\rho=0.892$ against the same Stage~0 reference, within $0.01$ of Stage~3 without item pruning ($0.901$; \cref{app:end2end}).

\paragraph{External validation.}
On $231$ of $265$ OpenVLM models scored on at least half of the $14$ benchmarks overlapping with \ptbench, Stage~3 exhaustive selection at $k=5$ achieves $\rho=0.988$ against the $14$-benchmark ranking. Era-stratified spectral analysis shows that the top five components hold $86$--$89\%$ of the energy in every release-date cohort (\cref{tab:openvlm_eras}); details are in \cref{app:external_validation}.

\section{Discussion}
\label{sec:discussion}

The results show that \ptbench{} preserves model rankings at a fraction of the original evaluation cost. \rev{We ask why VAW works, whether the selection ages, what it misses, and where else it applies.}

\begin{revblock}
\paragraph{(1) Where the gain of VAW comes from.}\label{par:disc_four_constraints}
VAW is vision-aware in two places: its coverage cells are drawn on vision-text embeddings, and its score adds a vision-dependence tilt. At our panel size, most of the gain comes from the cells. In the held-out ablation (\cref{app:ablation_cells}), VAW with benchmark shares alone reaches $0.908$ at $r{=}5\%$, essentially matching Variance ($0.906$); adding vision-text coverage cells raises it to $0.925$ and outperforms random partitions of the same size. The tilt follows a theoretical scaling with panel size $M$: the plug-in variance concentrates at rate $M^{-1/2}$, so VAW anneals the tilt at the same rate, $\alpha/\sqrt{M}$, and its selection approaches the variance selector as $M$ grows. The OpenVLM scan shows that the best in-cell tilt tends to decrease with $M$ and is of the same order as the deployed $\alpha/\sqrt{M}$ on large panels. At $r{=}2\%$, the cross-seed-selected tilt improves $\rho$ over variance-only selection in the same cells at every tested size from $M{=}4$ to $192$; from $M{=}14$ onward, the gain is $0.005$--$0.017$ and exceeds two standard errors at every tested size. These results support smaller tilts on larger panels (\cref{app:scaling_law,app:finite_m_v2}).
\end{revblock}

\begin{revblock}
\paragraph{(2) Shelf life: does a frozen benchmark selection age as newer models arrive?}\label{par:disc_shelf_life}
We order the OpenVLM models with a known release date by release date, select the Stage~3 subset on the oldest $132$ models, freeze it, and evaluate it as the panel grows to $166$, $200$, and $203$ models. Its ranking fidelity stays at $\rho{\approx}0.99$, within $0.0004$ of the best subset refitted on each enlarged panel. Because the OpenVLM score matrix is low-rank, even random five-benchmark subsets preserve rankings reasonably well, so we also rank the frozen subset among all five-benchmark subsets; it stays at or above the $99.9$th percentile throughout the dated range. The benchmark-level selection therefore remains stable over the dated range studied (\cref{app:robustness}).
\end{revblock}

\begin{revblock}
\paragraph{(3) Capability Frontier: items that every current family finds hard.}\label{par:disc_frontier}
Stage~2 favours items that separate today's models, so items that most current models still fail tend to be pruned, even though such hard cases may become informative as capabilities improve. We therefore release a second set, the \emph{Capability Frontier}: among items that every current model family finds hard (each family's mean score at most half the maximum, panel mean at least $0.10$), we keep those with the largest difficulty-weighted variance (the plug-in variance times the mean shortfall $1-\bar x_i$); as a heuristic budget, the set size is the in-band pool size times the panel's largest per-generation accuracy gain. The resulting $1{,}653$ items still separate the oldest and newest Qwen generations (\cref{app:capability_frontier}). Whether they will separate future models is untested, so we treat the set as a diagnostic of the current panel rather than a forecast.
\end{revblock}

\begin{figure}[t]
\centering
  \begin{subfigure}{0.46\linewidth}
    \centering
    \includegraphics[width=\linewidth]{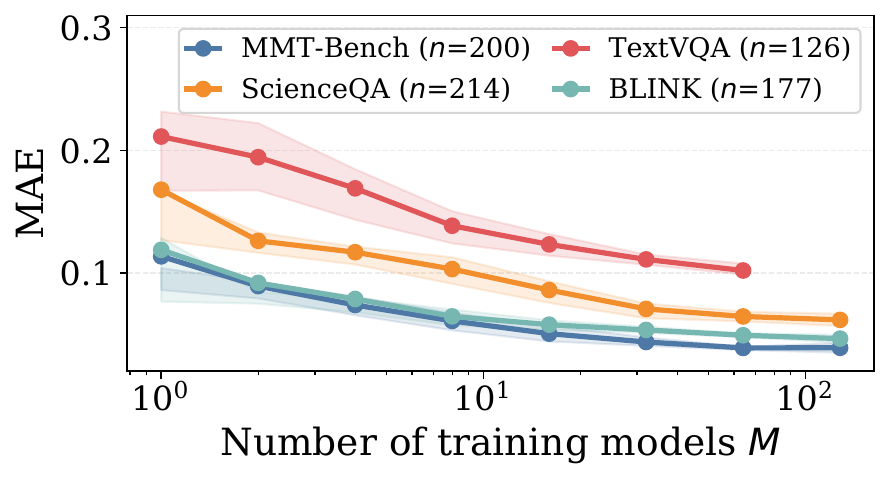}
    \subcaption{Left-out benchmark prediction.}\label{fig:d1_mae_scaling}
  \end{subfigure}\hfill
  \begin{subfigure}{0.46\linewidth}
    \centering
    \includegraphics[width=\linewidth]{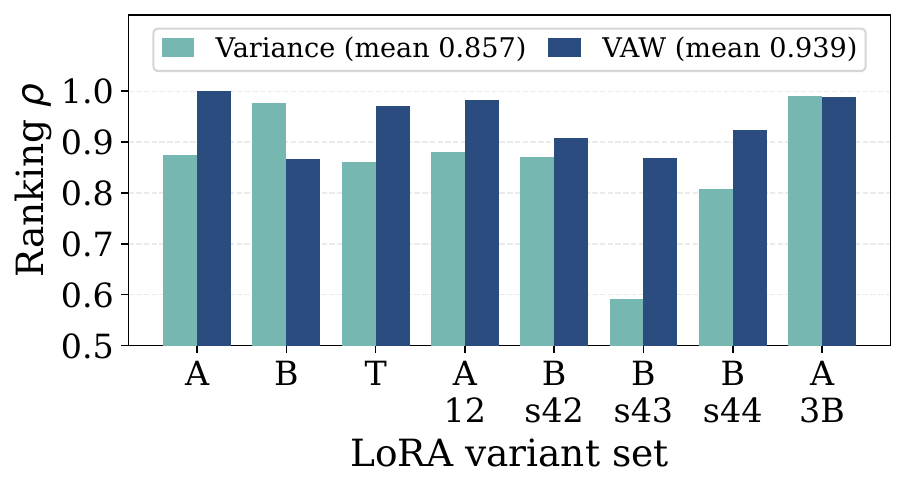}
    \subcaption{Ranking LoRA variants of one model.}\label{fig:ckpt_ranking_gain}
  \end{subfigure}
  \caption{\textbf{Benchmark scores predict a left-out benchmark better as models accumulate, and VAW ranks close variants of one model better than Variance.} (a) Mean absolute error (MAE) with the $p25$--$p75$ band over $30$ splits; $n$: models scored. (b) Variant-ranking $\rho$ at $r{=}5\%$ on the eight stable LoRA variant sets (A, B: learning-rate grids; T: one run; 12: twelve rates; s42--s44: seeds; 3B: Qwen2.5-VL-3B-Instruct).}
  \label{fig:disc_diagnostics}
\end{figure}

\paragraph{\rev{(4) Beyond ranking: predicting a left-out benchmark from the others.}}\label{par:disc_scaling_law}
Ranking fidelity is our primary objective, but benchmark scores may also support performance estimation. We predict a model's score on a left-out OpenVLM benchmark from its scores on the others, and vary the number $M$ of models used to fit the predictor. The prediction error falls as $M$ grows (\cref{fig:d1_mae_scaling}), paralleling LLM studies that estimate a benchmark score from a few of its items~\citep{polo2024tinybenchmarks,disco2026}; our setting instead predicts across benchmarks. Because VLM leaderboards still have sparse cross-benchmark coverage, these estimates may improve as coverage grows. The predictors are full benchmark scores, not the released suite (\cref{app:cross_benchmark}).

\begin{revblock}
\paragraph{(5) Another use: ranking a few checkpoints of one model.}\label{par:disc_ckpt}
Model developers often use a benchmark to choose among checkpoints or hyper-parameters of one model rather than to rank model families, a harder test because such variants differ little. We fine-tune Qwen2-VL-2B on ScienceQA with LoRA (low-rank adaptation; rank $16$, two epochs) and form three kinds of variant sets: \emph{learning-rate grids}, the final adapters of runs with $8$ or $12$ learning rates (a wide grid~A and a narrower grid~B, whose $12$-rate version is repeated with three training seeds); a \emph{trajectory} of $8$ checkpoints spread over one run; and the $12$-rate grid~A repeated on Qwen2.5-VL-3B-Instruct. No variant takes part in item selection, and we compare their ranking on the selected items with that on the full test set. On the eight variant sets whose full-test ranking is stable (bootstrap $\rho\ge0.90$), VAW raises this $\rho$ over Variance by $0.082$ on average at $r{=}5\%$ (\cref{fig:ckpt_ranking_gain}): it is ahead in six sets, behind on the $8$-rate grid~B and level on the 3B grid, and its lead ranges from $0.04$ to $0.28$ across the three seeds of the $12$-rate grid~B (\cref{app:ckpt_ranking}). Here only $8$ or $12$ closely related variants are ranked, so a few uninformative items can more easily reorder them; VAW's gain over Variance here is about four times its gain on the broader model panel ($0.019$). The two settings differ and the variants of one run are not independent draws, so this comparison is descriptive, but it suggests that VAW may be particularly useful for ranking small sets of closely related model variants.
\end{revblock}

\begin{revblock}
\paragraph{(6) Do evaluation-side item scores also prune training data? A negative control.}\label{par:disc_training_negative}
An item score that finds informative evaluation items could in principle also select informative training data. We fine-tune a SigLIP+BERT multiple-choice model on ScienceQA training subsets chosen by six selectors (Random, Variance, VAW, DatBench~$r_{pb}$, AnchorPoints and IRT~2PL) at retentions from $5\%$ to $90\%$, with three seeds. No selector beats random sampling by more than the seed spread, and Variance trails it by $10.5$--$20.2$ points of test accuracy at $r\le30\%$ (\cref{app:d6_training_pruning}). \emph{Training selection and evaluation compression are distinct problems}: the former rewards informative training examples, the latter items that separate models. We therefore do not recommend VAW or other evaluation-side scores for training-data pruning; uniform random sampling remains the safer baseline there.
\end{revblock}

\section{Conclusion}
\label{sec:conclusion}

We presented \ptbench, a vision-aware hierarchical framework for compressing VLM evaluation across both benchmarks and items. Its four-stage pipeline reduces the items of the original $43$-benchmark suite by over $97\%$ while closely preserving model rankings, and its item selector, Vision-Aware Variance, performs best at the released $5\%$ retention on models held out from selection. We further validate the benchmark-level compression mechanism on a substantially larger external OpenVLM panel.

Overall, our results show that hierarchical compression can substantially reduce the cost of broad VLM evaluation while retaining the ranking information needed to compare new models. Practical limitations and broader implications are discussed in \cref{app:limitations,app:broader_impact}.

\bibliographystyle{iclr2027_conference}
\bibliography{references}

@inproceedings{zhang2025lmmseval,
  title={Lmms-eval: Reality check on the evaluation of large multimodal models},
  author={Zhang, Kaichen and Li, Bo and Zhang, Peiyuan and Pu, Fanyi and Cahyono, Joshua Adrian and Hu, Kairui and Liu, Shuai and Zhang, Yuanhan and Yang, Jingkang and Li, Chunyuan and others},
  booktitle={Findings of the Association for Computational Linguistics: NAACL 2025},
  year={2025}
}

@article{
meng2025vlm2vecv2,
title={{VLM2Vec-V2}: Advancing Multimodal Embedding for Videos, Images, and Visual Documents},
author={Rui Meng and Ziyan Jiang and Ye Liu and Mingyi Su and Xinyi Yang and Yuepeng Fu and Can Qin and Raghuveer Thirukovalluru and Xuan Zhang and Zeyuan Chen and Ran Xu and Caiming Xiong and Yingbo Zhou and Wenhu Chen and Semih Yavuz},
journal={Transactions on Machine Learning Research},
year={2026},
}

@article{shanbhogue2026gemini,
  title={Gemini Embedding 2: A Native Multimodal Embedding Model from Gemini},
  author={Shanbhogue, Madhuri and Li, Zhe and Zhang, Shanfeng and {\'A}brego, Gustavo Hern{\'a}ndez and Huang, Shih-Cheng and Jain, Aashi and Salz, Daniel and Goenka, Sonam and Hegde, Chaitra and Ma, Ji and others},
  journal={arXiv preprint arXiv:2605.27295},
  year={2026}
}

@article{chen2024mmstar,
  title={Are we on the right way for evaluating large vision-language models?},
  author={Chen, Lin and Li, Jinsong and Dong, Xiaoyi and Zhang, Pan and Zang, Yuhang and Chen, Zehui and Duan, Haodong and Wang, Jiaqi and Qiao, Yu and Lin, Dahua and others},
  journal={Advances in Neural Information Processing Systems},
  year={2024}
}

@inproceedings{goyal2017vqav2,
  title={Making the {V} in {VQA} Matter: Elevating the Role of Image Understanding in Visual Question Answering},
  author={Goyal, Yash and Khot, Tejas and Summers-Stay, Douglas and Batra, Dhruv and Parikh, Devi},
  booktitle={Proceedings of the IEEE conference on computer vision and pattern recognition},
  year={2017}
}

@inproceedings{hudson2019gqa,
  title={Gqa: A new dataset for real-world visual reasoning and compositional question answering},
  author={Hudson, Drew A and Manning, Christopher D},
  booktitle={2019 IEEE/CVF Conference on Computer Vision and Pattern Recognition (CVPR)},
  year={2019},
  organization={IEEE}
}

@inproceedings{mathew2021docvqa,
  title={Docvqa: A dataset for vqa on document images},
  author={Mathew, Minesh and Karatzas, Dimosthenis and Jawahar, CV},
  booktitle={2021 IEEE Winter Conference on Applications of Computer Vision (WACV)},
  year={2021},
  organization={IEEE}
}

@inproceedings{singh2019textvqa,
  title={Towards vqa models that can read},
  author={Singh, Amanpreet and Natarajan, Vivek and Shah, Meet and Jiang, Yu and Chen, Xinlei and Batra, Dhruv and Parikh, Devi and Rohrbach, Marcus},
  booktitle={2019 IEEE/CVF Conference on Computer Vision and Pattern Recognition (CVPR)},
  year={2019},
  organization={IEEE}
}

@inproceedings{masry2022chartqa,
  title={Chartqa: A benchmark for question answering about charts with visual and logical reasoning},
  author={Masry, Ahmed and Tan, Jia Qing and Joty, Shafiq and Hoque, Enamul and others},
  booktitle={Findings of the association for computational linguistics: ACL 2022},
  year={2022}
}

@inproceedings{yue2024mmmu,
  title={Mmmu: A massive multi-discipline multimodal understanding and reasoning benchmark for expert agi},
  author={Yue, Xiang and Ni, Yuansheng and Zhang, Kai and Zheng, Tianyu and Liu, Ruoqi and Zhang, Ge and Stevens, Samuel and Jiang, Dongfu and Ren, Weiming and Sun, Yuxuan and others},
  booktitle={Proceedings of the IEEE/CVF conference on computer vision and pattern recognition},
  year={2024}
}

@inproceedings{liu2024mmbench,
  title={Mmbench: Is your multi-modal model an all-around player?},
  author={Liu, Yuan and Duan, Haodong and Zhang, Yuanhan and Li, Bo and Zhang, Songyang and Zhao, Wangbo and Yuan, Yike and Wang, Jiaqi and He, Conghui and Liu, Ziwei and others},
  booktitle={European conference on computer vision},
  year={2024},
  organization={Springer}
}

@inproceedings{fu2025mme,
  title={{MME}: A Comprehensive Evaluation Benchmark for Multimodal Large Language Models},
  author={Fu, Chaoyou and Chen, Peixian and Shen, Yunhang and Qin, Yulei and Zhang, Mengdan and Lin, Xu and Yang, Jinrui and Zheng, Xiawu and Li, Ke and Sun, Xing and others},
  booktitle={Advances in Neural Information Processing Systems (NeurIPS)},
  year={2025}
}

@article{li2023seedbench,
  title={Seed-bench: Benchmarking multimodal llms with generative comprehension},
  author={Li, Bohao and Wang, Rui and Wang, Guangzhi and Ge, Yuying and Ge, Yixiao and Shan, Ying},
  journal={arXiv preprint arXiv:2307.16125},
  year={2023}
}

@article{lu2022scienceqa,
  title={Learn to explain: Multimodal reasoning via thought chains for science question answering},
  author={Lu, Pan and Mishra, Swaroop and Xia, Tanglin and Qiu, Liang and Chang, Kai-Wei and Zhu, Song-Chun and Tafjord, Oyvind and Clark, Peter and Kalyan, Ashwin},
  journal={Advances in neural information processing systems},
  year={2022}
}

@inproceedings{li2023pope,
  title={Evaluating object hallucination in large vision-language models},
  author={Li, Yifan and Du, Yifan and Zhou, Kun and Wang, Jinpeng and Zhao, Xin and Wen, Ji-Rong},
  booktitle={Proceedings of the 2023 conference on empirical methods in natural language processing},
  year={2023}
}

@inproceedings{lu2024mathvista,
  title={Mathvista: Evaluating mathematical reasoning of foundation models in visual contexts},
  author={Lu, Pan and Bansal, Hritik and Xia, Tony and Liu, Jiacheng and Li, Chunyuan and Hajishirzi, Hannaneh and Cheng, Hao and Chang, Kai-Wei and Galley, Michel and Gao, Jianfeng},
  booktitle={International Conference on Learning Representations},
  year={2024}
}

@inproceedings{kembhavi2016ai2d,
  title={A diagram is worth a dozen images},
  author={Kembhavi, Aniruddha and Salvato, Mike and Kolve, Eric and Seo, Minjoon and Hajishirzi, Hannaneh and Farhadi, Ali},
  booktitle={European conference on computer vision},
  year={2016},
  organization={Springer}
}

@article{liu2024ocrbench,
  title={Ocrbench: on the hidden mystery of ocr in large multimodal models},
  author={Liu, Yuliang and Li, Zhang and Huang, Mingxin and Yang, Biao and Yu, Wenwen and Li, Chunyuan and Yin, Xu-Cheng and Liu, Cheng-Lin and Jin, Lianwen and Bai, Xiang},
  journal={Science China Information Sciences},
  year={2024},
  publisher={Springer}
}

@inproceedings{mathew2022infographicvqa,
  title={Infographicvqa},
  author={Mathew, Minesh and Bagal, Viraj and Tito, Rub{\`e}n and Karatzas, Dimosthenis and Valveny, Ernest and Jawahar, CV},
  booktitle={2022 IEEE/CVF Winter Conference on Applications of Computer Vision (WACV)},
  year={2022},
  organization={IEEE}
}

@inproceedings{duan2024vlmevalkit,
  title={Vlmevalkit: An open-source toolkit for evaluating large multi-modality models},
  author={Duan, Haodong and Yang, Junming and Qiao, Yuxuan and Fang, Xinyu and Chen, Lin and Liu, Yuan and Dong, Xiaoyi and Zang, Yuhang and Zhang, Pan and Wang, Jiaqi and others},
  booktitle={Proceedings of the 32nd ACM international conference on multimedia},
  year={2024}
}

@inproceedings{kiela2021dynabench,
  title={Dynabench: Rethinking benchmarking in NLP},
  author={Kiela, Douwe and Bartolo, Max and Nie, Yixin and Kaushik, Divyansh and Geiger, Atticus and Wu, Zhengxuan and Vidgen, Bertie and Prasad, Grusha and Singh, Amanpreet and Ringshia, Pratik and others},
  booktitle={Proceedings of the 2021 conference of the North American chapter of the Association for Computational Linguistics: human language technologies},
  year={2021}
}

@inproceedings{
polo2024tinybenchmarks,
title={tinyBenchmarks: evaluating {LLM}s with fewer examples},
author={Felipe Maia Polo and Lucas Weber and Leshem Choshen and Yuekai Sun and Gongjun Xu and Mikhail Yurochkin},
booktitle={Forty-first International Conference on Machine Learning},
year={2024},
}

@inproceedings{kipnis2025metabench,
  title={metabench-A Sparse Benchmark of Reasoning and Knowledge in Large Language Models},
  author={Kipnis, Alex and Voudouris, Konstantinos and Schulze Buschoff, Luca and Schulz, Eric},
  booktitle={International Conference on Learning Representations},
  year={2025}
}

@inproceedings{saranathan2025sublime,
    title = "{S}ub{LIME}: Subset Selection via Rank Correlation Prediction for Data-Efficient {LLM} Evaluation",
    author = "Saranathan, Gayathri  and
      Xu, Cong  and
      Alam, Mahammad Parwez  and
      Kumar, Tarun  and
      Foltin, Martin  and
      Wong, Soon Yee  and
      Bhattacharya, Suparna",
    booktitle = "Proceedings of the 63rd Annual Meeting of the Association for Computational Linguistics (Volume 1: Long Papers)",
    year = "2025",
}

@inproceedings{disco2026,
  title={Disco: Diversifying sample condensation for efficient model evaluation},
  author={Rubinstein, Alexander and Raible, Benjamin and Gubri, Martin and Oh, Seong Joon},
  booktitle={International Conference on Learning Representations},
  year={2026}
}

@inproceedings{essencebench2026,
  title={Rethinking LLM Evaluation: Can We Evaluate LLMs with 200$\times$ Less Data?},
  author={Wang, Shaobo and Wang, Cong and Fu, Wenjie and Min, Yue and Feng, Mingquan and Guan, Isabel and Hu, Xuming and He, Conghui and Wang, Cunxiang and Yang, Kexin and others},
  booktitle={International Conference on Learning Representations},
  year={2026}
}

@inproceedings{
sparseeval2026,
title={SparseEval: Efficient Evaluation of Large Language Models by Sparse Optimization},
author={Taolin Zhang and Hang Guo and Wang Lu and Tao Dai and Shu-Tao Xia and Jindong Wang},
booktitle={The Fourteenth International Conference on Learning Representations},
year={2026},
}

@inproceedings{vivek2024anchor,
  title={Anchor points: Benchmarking models with much fewer examples},
  author={Vivek, Rajan and Ethayarajh, Kawin and Yang, Diyi and Kiela, Douwe},
  booktitle={Proceedings of the 18th Conference of the European Chapter of the Association for Computational Linguistics (Volume 1: Long Papers)},
  year={2024}
}

@article{bai2023qwenvl,
  title={{Qwen-VL}: A Versatile Vision-Language Model for Understanding, Localization, Text Reading, and Beyond},
  author={Jinze Bai and Shuai Bai and Shusheng Yang and Shijie Wang and Sinan Tan and Peng Wang and Junyang Lin and Chang Zhou and Jingren Zhou},
  journal={arXiv preprint arXiv:2308.12966},
  year={2023}
}

@article{wang2024qwen2vl,
  title={Qwen2-vl: Enhancing vision-language model's perception of the world at any resolution},
  author={Wang, Peng and Bai, Shuai and Tan, Sinan and Wang, Shijie and Fan, Zhihao and Bai, Jinze and Chen, Keqin and Liu, Xuejing and Wang, Jialin and Ge, Wenbin and others},
  journal={arXiv preprint arXiv:2409.12191},
  year={2024}
}

@article{qwen25vl2025,
  title={{Qwen2.5-VL} Technical Report},
  author={Shuai Bai and Keqin Chen and Xuejing Liu and Jialin Wang and Wenbin Ge and Sibo Song and Kai Dang and Peng Wang and Shijie Wang and Jun Tang and Humen Zhong and Yuanzhi Zhu and Mingkun Yang and Zhaohai Li and Jianqiang Wan and Pengfei Wang and Wei Ding and Zheren Fu and Yiheng Xu and Jiabo Ye and Xi Zhang and Tianbao Xie and Zesen Cheng and Hang Zhang and Zhibo Yang and Haiyang Xu and Junyang Lin and others},
  journal={arXiv preprint arXiv:2502.13923},
  year={2025}
}

@article{qwen3vl2025,
  title={Qwen3-vl technical report},
  author={Bai, Shuai and Cai, Yuxuan and Chen, Ruizhe and Chen, Keqin and Chen, Xionghui and Cheng, Zesen and Deng, Lianghao and Ding, Wei and Gao, Chang and Ge, Chunjiang and others},
  journal={arXiv preprint arXiv:2511.21631},
  year={2025}
}

@article{chen2024expanding,
  title={Expanding performance boundaries of open-source multimodal models with model, data, and test-time scaling},
  author={Chen, Zhe and Wang, Weiyun and Cao, Yue and Liu, Yangzhou and Gao, Zhangwei and Cui, Erfei and Zhu, Jinguo and Ye, Shenglong and Tian, Hao and Liu, Zhaoyang and others},
  journal={arXiv preprint arXiv:2412.05271},
  year={2024}
}

@inproceedings{mao2016generation,
  title={Generation and comprehension of unambiguous object descriptions},
  author={Mao, Junhua and Huang, Jonathan and Toshev, Alexander and Camburu, Oana and Yuille, Alan L and Murphy, Kevin},
  booktitle={Proceedings of the IEEE conference on computer vision and pattern recognition},
  year={2016}
}

@article{chen2024internvl2,
  title={How far are we to gpt-4v? closing the gap to commercial multimodal models with open-source suites},
  author={Chen, Zhe and Wang, Weiyun and Tian, Hao and Ye, Shenglong and Gao, Zhangwei and Cui, Erfei and Tong, Wenwen and Hu, Kongzhi and Luo, Jiapeng and Ma, Zheng and others},
  journal={Science China Information Sciences},
  year={2024},
  publisher={Springer}
}

@article{internvl3_2025,
  title={Internvl3: Exploring advanced training and test-time recipes for open-source multimodal models},
  author={Zhu, Jinguo and Wang, Weiyun and Chen, Zhe and Liu, Zhaoyang and Ye, Shenglong and Gu, Lixin and Tian, Hao and Duan, Yuchen and Su, Weijie and Shao, Jie and others},
  journal={arXiv preprint arXiv:2504.10479},
  year={2025}
}

@article{internvl3_5_2025,
  title={Internvl3. 5: Advancing open-source multimodal models in versatility, reasoning, and efficiency},
  author={Wang, Weiyun and Gao, Zhangwei and Gu, Lixin and Pu, Hengjun and Cui, Long and Wei, Xingguang and Liu, Zhaoyang and Jing, Linglin and Ye, Shenglong and Shao, Jie and others},
  journal={arXiv preprint arXiv:2508.18265},
  year={2025}
}

@article{glm4_1v_2025,
  title={Glm-4.5 v and glm-4.1 v-thinking: Towards versatile multimodal reasoning with scalable reinforcement learning},
  author={Hong, Wenyi and Yu, Wenmeng and Gu, Xiaotao and Wang, Guo and Gan, Guobing and Tang, Haomiao and Cheng, Jiale and Qi, Ji and Ji, Junhui and Pan, Lihang and others},
  journal={arXiv preprint arXiv:2507.01006},
  year={2025}
}

@article{team2025gemma3,
  title={Gemma 3 technical report},
  author={Team, Gemma and Kamath, Aishwarya and Ferret, Johan and Pathak, Shreya and Vieillard, Nino and Merhej, Ramona and Perrin, Sarah and Matejovicova, Tatiana and Ram{\'e}, Alexandre and Rivi{\`e}re, Morgane and others},
  journal={arXiv preprint arXiv:2503.19786},
  year={2025}
}

@misc{realworldqa2024,
  title = {{RealWorldQA}},
  author = {{xAI}},
  year = {2024},
  howpublished = {\url{https://huggingface.co/datasets/xai-org/RealworldQA}},
}

@inproceedings{marino2019okvqa,
  title={Ok-vqa: A visual question answering benchmark requiring external knowledge},
  author={Marino, Kenneth and Rastegari, Mohammad and Farhadi, Ali and Mottaghi, Roozbeh},
  booktitle={2019 IEEE/CVF conference on computer vision and pattern recognition (CVPR)},
  year={2019},
  organization={IEEE}
}

@article{zhang2024cmmmu,
  title={Cmmmu: A chinese massive multi-discipline multimodal understanding benchmark},
  author={Zhang, Ge and Du, Xinrun and Chen, Bei and Liang, Yiming and Luo, Tongxu and Zheng, Tianyu and Zhu, Kang and Cheng, Yuyang and Xu, Chunpu and Guo, Shuyue and others},
  journal={arXiv preprint arXiv:2401.11944},
  year={2024}
}

@article{cvbench2024,
  title={Cambrian-1: A fully open, vision-centric exploration of multimodal llms},
  author={Tong, Shengbang and Brown, Ellis and Wu, Penghao and Woo, Sanghyun and Middepogu, Manoj and Akula, Sai C and Yang, Jihan and Yang, Shusheng and Iyer, Adithya and Pan, Xichen and others},
  journal={Advances in Neural Information Processing Systems},
  year={2024}
}

@article{joshi2026datbench,
  title={Datbench: Discriminative, faithful, and efficient vlm evaluations},
  author={Joshi, Siddharth and Yin, Haoli and Adiga, Rishabh and Monti, Ricardo and Carranza, Aldo and Fang, Alex and Deng, Alvin and Abbas, Amro and Larsen, Brett and Blakeney, Cody and others},
  journal={arXiv preprint arXiv:2601.02316},
  year={2026}
}

@article{wang2024charxiv,
  title={Charxiv: Charting gaps in realistic chart understanding in multimodal llms},
  author={Wang, Zirui and Xia, Mengzhou and He, Luxi and Chen, Howard and Liu, Yitao and Zhu, Richard and Liang, Kaiqu and Wu, Xindi and Liu, Haotian and Malladi, Sadhika and others},
  journal={Advances in Neural Information Processing Systems},
  year={2024}
}

@article{spearman1904,
  title={The Proof and Measurement of Association between Two Things},
  author={Spearman, Charles},
  journal={The American Journal of Psychology},
  year={1904},
}

@inproceedings{
ying2024mmtbench,
title={{MMT}-Bench: A Comprehensive Multimodal Benchmark for Evaluating Large Vision-Language Models Towards Multitask {AGI}},
author={Kaining Ying and Fanqing Meng and Jin Wang and Zhiqian Li and Han Lin and Yue Yang and Hao Zhang and Wenbo Zhang and Yuqi Lin and Shuo Liu and jiayi lei and Quanfeng Lu and Runjian Chen and Peng Xu and Renrui Zhang and Haozhe Zhang and Peng Gao and Yali Wang and Yu Qiao and Ping Luo and Kaipeng Zhang and Wenqi Shao},
booktitle={Forty-first International Conference on Machine Learning},
year={2024},
}

@article{teamgemini2025erqa,
  title={Gemini robotics: Bringing ai into the physical world},
  author={Team, Gemini Robotics and Abeyruwan, Saminda and Ainslie, Joshua and Alayrac, Jean-Baptiste and Arenas, Montserrat Gonzalez and Armstrong, Travis and Balakrishna, Ashwin and Baruch, Robert and Bauza, Maria and Blokzijl, Michiel and others},
  journal={arXiv preprint arXiv:2503.20020},
  year={2025}
}

@inproceedings{guan2024hallusionbench,
  title={Hallusionbench: an advanced diagnostic suite for entangled language hallucination and visual illusion in large vision-language models},
  author={Guan, Tianrui and Liu, Fuxiao and Wu, Xiyang and Xian, Ruiqi and Li, Zongxia and Liu, Xiaoyu and Wang, Xijun and Chen, Lichang and Huang, Furong and Yacoob, Yaser and others},
  booktitle={2024 IEEE/CVF Conference on Computer Vision and Pattern Recognition (CVPR)},
  year={2024},
  organization={IEEE}
}

@InProceedings{yu2024mmvet,
  title = 	 {{MM}-Vet: Evaluating Large Multimodal Models for Integrated Capabilities},
  author =       {Yu, Weihao and Yang, Zhengyuan and Li, Linjie and Wang, Jianfeng and Lin, Kevin and Liu, Zicheng and Wang, Xinchao and Wang, Lijuan},
  booktitle = 	 {Proceedings of the 41st International Conference on Machine Learning},
  year = 	 {2024},
}

@inproceedings{
liu2024visualwebbench,
title={VisualWebBench: How Far Have Multimodal {LLM}s Evolved in Web Page Understanding and Grounding?},
author={Junpeng Liu and Yifan Song and Bill Yuchen Lin and Wai Lam and Graham Neubig and Yuanzhi Li and Xiang Yue},
booktitle={First Conference on Language Modeling},
year={2024},
}

@inproceedings{chen2021websrc,
  title={Websrc: A dataset for web-based structural reading comprehension},
  author={Chen, Xingyu and Zhao, Zihan and Chen, Lu and Ji, JiaBao and Zhang, Danyang and Luo, Ao and Xiong, Yuxuan and Yu, Kai},
  booktitle={Proceedings of the 2021 Conference on Empirical Methods in Natural Language Processing},
  year={2021}
}

@article{wang2023amber,
  title={Amber: An llm-free multi-dimensional benchmark for mllms hallucination evaluation},
  author={Wang, Junyang and Wang, Yuhang and Xu, Guohai and Zhang, Jing and Gu, Yukai and Jia, Haitao and Wang, Jiaqi and Xu, Haiyang and Yan, Ming and Zhang, Ji and others},
  journal={arXiv preprint arXiv:2311.07397},
  year={2023}
}

@inproceedings{agrawal2019nocaps,
  title={Nocaps: Novel object captioning at scale},
  author={Agrawal, Harsh and Desai, Karan and Wang, Yufei and Chen, Xinlei and Jain, Rishabh and Johnson, Mark and Batra, Dhruv and Parikh, Devi and Lee, Stefan and Anderson, Peter},
  booktitle={2019 IEEE/CVF International Conference on Computer Vision (ICCV)},
  year={2019},
  organization={IEEE}
}

@inproceedings{sidorov2020textcaps,
  title={Textcaps: a dataset for image captioning with reading comprehension},
  author={Sidorov, Oleksii and Hu, Ronghang and Rohrbach, Marcus and Singh, Amanpreet},
  booktitle={European conference on computer vision},
  year={2020},
  organization={Springer}
}

@article{chen2015cococaptions,
  title={Microsoft coco captions: Data collection and evaluation server},
  author={Chen, Xinlei and Fang, Hao and Lin, Tsung-Yi and Vedantam, Ramakrishna and Gupta, Saurabh and Doll{\'a}r, Piotr and Zitnick, C Lawrence},
  journal={arXiv preprint arXiv:1504.00325},
  year={2015}
}

@inproceedings{qian2025miabench,
  title={Mia-bench: Towards better instruction following evaluation of multimodal llms},
  author={Qian, Yusu and Ye, Hanrong and Fauconnier, Jean-Philippe and Grasch, Peter and Yang, Yinfei and Gan, Zhe},
  booktitle={International Conference on Learning Representations},
  year={2025}
}

@article{zhang2025sparbench,
  title={From flatland to space: Teaching vision-language models to perceive and reason in 3d},
  author={Zhang, Jiahui and Chen, Yurui and Xu, Yueming and Huang, Ze and Mei, Jilin and Chen, Chunhui and Zhou, Yanpeng and Yuan, Yu-Jie and Cai, Xinyue and Huang, Guowei and others},
  journal={Advances in Neural Information Processing Systems},
  year={2025}
}

@inproceedings{yue2025mmmupro,
  title={Mmmu-pro: A more robust multi-discipline multimodal understanding benchmark},
  author={Yue, Xiang and Zheng, Tianyu and Ni, Yuansheng and Wang, Yubo and Zhang, Kai and Tong, Shengbang and Sun, Yuxuan and Yu, Botao and Zhang, Ge and Sun, Huan and others},
  booktitle={Proceedings of the 63rd Annual Meeting of the Association for Computational Linguistics (Volume 1: Long Papers)},
  year={2025}
}

@article{uebayashi2026m3irt,
  title={Evaluating Cross-Modal Reasoning Ability and Problem Characteristics with Multimodal Item Response Theory},
  author={Uebayashi, Shunki and Masui, Kento and Atarashi, Kyohei and Bao, Han and Kashima, Hisashi and Inoue, Naoto and Otani, Mayu and Takeuchi, Koh},
  journal={arXiv preprint arXiv:2603.02663},
  year={2026}
}

@inproceedings{cheng2024seeclick,
  title={Seeclick: Harnessing gui grounding for advanced visual gui agents},
  author={Cheng, Kanzhi and Sun, Qiushi and Chu, Yougang and Xu, Fangzhi and YanTao, Li and Zhang, Jianbing and Wu, Zhiyong},
  booktitle={Proceedings of the 62nd Annual Meeting of the Association for Computational Linguistics (Volume 1: Long Papers)},
  year={2024}
}

@inproceedings{koh2024visualwebarena,
  title={Visualwebarena: Evaluating multimodal agents on realistic visual web tasks},
  author={Koh, Jing Yu and Lo, Robert and Jang, Lawrence and Duvvur, Vikram and Lim, Ming and Huang, Po-Yu and Neubig, Graham and Zhou, Shuyan and Salakhutdinov, Russ and Fried, Daniel},
  booktitle={Proceedings of the 62nd Annual Meeting of the Association for Computational Linguistics (Volume 1: Long Papers)},
  year={2024}
}

@article{liang2023helm,
  title={Holistic evaluation of language models},
  author={Liang, Percy and Bommasani, Rishi and Lee, Tony and Tsipras, Dimitris and Soylu, Dilara and Yasunaga, Michihiro and Zhang, Yian and Narayanan, Deepak and Wu, Yuhuai and Kumar, Ananya and others},
  journal={arXiv preprint arXiv:2211.09110},
  year={2022}
}

@article{tsybakov2004,
  title={Optimal Aggregation of Classifiers in Statistical Learning},
  author={Alexandre B. Tsybakov},
  journal={The Annals of Statistics},
  year={2004},
}

@article{qwen3vl_embedding_2026,
  title={Qwen3-vl-embedding and qwen3-vl-reranker: A unified framework for state-of-the-art multimodal retrieval and ranking},
  author={Li, Mingxin and Zhang, Yanzhao and Long, Dingkun and Chen, Keqin and Song, Sibo and Bai, Shuai and Yang, Zhibo and Xie, Pengjun and Yang, An and Liu, Dayiheng and others},
  journal={arXiv preprint arXiv:2601.04720},
  year={2026}
}

@article{teamgemma4_2026,
  title={Gemma 4 technical report},
  author={Team, Gemma and Abd, Sherif El and Aggarwal, Vaibhav and Algayres, Robin and Andreev, Alek and Bachem, Olivier and Ballantyne, Ian and Brick, Cormac and C{\u{a}}rbune, Victor and Casbon, Michelle and others},
  journal={arXiv preprint arXiv:2607.02770},
  year={2026}
}

@inproceedings{fu2024blink,
  title={Blink: Multimodal large language models can see but not perceive},
  author={Fu, Xingyu and Hu, Yushi and Li, Bangzheng and Feng, Yu and Wang, Haoyu and Lin, Xudong and Roth, Dan and Smith, Noah A and Ma, Wei-Chiu and Krishna, Ranjay},
  booktitle={European Conference on Computer Vision},
  year={2024},
  organization={Springer}
}

@article{wang2024mathvision,
  title={Measuring multimodal mathematical reasoning with math-vision dataset},
  author={Wang, Ke and Pan, Junting and Shi, Weikang and Lu, Zimu and Ren, Houxing and Zhou, Aojun and Zhan, Mingjie and Li, Hongsheng},
  journal={Advances in Neural Information Processing Systems},
  year={2024}
}

@inproceedings{he2024olympiad,
  title={Olympiadbench: A challenging benchmark for promoting agi with olympiad-level bilingual multimodal scientific problems},
  author={He, Chaoqun and Luo, Renjie and Bai, Yuzhuo and Hu, Shengding and Thai, Zhen and Shen, Junhao and Hu, Jinyi and Han, Xu and Huang, Yujie and Zhang, Yuxiang and others},
  booktitle={Proceedings of the 62nd Annual Meeting of the Association for Computational Linguistics (Volume 1: Long Papers)},
  year={2024}
}

@inproceedings{kazemzadeh2014refcoco,
    title = "{R}efer{I}t{G}ame: Referring to Objects in Photographs of Natural Scenes",
    author = "Kazemzadeh, Sahar  and
      Ordonez, Vicente  and
      Matten, Mark  and
      Berg, Tamara",
    booktitle = "Proceedings of the 2014 Conference on Empirical Methods in Natural Language Processing ({EMNLP})",
    year = "2014",
}

@inproceedings{yu2016refcocog,
  title={Modeling context in referring expressions},
  author={Yu, Licheng and Poirson, Patrick and Yang, Shan and Berg, Alexander C and Berg, Tamara L},
  booktitle={European conference on computer vision},
  year={2016},
  organization={Springer}
}

@article{liu2023llava,
  title={Visual instruction tuning},
  author={Liu, Haotian and Li, Chunyuan and Wu, Qingyang and Lee, Yong Jae},
  journal={Advances in neural information processing systems},
  year={2023}
}

@article{xu2025qwen25omni,
  title={Qwen2.5-Omni Technical Report}, 
  author={Jin Xu and Zhifang Guo and Jinzheng He and Hangrui Hu and Ting He and Shuai Bai and Keqin Chen and Jialin Wang and Yang Fan and Kai Dang and Bin Zhang and Xiong Wang and Yunfei Chu and Junyang Lin},
  journal={arXiv preprint arXiv:2503.20215},
  year={2025},
  url={https://arxiv.org/abs/2503.20215}
}

@article{xu2025qwen3omni,
  title={Qwen3-omni technical report},
  author={Xu, Jin and Guo, Zhifang and Hu, Hangrui and Chu, Yunfei and Wang, Xiong and He, Jinzheng and Wang, Yuxuan and Shi, Xian and He, Ting and Zhu, Xinfa and others},
  journal={arXiv preprint arXiv:2509.17765},
  year={2025}
}

@article{
li2024llavaonevision,
title={{LL}a{VA}-OneVision: Easy Visual Task Transfer},
author={Bo Li and Yuanhan Zhang and Dong Guo and Renrui Zhang and Feng Li and Hao Zhang and Kaichen Zhang and Peiyuan Zhang and Yanwei Li and Ziwei Liu and Chunyuan Li},
journal={Transactions on Machine Learning Research},
issn={2835-8856},
year={2025}
}

@article{an2025llavaonevision15,
  title={Llava-onevision-1.5: Fully open framework for democratized multimodal training},
  author={An, Xiang and Xie, Yin and Yang, Kaicheng and Zhang, Wenkang and Zhao, Xiuwei and Cheng, Zheng and Wang, Yirui and Xu, Songcen and Chen, Changrui and Zhu, Didi and others},
  journal={arXiv preprint arXiv:2509.23661},
  year={2025}
}

@article{an2026llavaonevision2,
  title={Llava-onevision-2: Towards next-generation perceptual intelligence},
  author={An, Xiang and Xie, Yin and Tang, Feilong and Yan, Yunyao and Tan, Huajie and Zhu, Didi and Chen, Changrui and Zhao, Xiuwei and Qin, Bin and Yang, Kaicheng and others},
  journal={arXiv preprint arXiv:2605.25979},
  year={2026}
}

@article{marafioti2025smolvlm,
  title={Smolvlm: Redefining small and efficient multimodal models},
  author={Marafioti, Andr{\'e}s and Zohar, Orr and Farr{\'e}, Miquel and Noyan, Merve and Bakouch, Elie and Cuenca, Pedro and Zakka, Cyril and Allal, Loubna Ben and Lozhkov, Anton and Tazi, Nouamane and others},
  journal={arXiv preprint arXiv:2504.05299},
  year={2025}
}

@article{team2025kimivl,
  title={Kimi-vl technical report},
  author={Team, Kimi and Du, Angang and Yin, Bohong and Xing, Bowei and Qu, Bowen and Wang, Bowen and Chen, Cheng and Zhang, Chenlin and Du, Chenzhuang and Wei, Chu and others},
  journal={arXiv preprint arXiv:2504.07491},
  year={2025}
}

@inproceedings{yu2025minicpmv45,
  title={Minicpm-v 4.5: Cooking efficient mllms via architecture, data, and training recipe},
  author={Yu, Tianyu and Wang, Zefan and Wang, Chongyi and Huang, Fuwei and Ma, Wenshuo and He, Zhihui and Cai, Tianchi and Chen, Weize and Huang, Yuxiang and Zhao, Ranchi and others},
  booktitle={Proceedings of the IEEE/CVF Conference on Computer Vision and Pattern Recognition},
  year={2026}
}

@article{naver2026hyperclovaomni,
  title={Hyperclova x 8b omni},
  author={Team, NAVER Cloud HyperCLOVA X},
  journal={arXiv preprint arXiv:2601.01792},
  year={2026}
}

@article{meta2024llama32vision,
  title={The llama 3 herd of models},
  author={Grattafiori, Aaron and Dubey, Abhimanyu and Jauhri, Abhinav and Pandey, Abhinav and Kadian, Abhishek and Al-Dahle, Ahmad and Letman, Aiesha and Mathur, Akhil and Schelten, Alan and Vaughan, Alex and others},
  journal={arXiv preprint arXiv:2407.21783},
  year={2024}
}

@misc{openbmb2026minicpmv46,
  title={{MiniCPM-V 4.6}},
  author={{OpenBMB}},
  year={2026},
  howpublished={\url{https://huggingface.co/openbmb/MiniCPM-V-4.6}},
  note={Official model card; accessed 2026-09-25}
}

\clearpage

\begin{appendices}
\onecolumn

\etocdepthtag.toc{mtappendix}
\etocsettagdepth{mtchapter}{none}
\etocsettagdepth{mtappendix}{subsection}

\begin{center}{\Large\bf Appendix}\end{center}
\vspace{0.5em}

\hypersetup{linkcolor=black}
\tableofcontents
\hypersetup{linkcolor={[rgb]{0.20,0.30,0.65}}}


\section{Model and Benchmark Details}
\label{app:details}

\subsection{Model details}
\label{app:models}

\begin{revblock}
The model panel has $30$ \rev{vision--language and omni models} from ten families (\cref{tab:models}). Selection experiments split the panel into build models that choose the items and held-out models that are scored (\cref{app:unseen_protocol}). Every model is evaluated through lmms-eval~\citep{zhang2025lmmseval} with greedy decoding.
\end{revblock}

\begin{table}[!htbp]
  \caption{\rev{\textbf{The panel has $30$ vision--language and omni models from ten families.}}}
  \label{tab:models}
  \centering
  \small
  \setlength{\tabcolsep}{4pt}
  \begin{revblock}
  \begin{adjustbox}{max width=\linewidth}
\begin{tabular}{ll}
    \toprule
    Model & Family \\
    \midrule
    Qwen-VL-Chat~\citep{bai2023qwenvl}           & Qwen \\
    Qwen2-VL-7B-Instruct~\citep{wang2024qwen2vl} & Qwen \\
    Qwen2.5-VL-7B-Instruct~\citep{qwen25vl2025}  & Qwen \\
    Qwen2.5-Omni-3B~\rev{\citep{xu2025qwen25omni}} & Qwen \\
    Qwen2.5-Omni-7B~\rev{\citep{xu2025qwen25omni}} & Qwen \\
    Qwen3-VL-2B-Instruct~\citep{qwen3vl2025}     & Qwen \\
    Qwen3-VL-4B-Instruct~\citep{qwen3vl2025}     & Qwen \\
    Qwen3-VL-8B-Instruct~\citep{qwen3vl2025}     & Qwen \\
    Qwen3-VL-30B-A3B-Instruct~\rev{\citep{qwen3vl2025}} & Qwen \\
    Qwen3-Omni-30B-A3B-Instruct~\rev{\citep{xu2025qwen3omni}} & Qwen \\
    InternVL2-8B~\citep{chen2024internvl2}       & InternVL \\
    InternVL2.5-8B~\citep{chen2024expanding}     & InternVL \\
    InternVL3-8B~\citep{internvl3_2025}          & InternVL \\
    InternVL3.5-8B~\citep{internvl3_5_2025}      & InternVL \\
    Gemma3-4B-IT~\citep{team2025gemma3}          & Gemma3 \\
    Gemma3-12B-IT~\citep{team2025gemma3}         & Gemma3 \\
    Gemma4-E2B-IT~\citep{teamgemma4_2026}        & Gemma4 \\
    Gemma4-E4B-IT~\citep{teamgemma4_2026}        & Gemma4 \\
    Gemma4-26B-A4B-IT~\citep{teamgemma4_2026}    & Gemma4 \\
    Gemma4-31B-IT~\citep{teamgemma4_2026}        & Gemma4 \\
    LLaVA-OneVision-Qwen2-7B-OV~\rev{\citep{li2024llavaonevision}} & LLaVA-OneVision \\
    LLaVA-OneVision-1.5-4B-Instruct~\rev{\citep{an2025llavaonevision15}} & LLaVA-OneVision \\
    LLaVA-OneVision-1.5-8B-Instruct~\rev{\citep{an2025llavaonevision15}} & LLaVA-OneVision \\
    LLaVA-OneVision-2-8B-Instruct~\rev{\citep{an2026llavaonevision2}} & LLaVA-OneVision \\
    Llama-3.2-11B-Vision-Instruct~\rev{\citep{meta2024llama32vision}} & Llama \\
    SmolVLM2-2.2B-Instruct~\rev{\citep{marafioti2025smolvlm}} & SmolVLM \\
    Kimi-VL-A3B-Instruct~\rev{\citep{team2025kimivl}} & Kimi-VL \\
    MiniCPM-V-4.5~\rev{\citep{yu2025minicpmv45}} & MiniCPM-V \\
    MiniCPM-V-4.6~\rev{\citep{openbmb2026minicpmv46}} & MiniCPM-V \\
    HyperCLOVAX-SEED-Omni-8B~\rev{\citep{naver2026hyperclovaomni}} & HyperCLOVAX \\
    \bottomrule
  \end{tabular}
\end{adjustbox}
  \end{revblock}
\end{table}

\subsection{Benchmark details}
\label{app:benchmarks}

\cref{tab:benchmark_details} lists all benchmarks in the catalogue, grouped by the \rev{$18$} capability categories used in Stage~1.

\begin{table}[!htbp]
  \caption{\textbf{The catalogue mixes binary, continuous, GPT-judged, and aggregate scoring.} \rev{Scoring = the per-benchmark metric that feeds Stage~1; Items = the benchmark's items before Stage~0 cleaning ($386{,}324$ in total); MCQ = multiple-choice question, CoT = chain-of-thought, ANLS = average normalised Levenshtein similarity, MRA = mean relative accuracy, aAcc = HallusionBench all-question accuracy, Cover = AMBER object coverage, GPT = judged by a GPT model.}}
  \label{tab:benchmark_details}
  \centering
  \small
  \setlength{\tabcolsep}{4pt}
  \begin{revblock}
  \begin{adjustbox}{max width=\linewidth}
\begin{tabular}{lllr}
    \toprule
    Category & Benchmark & Scoring & Items \\
    \midrule
    Comprehensive & MME~\citep{fu2025mme} & Perception aggregate & $2{,}374$ \\
     & MMBench~\citep{liu2024mmbench} & GPT & $4{,}329$ \\
     & MM-Vet~\citep{yu2024mmvet} & GPT & $218$ \\
    \midrule
    Academic Knowledge & MMMU~\citep{yue2024mmmu} & MCQ exact match & $900$ \\
     & MMMU-Pro~\citep{yue2025mmmupro} & MCQ exact match & $1{,}730$ \\
     & CMMMU~\citep{zhang2024cmmmu} & MCQ exact match & $900$ \\
    \midrule
    Visual Reasoning & VQAv2~\citep{goyal2017vqav2} & Exact match & $214{,}354$ \\
     & GQA~\citep{hudson2019gqa} & Exact match & $12{,}578$ \\
     & BLINK~\citep{fu2024blink} & Aggregate & $1{,}901$ \\
    \midrule
    Math Reasoning & MathVista~\citep{lu2024mathvista} & CoT, GPT & $1{,}000$ \\
     & MathVision~\citep{wang2024mathvision} & Exact match & $304$ \\
     & OlympiadBench~\citep{he2024olympiad} & GPT & $150$ \\
    \midrule
    Document Understanding & DocVQA~\citep{mathew2021docvqa} & ANLS & $5{,}349$ \\
     & InfoVQA~\citep{mathew2022infographicvqa} & ANLS & $2{,}801$ \\
    \midrule
    Text Recognition & TextVQA~\citep{singh2019textvqa} & Exact match & $5{,}000$ \\
     & OCRBench~\citep{liu2024ocrbench} & Accuracy & $1{,}000$ \\
    \midrule
    Chart Understanding & ChartQA~\citep{masry2022chartqa} & Relaxed accuracy & $2{,}500$ \\
     & CharXiv (descriptive)~\citep{wang2024charxiv} & GPT & $4{,}000$ \\
     & CharXiv (reasoning)~\citep{wang2024charxiv} & GPT & $1{,}000$ \\
    \midrule
    Science Knowledge & ScienceQA~\citep{lu2022scienceqa} & MCQ exact match & $4{,}241$ \\
     & AI2D~\citep{kembhavi2016ai2d} & MCQ exact match & $3{,}088$ \\
    \midrule
    Object Grounding & RefCOCO~\citep{kazemzadeh2014refcoco} & BLEU (region caption) & $1{,}975$ \\
     & RefCOCO+~\citep{kazemzadeh2014refcoco} & BLEU (region caption) & $1{,}975$ \\
     & RefCOCOg~\citep{mao2016generation,yu2016refcocog} & BLEU (region caption) & $5{,}023$ \\
    \midrule
    Spatial Reasoning & CV-Bench~\citep{cvbench2024} & Exact match & $2{,}638$ \\
     & SPARBench~\citep{zhang2025sparbench} & MRA & $2{,}842$ \\
    \midrule
    Image Understanding & SEEDBench~\citep{li2023seedbench} & MCQ exact match & $14{,}233$ \\
     & RealWorldQA~\citep{realworldqa2024} & Exact match & $765$ \\
     & LLaVA-W~\citep{liu2023llava} & GPT & \rev{$60$} \\
    \midrule
    Image Captioning & COCO-Cap~\citep{chen2015cococaptions} & BLEU & $5{,}000$ \\
     & NoCaps~\citep{agrawal2019nocaps} & BLEU & $4{,}500$ \\
     & TextCaps~\citep{sidorov2020textcaps} & BLEU & $3{,}166$ \\
    \midrule
    Hallucination Detection & POPE~\citep{li2023pope} & Accuracy & $9{,}000$ \\
     & HallusionBench~\citep{guan2024hallusionbench} & aAcc & $951$ \\
     & AMBER~\citep{wang2023amber} & Cover & $1{,}004$ \\
    \midrule
    Commonsense & OK-VQA~\citep{marino2019okvqa} & Exact match & $5{,}046$ \\
    \midrule
    Multi-Turn/Multilingual & MMT-Bench~\citep{ying2024mmtbench} & MCQ exact match & $3{,}127$ \\
     & MIA-Bench~\citep{qian2025miabench} & GPT & $400$ \\
    \midrule
    Web/UI Understanding & ScreenSpot~\citep{cheng2024seeclick} & Centre accuracy & $1{,}272$ \\
     & VisualWebBench~\citep{liu2024visualwebbench} & F1 & $314$ \\
     & WebSRC~\citep{chen2021websrc} & SQuAD F1 & $52{,}826$ \\
    \midrule
    Object Recognition & ERQA~\citep{teamgemini2025erqa} & Exact match & $400$ \\
    \midrule
    Instruction Following & LLaVA-Bench-COCO~\citep{liu2023llava} & GPT & $90$ \\
    \bottomrule
  \end{tabular}
\end{adjustbox}
  \end{revblock}
\end{table}

\paragraph{Asset licences.}
\label{app:licenses}
\begin{revblock}
All upstream benchmarks and model checkpoints are publicly released for research use, and we use them for evaluation and for diagnostic fine-tuning on ScienceQA (\cref{app:ckpt_ranking,app:d6_training_pruning}).
\end{revblock}

\FloatBarrier
\section{Stage 0: Data Cleaning}
\label{app:stage0}
\label{app:cleaning}

\begin{revblock}
Stage~0 runs on all \rev{$43$} benchmarks, with every item score scaled to $[0,1]$. Both filters run once, before any split; the all-correct filter uses the whole panel, and the blind-solvability filter uses the image-free runs described next. The \emph{all-correct} filter removes an item when every evaluated model with a valid score gets the maximum score. The \emph{blind-solvability} filter is inspired by MMStar~\citep{chen2024mmstar}: the models we also ran without images (Qwen-VL-Chat, Qwen2-VL, Qwen2.5-VL, Qwen3-VL 2B--8B, InternVL2--3.5 and Gemma3) answer each item from its text alone, and the item is removed when their mean score is at least $2/3$, a threshold not tuned on any ranking result. Benchmarks with too few valid image-free scores (COCO-Cap, MathVista, OlympiadBench, CharXiv, MM-Vet, MIA-Bench, LLaVA-Bench-COCO and LLaVA-W) go through the all-correct filter only.

All-wrong items are kept. Later models may solve them; with zero variance, Stage~2 rarely keeps them. \Cref{tab:cleaning} gives the per-benchmark counts. Stage~0 removes \rev{115{,}916} of the \rev{386{,}324} items (\rev{30.0\%}), and \rev{11{,}247} of them fail both filters.
\end{revblock}

\begin{table}[!htbp]
\centering
\small
\caption{\rev{\textbf{Stage~0 removes \rev{115{,}916} of the \rev{386{,}324} items of the $43$ benchmarks.} The all-correct filter uses every evaluated model, the blind-solvability filter the image-free runs. An item can fail both filters, so the removed count is the union. The fraction removed ranges from \rev{0.0\%} ($14$ benchmarks) to \rev{78.3\%} (\texttt{scienceqa}).}}
\label{tab:cleaning}
\setlength{\tabcolsep}{4pt}
\begin{revblock}
\begin{adjustbox}{max width=\linewidth}
\begin{tabular}{lrrrrr}
\toprule
Benchmark & Items & All-correct & Blind-solvable & Removed & \% \\
\midrule
\texttt{vqav2\_val} & 214{,}354 & 19{,}983 & 59{,}226 & 75{,}349 & 35.2 \\
\texttt{websrc\_val} & 52{,}826 & 7{,}825 & 3{,}552 & 10{,}537 & 19.9 \\
\texttt{seedbench} & 14{,}233 & 3{,}212 & 3{,}620 & 5{,}780 & 40.6 \\
\texttt{gqa} & 12{,}578 & 316 & 2{,}660 & 2{,}904 & 23.1 \\
\texttt{pope} & 9{,}000 & 5{,}304 & 4{,}500 & 6{,}831 & 75.9 \\
\texttt{docvqa\_val} & 5{,}349 & 13 & 107 & 119 & 2.2 \\
\texttt{ok\_vqa\_val2014} & 5{,}046 & 188 & 94 & 264 & 5.2 \\
\texttt{refcocog\_bbox\_test} & 5{,}023 & 0 & 0 & 0 & 0.0 \\
\texttt{textvqa\_val} & 5{,}000 & 912 & 151 & 1{,}028 & 20.6 \\
\texttt{coco2017\_cap\_val} & 5{,}000 & 0 & 0 & 0 & 0.0 \\
\texttt{nocaps\_val} & 4{,}500 & 0 & 0 & 0 & 0.0 \\
\texttt{mmbench\_en\_dev} & 4{,}329 & 1{,}320 & 993 & 1{,}983 & 45.8 \\
\texttt{scienceqa} & 4{,}241 & 1{,}200 & 3{,}244 & 3{,}321 & 78.3 \\
\texttt{charxiv\_val\_descriptive} & 4{,}000 & 117 & 0 & 117 & 2.9 \\
\texttt{textcaps\_val} & 3{,}166 & 0 & 0 & 0 & 0.0 \\
\texttt{mmt\_val} & 3{,}127 & 395 & 652 & 949 & 30.3 \\
\texttt{ai2d} & 3{,}088 & 322 & 1{,}635 & 1{,}672 & 54.1 \\
\texttt{sparbench} & 2{,}842 & 0 & 33 & 33 & 1.2 \\
\texttt{infovqa\_val} & 2{,}801 & 0 & 269 & 269 & 9.6 \\
\texttt{cv\_bench} & 2{,}638 & 225 & 883 & 1{,}039 & 39.4 \\
\texttt{chartqa} & 2{,}500 & 496 & 208 & 584 & 23.4 \\
\texttt{mme} & 2{,}374 & 473 & 1{,}012 & 1{,}201 & 50.6 \\
\texttt{refcoco\_bbox\_testA} & 1{,}975 & 0 & 0 & 0 & 0.0 \\
\texttt{refcoco+\_bbox\_testA} & 1{,}975 & 0 & 0 & 0 & 0.0 \\
\texttt{blink} & 1{,}901 & 44 & 587 & 614 & 32.3 \\
\texttt{mmmu\_pro\_standard} & 1{,}730 & 22 & 162 & 172 & 9.9 \\
\texttt{screenspot\_rec\_test} & 1{,}272 & 0 & 0 & 0 & 0.0 \\
\texttt{amber\_g} & 1{,}004 & 0 & 0 & 0 & 0.0 \\
\texttt{mathvista\_testmini\_cot} & 1{,}000 & 40 & 0 & 40 & 4.0 \\
\texttt{ocrbench} & 1{,}000 & 227 & 12 & 234 & 23.4 \\
\texttt{charxiv\_val\_reasoning} & 1{,}000 & 0 & 0 & 0 & 0.0 \\
\texttt{hallusion\_bench\_image} & 951 & 45 & 161 & 200 & 21.0 \\
\texttt{mmmu\_val} & 900 & 30 & 264 & 268 & 29.8 \\
\texttt{cmmmu\_val} & 900 & 9 & 152 & 153 & 17.0 \\
\texttt{realworldqa} & 765 & 23 & 176 & 190 & 24.8 \\
\texttt{mia\_bench} & 400 & 0 & 0 & 0 & 0.0 \\
\texttt{erqa} & 400 & 6 & 56 & 58 & 14.5 \\
\texttt{visualwebbench\_webqa} & 314 & 0 & 6 & 6 & 1.9 \\
\texttt{mathvision\_testmini} & 304 & 0 & 1 & 1 & 0.3 \\
\texttt{mmvet} & 218 & 0 & 0 & 0 & 0.0 \\
\texttt{olympiadbench\_OE\_MM\_maths\_en\_COMP} & 150 & 0 & 0 & 0 & 0.0 \\
\texttt{llava\_bench\_coco} & 90 & 0 & 0 & 0 & 0.0 \\
\texttt{llava\_in\_the\_wild} & 60 & 0 & 0 & 0 & 0.0 \\
\midrule
Total & 386{,}324 & 42{,}747 & 84{,}416 & 115{,}916 & 30.0 \\
\bottomrule
\end{tabular}
\end{adjustbox}
\end{revblock}
\end{table}

\FloatBarrier
\section{Stage 1: Category Representative Selection}
\label{app:stage1}

\subsection{Category representative selection}
\label{app:category_rep}

\begin{revblock}
The $18$ categories follow the capability groupings used in recent frontier VLM technical reports~\citep{qwen3vl2025,internvl3_5_2025,glm4_1v_2025}. Each benchmark is assigned to the category of its primary capability (\cref{tab:benchmark_details}), and the mapping is fixed before any selection. For each of the \rev{$18$} categories, Stage~1 picks the benchmark whose per-model score correlates best with the category mean of $z$-normalised scores (\cref{eq:rep}), computed on the cleaned pool that Stage~2 consumes. \Cref{tab:category_rep} lists the representatives and \cref{tab:all_task_rho} every candidate. The representative correlations have median \rev{0.925}; they fall below $0.90$ in \rev{eight} categories, lowest in Hallucination Detection (\rev{0.699}) and Visual Reasoning (\rev{0.758}), where no benchmark tracks the category mean closely.
\end{revblock}

\begin{table}[!htbp]
\centering
\small
\caption{\rev{\textbf{Stage~1 representatives of the \rev{$18$} capability categories.} The representative maximises the Spearman correlation between its per-model score and the category mean of $z$-normalised scores (\cref{eq:rep}); the last column gives the correlation across models after pooling the same-category supplements (\cref{app:multisource}). Per-candidate correlations are in \cref{tab:all_task_rho}.}}
\label{tab:category_rep}
\setlength{\tabcolsep}{4pt}
\begin{revblock}
\begin{adjustbox}{max width=\linewidth}
\begin{tabular}{l l c c}
\toprule
Category & Representative & Stage~1 $\rho$ & Supplemented $\rho$ \\
\midrule
Academic Knowledge & \texttt{mmmu\_val} & 0.941 & 0.951 \\
Chart Understanding & \texttt{charxiv\_val\_descriptive} & 0.858 & 0.951 \\
Commonsense & \texttt{ok\_vqa\_val2014} & 1.000 & 1.000 \\
Comprehensive & \texttt{mmbench\_en\_dev} & 0.875 & 0.942 \\
Document Understanding & \texttt{infovqa\_val} & 0.992 & 0.992 \\
Hallucination Detection & \texttt{amber\_g} & 0.699 & 0.947 \\
Image Captioning & \texttt{nocaps\_val} & 0.980 & 0.980 \\
Image Understanding & \texttt{seedbench} & 0.910 & 0.933 \\
Instruction Following & \texttt{llava\_bench\_coco} & 1.000 & 1.000 \\
Math Reasoning & \texttt{mathvision\_testmini} & 0.898 & 0.950 \\
Multi-Turn/Multilingual & \texttt{mmt\_val} & 0.863 & 0.924 \\
Object Grounding & \texttt{refcoco+\_bbox\_testA} & 0.950 & 0.953 \\
Object Recognition & \texttt{erqa} & 1.000 & 1.000 \\
Science Knowledge & \texttt{scienceqa} & 0.945 & 0.950 \\
Spatial Reasoning & \texttt{cv\_bench} & 0.860 & 0.951 \\
Text Recognition & \texttt{textvqa\_val} & 0.958 & 0.958 \\
Visual Reasoning & \texttt{vqav2\_val} & 0.758 & 0.764 \\
Web/UI Understanding & \texttt{visualwebbench\_webqa} & 0.794 & 0.800 \\
\bottomrule
\end{tabular}
\end{adjustbox}
\end{revblock}
\end{table}

\begin{table}[!htbp]
\centering
\small
\caption{\rev{\textbf{Per-candidate Stage~1 correlations in the multi-benchmark categories.} Spearman $\rho$ between each candidate's per-model score and the category mean of $z$-normalised scores; the representative is in bold. MM-Vet is scored on $29$ models because one model's score is missing.}}
\label{tab:all_task_rho}
\setlength{\tabcolsep}{4pt}
\begin{revblock}
\begin{adjustbox}{max width=\linewidth}
\begin{tabular}{l l c}
\toprule
Category & Benchmark & $\rho$ \\
\midrule
Academic Knowledge & \textbf{\texttt{mmmu\_val}} & 0.941 \\
 & \texttt{mmmu\_pro\_standard} & 0.933 \\
 & \texttt{cmmmu\_val} & 0.908 \\
\midrule
Chart Understanding & \texttt{chartqa} & 0.668 \\
 & \textbf{\texttt{charxiv\_val\_descriptive}} & 0.858 \\
 & \texttt{charxiv\_val\_reasoning} & 0.840 \\
\midrule
Comprehensive & \texttt{mme} & 0.540 \\
 & \textbf{\texttt{mmbench\_en\_dev}} & 0.875 \\
 & \texttt{mmvet} & 0.834 \\
\midrule
Document Understanding & \texttt{docvqa\_val} & 0.975 \\
 & \textbf{\texttt{infovqa\_val}} & 0.992 \\
\midrule
Hallucination Detection & \texttt{pope} & 0.683 \\
 & \texttt{hallusion\_bench\_image} & 0.510 \\
 & \textbf{\texttt{amber\_g}} & 0.699 \\
\midrule
Image Captioning & \texttt{coco2017\_cap\_val} & 0.975 \\
 & \textbf{\texttt{nocaps\_val}} & 0.980 \\
 & \texttt{textcaps\_val} & 0.917 \\
\midrule
Image Understanding & \textbf{\texttt{seedbench}} & 0.910 \\
 & \texttt{realworldqa} & 0.858 \\
 & \texttt{llava\_in\_the\_wild} & 0.752 \\
\midrule
Math Reasoning & \texttt{mathvista\_testmini\_cot} & 0.807 \\
 & \textbf{\texttt{mathvision\_testmini}} & 0.898 \\
 & \texttt{olympiadbench\_OE\_MM\_maths\_en\_COMP} & 0.848 \\
\midrule
Multi-Turn/Multilingual & \textbf{\texttt{mmt\_val}} & 0.863 \\
 & \texttt{mia\_bench} & 0.762 \\
\midrule
Object Grounding & \texttt{refcoco\_bbox\_testA} & 0.945 \\
 & \textbf{\texttt{refcoco+\_bbox\_testA}} & 0.950 \\
 & \texttt{refcocog\_bbox\_test} & 0.860 \\
\midrule
Science Knowledge & \textbf{\texttt{scienceqa}} & 0.945 \\
 & \texttt{ai2d} & 0.916 \\
\midrule
Spatial Reasoning & \textbf{\texttt{cv\_bench}} & 0.860 \\
 & \texttt{sparbench} & 0.680 \\
\midrule
Text Recognition & \textbf{\texttt{textvqa\_val}} & 0.958 \\
 & \texttt{ocrbench} & 0.867 \\
\midrule
Visual Reasoning & \textbf{\texttt{vqav2\_val}} & 0.758 \\
 & \texttt{gqa} & 0.687 \\
 & \texttt{blink} & 0.562 \\
\midrule
Web/UI Understanding & \texttt{screenspot\_rec\_test} & 0.534 \\
 & \textbf{\texttt{visualwebbench\_webqa}} & 0.794 \\
 & \texttt{websrc\_val} & 0.781 \\
\bottomrule
\end{tabular}
\end{adjustbox}
\end{revblock}
\end{table}

\subsection{Multi-source item supplementation for low-$\rho$ categories}
\label{app:multisource}

\begin{revblock}
When the representative alone has $\rho<0.95$, Stage~1 adds items from the other benchmarks of the same category, starting from the representative's full cleaned item set. Supplements are added in descending variance of their scaled scores, with ties broken by benchmark order and item index, and after each addition we recompute the Spearman correlation between the item-pooled score of the supplemented pool and the category mean. Supplementation stops at the first $\rho \ge 0.95$, or at the maximum $\rho$ when the target is not reached. Supplement items come from the Stage~0-cleaned pool, like the representatives. This pooled correlation (the last column of \cref{tab:category_rep}) can differ from the Stage~1 correlation of the representative, because it pools items of benchmarks with different metrics and sizes.

\Cref{tab:augmentation_thresholds} shows three regimes. \rev{Six} categories need no supplement, \rev{six} reach $\rho \ge 0.95$ with \rev{2}--\rev{1{,}287} added items, and \rev{six} stay below the target: \rev{Comprehensive (0.942)}, Hallucination Detection (\rev{0.947}), Image Understanding (\rev{0.933}), \rev{Multi-Turn/Multilingual (0.924)}, Visual Reasoning (\rev{0.764}) and \rev{Web/UI Understanding (0.800)}. The supplements become part of the category pool, so the categories with large supplements also carry large pools after pruning.
\end{revblock}

\begin{table}[!htbp]
\centering
\small
\caption{\rev{\textbf{\rev{Twelve} of the \rev{$18$} categories reach $\rho\ge 0.95$, \rev{six of them through supplementation}; \rev{six} stay below it even at the best prefix of the supplements.} Tier~1 needs no supplement, tier~2 reaches the target, tier~3 does not. Items are added from the other benchmarks of the category in descending variance of their scaled scores, and $\rho$ is the correlation across models between the item-pooled score and the category mean. The supplements enter the category pool. Object Grounding is supplemented because its representative has $\rho{=}0.9497$ before rounding ($0.950$ in \cref{tab:category_rep}).}}
\label{tab:augmentation_thresholds}
\setlength{\tabcolsep}{3pt}
\begin{revblock}
\begin{adjustbox}{max width=\linewidth}
\begin{tabular}{l l c r r r c}
\toprule
Category & Representative & Tier & Rep.\ items & Added & Available & Final $\rho$ \\
\midrule
Commonsense & \texttt{ok\_vqa\_val2014} & 1 & 4{,}782 & 0 & 0 & 1.000 \\
Document Understanding & \texttt{infovqa\_val} & 1 & 2{,}532 & 0 & 5{,}230 & 0.992 \\
Image Captioning & \texttt{nocaps\_val} & 1 & 4{,}500 & 0 & 8{,}166 & 0.980 \\
Instruction Following & \texttt{llava\_bench\_coco} & 1 & 90 & 0 & 0 & 1.000 \\
Object Recognition & \texttt{erqa} & 1 & 342 & 0 & 0 & 1.000 \\
Text Recognition & \texttt{textvqa\_val} & 1 & 3{,}972 & 0 & 766 & 0.958 \\
Object Grounding & \texttt{refcoco+\_bbox\_testA} & 2 & 1{,}975 & 2 & 6{,}998 & 0.953 \\
Academic Knowledge & \texttt{mmmu\_val} & 2 & 632 & 5 & 2{,}305 & 0.951 \\
Math Reasoning & \texttt{mathvision\_testmini} & 2 & 303 & 66 & 1{,}110 & 0.950 \\
Science Knowledge & \texttt{scienceqa} & 2 & 920 & 93 & 1{,}416 & 0.950 \\
Spatial Reasoning & \texttt{cv\_bench} & 2 & 1{,}599 & 673 & 2{,}809 & 0.951 \\
Chart Understanding & \texttt{charxiv\_val\_descriptive} & 2 & 3{,}883 & 1{,}287 & 2{,}916 & 0.951 \\
Image Understanding & \texttt{seedbench} & 3 & 8{,}453 & 293 & 635 & 0.933 \\
Multi-Turn/Multilingual & \texttt{mmt\_val} & 3 & 2{,}178 & 393 & 400 & 0.924 \\
Hallucination Detection & \texttt{amber\_g} & 3 & 1{,}004 & 709 & 2{,}920 & 0.947 \\
Comprehensive & \texttt{mmbench\_en\_dev} & 3 & 2{,}346 & 961 & 1{,}391 & 0.942 \\
Visual Reasoning & \texttt{vqav2\_val} & 3 & 139{,}005 & 1{,}656 & 10{,}961 & 0.764 \\
Web/UI Understanding & \texttt{visualwebbench\_webqa} & 3 & 308 & 37{,}414 & 43{,}561 & 0.800 \\
\bottomrule
\end{tabular}
\end{adjustbox}
\end{revblock}
\end{table}

\FloatBarrier
\section{Stage 2: Item Pruning with VAW}
\label{app:stage2}\label{app:perbench}

\subsection{\rev{Selection rule}}
\label{app:stage2_rule}

\begin{revblock}
Item scores are scaled linearly to $[0,1]$, and an invalid record counts as missing, not as zero. IRT 2PL \rev{needs} binary responses, so \rev{it receives} the scores binarised, VQA-style soft and judge scores at half their scale, and captioning, referring-expression and the remaining graded scores at the benchmark median; every other selector and every ranking target use the scaled scores, so DatBench's $r_{pb}$ becomes the Pearson item--rest correlation on continuous scores, which equals the point-biserial correlation on binary ones.

Within benchmark $b$, the cells are drawn from the vision-text embeddings $E_i^{\mathrm{VL}}$ of Qwen3-VL-Embedding-8B~\citep{qwen3vl_embedding_2026}. They use item content only, so they stay fixed across panels and model subsamples. The $64$-dimensional projection only speeds up $k$-means on the $4{,}096$-dimensional embeddings of \rev{222{,}149} items.

Items without a vision-text embedding enter no cell and have $\delta_i^{\mathrm{VL}}=0$; a share that the cells cannot fill is completed with the benchmark's highest-scoring remaining items.
\end{revblock}

\begin{table}[!htbp]
\centering
\small
\caption{\rev{\textbf{VAW keeps $\rho\ge 0.90$ on unseen models in \rev{thirteen} of \rev{$18$} categories at $r{=}5\%$; the weakest are \rev{Spatial Reasoning, Academic Knowledge and Object Recognition}.} Per-category Spearman $\rho$ on the held-out models, mean over the $20$ random splits (\cref{app:unseen_protocol}). Pool = cleaned representative plus supplements; $K=\rev{\max(\lfloor rN\rfloor,\min(N,100))}$. The macro $\rho$ values are \rev{0.925} (VAW) and \rev{0.906} (Variance, seeded random tie-breaking). The higher of the two selectors is in bold.}}
\label{tab:perbench_16vlm}
\setlength{\tabcolsep}{3pt}
\begin{revblock}
\begin{adjustbox}{max width=\linewidth}
\begin{tabular}{l l r r c c}
\toprule
Category & Representative & Pool & $K$ & VAW & Variance \\
\midrule
Image Captioning & \texttt{nocaps\_val} & 4{,}500 & 225 & \textbf{0.984} & \textbf{0.984} \\
Visual Reasoning & \texttt{vqav2\_val} & 140{,}661 & 7{,}033 & \textbf{0.979} & 0.952 \\
Commonsense & \texttt{ok\_vqa\_val2014} & 4{,}782 & 239 & \textbf{0.970} & 0.968 \\
Document Understanding & \texttt{infovqa\_val} & 2{,}532 & 126 & \textbf{0.966} & 0.955 \\
Object Grounding & \texttt{refcoco+\_bbox\_testA} & 1{,}977 & 100 & 0.942 & \textbf{0.949} \\
Text Recognition & \texttt{textvqa\_val} & 3{,}972 & 198 & \textbf{0.948} & 0.935 \\
Chart Understanding & \texttt{charxiv\_val\_descriptive} & 5{,}170 & 258 & \textbf{0.936} & 0.906 \\
Web/UI Understanding & \texttt{visualwebbench\_webqa} & 37{,}722 & 1{,}886 & \textbf{0.936} & 0.852 \\
Comprehensive & \texttt{mmbench\_en\_dev} & 3{,}307 & 165 & \textbf{0.928} & 0.883 \\
Multi-Turn/Multilingual & \texttt{mmt\_val} & 2{,}571 & 128 & \textbf{0.907} & 0.902 \\
Science Knowledge & \texttt{scienceqa} & 1{,}013 & 100 & \textbf{0.930} & 0.907 \\
Hallucination Detection & \texttt{amber\_g} & 1{,}713 & 100 & \textbf{0.887} & 0.748 \\
Image Understanding & \texttt{seedbench} & 8{,}746 & 437 & 0.884 & \textbf{0.897} \\
Spatial Reasoning & \texttt{cv\_bench} & 2{,}272 & 113 & 0.801 & \textbf{0.837} \\
Math Reasoning & \texttt{mathvision\_testmini} & 369 & 100 & \textbf{0.943} & 0.905 \\
Academic Knowledge & \texttt{mmmu\_val} & 637 & 100 & 0.848 & \textbf{0.873} \\
Instruction Following & \texttt{llava\_bench\_coco} & 90 & 90 & \textbf{1.000} & \textbf{1.000} \\
Object Recognition & \texttt{erqa} & 342 & 100 & \textbf{0.858} & 0.855 \\
\bottomrule
\end{tabular}
\end{adjustbox}
\end{revblock}
\end{table}

\subsection{Per-category variation}
\label{app:per_benchmark_variation_detail}

\begin{revblock}
On unseen models VAW keeps $\rho\ge 0.90$ in \rev{thirteen} of the \rev{$18$} categories at $r{=}5\%$ (\cref{tab:perbench_16vlm}). The weakest are \rev{Spatial Reasoning (0.801, 113 kept items), Academic Knowledge (0.848, 100 items)} and Object Recognition (\rev{0.858}, \rev{100} items). \rev{Academic Knowledge and Object Recognition are small pools ($637$ and $342$ items) that keep only the $100$-item floor, and Spatial Reasoning keeps $113$ items, so a small set of items carries each of these categories.} Users who need one of these categories at fine resolution should evaluate the full benchmark or use a larger retention $r$.
\end{revblock}

\subsection{\rev{Unseen-model protocols}}
\label{app:unseen_protocol}

\begin{revblock}
A split divides the panel into build models, which select the items, and held-out models, which are scored. Selection reads only the build columns. Stages~0 and~1 use the responses of the whole panel and are run once, so the pools are fixed before any split; the cells use no model responses, and every variance, IRT or AnchorPoints clustering fit uses the build models alone. The protocol therefore tests Stage~2 selection on unseen models, not the construction of the pools; the end-to-end protocol below refits Stages~1 to~3 on the build models. For each category we compute the Spearman correlation over the held-out models between their mean on the full pool and their mean on the selected items, and average over the categories with a finite value. A category whose correlation is undefined in a split, because every held-out model receives the same score on the selected items, is left out of that split's mean. Of the six retentions, the tables report $r\in\{2,5,10\}\%$.

\begin{table}[!t]
\centering
\begin{revblock}
\caption{\textbf{On unseen models, \rev{VAW has} the highest mean at $r{=}5\%$ and $10\%$, and \rev{AnchorPoints} at $r{=}2\%$.}
Selection uses half of the panel and $\rho$ is measured on the other half, averaged over the $18$ categories (all with a defined correlation) and $20$ random splits; the last column gives $\rho$ on the chronological split at $r{=}5\%$. Best in bold, second best underlined.}
\label{tab:stage2_unseen}
\small
\setlength{\tabcolsep}{5pt}
\begin{adjustbox}{max width=\linewidth}
\begin{tabular}{l ccc c}
\toprule
Selector & $r{=}2\%$ & $r{=}5\%$ & $r{=}10\%$ & Chronological \\
\midrule
\textbf{VAW (ours)} & \underline{0.908} & \textbf{0.925} & \textbf{0.944} & \textbf{0.936} \\
\midrule
Variance & 0.890 & 0.906 & 0.927 & 0.889 \\
IRT 2PL & 0.901 & 0.909 & 0.921 & 0.915 \\
AnchorPoints & \textbf{0.911} & \underline{0.920} & \underline{0.932} & \underline{0.929} \\
DatBench $r_{pb}$ & 0.882 & 0.894 & 0.908 & 0.854 \\
DISCO & 0.752 & 0.781 & 0.832 & 0.780 \\
\bottomrule
\end{tabular}
\end{adjustbox}
\end{revblock}
\end{table}

The random-split protocol draws $20$ random $15/15$ splits with a fixed seed and reports the mean (\cref{tab:stage2_unseen}). The chronological protocol orders the models by public release date, builds on the $15$ oldest and scores the $15$ newest; the $10/20$ and $20/10$ splits are sensitivity checks (\rev{VAW $0.917$ and $0.938$}). Models without a verifiable release date belong to the newest release waves and sort last, and no undated model enters a build set. Each correlation needs at least $10$ held-out models with finite scores. The per-split distribution is in \cref{tab:app_unseen_splits}.

Variance uses seeded random tie-breaking, as in the main comparison. At $r{=}5\%$ VAW reaches a mean of \rev{0.925} against $0.920$ for AnchorPoints, \rev{0.909} for IRT 2PL and \rev{0.906} for Variance. At $r{=}2\%$ \rev{AnchorPoints has the highest mean (0.911 against 0.908 for VAW).} At $r{=}5\%$ VAW is the best in $10$ of the $20$ splits (\cref{tab:app_unseen_splits}); with $15$ held-out models the split-to-split sd (\rev{0.01}--\rev{0.03}) is as large as the gaps between selectors. A single chronological split is subject to this variability as well, so we report it beside the random-split mean.
\end{revblock}

\begin{table}[!htbp]
\centering
\small
\caption{\rev{\textbf{On unseen models VAW has the highest mean at $r{=}5\%$ and $10\%$ and \rev{AnchorPoints} at $r{=}2\%$, but single splits overlap widely.} Macro Spearman $\rho$ (mean over categories) on the held-out models. Left: $20$ random $15/15$ splits; sd, min and max over splits at $r{=}5\%$, and the number of splits in which the selector is the best of the $6$ listed. Right: the chronological split, built on the $15$ oldest models and scored on the $15$ newest, at $r{=}5\%$. The split sd is \rev{$0.01$--$0.03$}, so the ranking rests on the means, not on single splits. Best in bold, second best underlined.}}
\label{tab:app_unseen_splits}
\footnotesize
\setlength{\tabcolsep}{3pt}
\begin{revblock}
\begin{adjustbox}{max width=\linewidth}
\begin{tabular}{l ccccccc c}
\toprule
 & \multicolumn{7}{c}{Random $15/15$ splits} & Chronological \\
\cmidrule(lr){2-8}\cmidrule(lr){9-9}
Selector & $2\%$ & $5\%$ & sd & min & max & best & $10\%$ & $5\%$ \\
\midrule
VAW & \underline{0.908} & \textbf{0.925} & 0.018 & 0.867 & 0.949 & \textbf{10} & \textbf{0.944} & \textbf{0.936} \\
IRT 2PL & 0.901 & 0.909 & 0.021 & 0.849 & 0.930 & 4 & 0.921 & 0.915 \\
AnchorPoints & \textbf{0.911} & \underline{0.920} & 0.012 & 0.891 & 0.942 & \underline{5} & \underline{0.932} & \underline{0.929} \\
Variance & 0.890 & 0.906 & 0.016 & 0.858 & 0.933 & 1 & 0.927 & 0.889 \\
DatBench $r_{pb}$ & 0.882 & 0.894 & 0.021 & 0.853 & 0.929 & 0 & 0.908 & 0.854 \\
DISCO & 0.752 & 0.781 & 0.033 & 0.722 & 0.840 & 0 & 0.832 & 0.780 \\
\bottomrule
\end{tabular}
\end{adjustbox}
\end{revblock}
\end{table}

\paragraph{\rev{End-to-end comparison.}}
\label{app:end2end}
\begin{revblock}
Stage~0 is a data-cleaning step applied once on all models: an item that every model answers correctly adds the same score to every model and cannot change a ranking, and a blind-solvable item does not belong to a vision--language ranking. For each split, Stage~1 is refitted on the build models (representatives and supplements from their scores), every Stage~2 selector selects items at retention $r$ on the build models, Stage~3 searches all $\binom{\rev{18}}{5}$ category subsets on the build columns of the pruned matrix, and the held-out models are scored. The end-to-end $\rho$ compares the held-out ranking by the category means over all Stage~0-cleaned benchmarks with the ranking by the pruned $k{=}5$ leaderboard. This reference replaces the benchmark mean $\bar{s}_m$ of \cref{sec:problem} by the mean of the category means, which weights each category equally. A second score keeps all pruned categories and omits Stage~3. On the $20$ random splits (\cref{tab:app_end2end}), Stage~3 only, without Stage~2 pruning, reaches \rev{0.901}. At $r{=}5\%$ every selector lies between \rev{0.870} and $0.893$, with AnchorPoints highest, Variance at \rev{0.889} and VAW at \rev{0.892}, and each selector beats Stage~3 only in \rev{6} to \rev{8} of the $20$ splits. The split-to-split sd of the end-to-end $\rho$ is $0.05$--\rev{0.09}, so the selectors overlap substantially, because Stage~3 chooses $5$ of \rev{$18$} categories from $15$ build models and the chosen set changes from split to split. At $r{=}5\%$, VAW is no more than $0.01$ below every alternative\rev{, including Stage~3 only (VAW 0.892 against 0.901 for Stage~3 only), though not ahead of all of them}. At $r{=}2\%$ and $10\%$ it is $0.04$ and $0.05$ below the best selector. Without Stage~3 the \rev{$18$} pruned categories preserve the ranking closely (VAW \rev{0.956}, IRT 2PL $0.958$).
\end{revblock}

\begin{table}[!htbp]
\centering
\small
\caption{\rev{\textbf{End to end, the Stage~2 selectors and Stage~3 alone differ little relative to split-to-split variability.} Spearman $\rho$ over held-out models between the ranking by the category means over all Stage~0-cleaned benchmarks and the $k{=}5$ leaderboard after Stage~2 at retention $r$ and Stage~3 on the build models; $20$ random splits. \emph{No Stage~3} scores all pruned categories; the next column counts splits in which the selector beats Stage~3 without Stage~2 pruning (Stage~3 only). The split-to-split sd of the end-to-end $\rho$ is \rev{$0.05$--$0.09$, as large as the gaps between selectors}. Variance here breaks ties by item index. Best in bold, second best underlined.}}
\label{tab:app_end2end}
\setlength{\tabcolsep}{4pt}
\begin{revblock}
\begin{adjustbox}{max width=\linewidth}
\begin{tabular}{l ccc cc}
\toprule
 & \multicolumn{3}{c}{End-to-end} & \multicolumn{2}{c}{$r{=}5\%$} \\
\cmidrule(lr){2-4}\cmidrule(lr){5-6}
Stage~2 selector & $2\%$ & $5\%$ & $10\%$ & No Stage~3 & $>$Stage~3 only \\
\midrule
Stage~3 only & \underline{0.901} & \textbf{0.901} & 0.901 & -- & -- \\
\midrule
VAW & 0.869 & 0.892 & 0.866 & \underline{0.956} & 6/20 \\
IRT 2PL & \textbf{0.907} & 0.890 & \underline{0.909} & \textbf{0.958} & 7/20 \\
AnchorPoints & 0.899 & \underline{0.893} & 0.898 & 0.950 & 8/20 \\
DatBench $r_{pb}$ & 0.870 & 0.870 & 0.883 & 0.944 & 8/20 \\
Variance & 0.873 & 0.889 & \textbf{0.917} & 0.947 & 7/20 \\
DISCO & 0.846 & 0.871 & 0.889 & 0.949 & 6/20 \\
\bottomrule
\end{tabular}
\end{adjustbox}
\end{revblock}
\end{table}

\subsection{\rev{Cell-space ablation}}
\label{app:ablation_cells}\label{app:visual_gain}\label{app:vlm_analysis}

\begin{revblock}
The ablation keeps the score, the benchmark shares and the number of cells, and varies the space in which the cells are drawn: the vision-text embedding (the default), the text-only embedding of the same encoder under the same projection, or a random partition of each benchmark's items into the same number of groups. Two further arms remove the cells and keep only the benchmark shares, or replace the cells with a DPP \rev{(determinantal point process)}. Inside each benchmark share it runs the greedy MAP \rev{(maximum a posteriori)} selection of $L=\mathrm{diag}(q)\,G\,\mathrm{diag}(q)$ with an RBF \rev{(radial basis function)} kernel $G$ on the projected embeddings, whose bandwidth is the median squared distance of $2{,}000$ random item pairs, so no weight is tuned. All arms run on the $20$ random splits of \cref{app:unseen_protocol} (\cref{tab:app_ablation_cells}).

With benchmark shares alone VAW reaches \rev{0.908} at $r{=}5\%$, level with Variance (\rev{0.906}), and vision-text cells add \rev{0.017} (\rev{18} of \rev{20} splits). Vision-text cells are ahead of random partitions of the same size by \rev{0.011} on average and in \rev{16} of \rev{20} splits, so the gain depends on grouping similar items together. Text-only cells are within about $0.005$ of vision-text cells (vision-text minus text-only mean difference \rev{$-0.005$}, positive in \rev{6} of \rev{20} splits), so the coverage gain does not require image inputs to build the cells. The DPP variant reaches \rev{0.919} against \rev{0.925} for the cells; both use the embeddings, but the DPP also needs a kernel and a bandwidth, while the number of cells is fixed by the budget, so the pipeline uses the cells. At $r{=}10\%$ the three cell spaces are within about \rev{0.007}, \rev{with text-only cells highest and the random partition lowest}.

Inside vision-text cells the vision-dependence score $\delta^{\mathrm{VL}}$ changes the macro $\rho$ by \rev{+0.002} at $r{=}2\%$ (\rev{9} of \rev{20} splits), \rev{+0.003} at $r{=}5\%$ and $-0.002$ at $r{=}10\%$. These differences, computed before rounding, are descriptive; \cref{app:finite_m_v2} reports a scan across panel sizes. The differences between cell spaces and against the DPP are at most \rev{0.019}, against split-to-split sds of \rev{0.010}--\rev{0.018} for these arms.
\end{revblock}

\begin{table}[!htbp]
\centering
\small
\caption{\rev{\textbf{At $r{=}5\%$ embedding cells beat random partitions with either the vision-text or the text-only embedding.} Macro Spearman $\rho$ (mean over categories) on held-out models, mean over the $20$ random splits of \cref{app:unseen_protocol}. $-\delta$: $\alpha{=}0$, without the vision-dependence score. Best in bold, second best underlined.}}
\label{tab:app_ablation_cells}
\setlength{\tabcolsep}{4pt}
\begin{revblock}
\begin{adjustbox}{max width=\linewidth}
\begin{tabular}{l ccc}
\toprule
Selector & $2\%$ & $5\%$ & $10\%$ \\
\midrule
VAW, vision-text cells & 0.908 & 0.925 & \underline{0.944} \\
VAW, text-only cells & \textbf{0.914} & \textbf{0.930} & \textbf{0.946} \\
VAW, random partition & 0.895 & 0.914 & 0.939 \\
\midrule
VAW$-\delta$, vision-text cells & 0.906 & 0.921 & \textbf{0.946} \\
VAW$-\delta$, text-only cells & \underline{0.910} & \underline{0.929} & \underline{0.944} \\
VAW$-\delta$, random partition & 0.895 & 0.914 & 0.937 \\
\midrule
DPP, vision-text & 0.897 & 0.919 & 0.942 \\
DPP, text-only & 0.900 & 0.924 & 0.943 \\
DPP$-\delta$, vision-text & 0.898 & 0.921 & 0.943 \\
\midrule
\rev{VAW, shares only} & 0.884 & 0.908 & 0.930 \\
\rev{Variance, shares only} & 0.893 & 0.908 & 0.932 \\
\midrule
Variance & 0.890 & 0.906 & 0.927 \\
\bottomrule
\end{tabular}
\end{adjustbox}
\end{revblock}
\end{table}

\paragraph{\rev{Encoder of the vision-dependence score.}}
\label{app:encoder}
\begin{revblock}
We computed $\delta^{\mathrm{VL}}$ and the cells with three other encoders (Qwen3-VL-Embedding-2B, VLM2Vec-V2~\citep{meng2025vlm2vecv2} and Gemini-Embedding-2~\citep{shanbhogue2026gemini}) and ran VAW with each on the held-out models of the $20$ random splits (\cref{tab:app_encoder}). At $r{=}5\%$ every other encoder is $0.011$--$0.012$ below the default, so the choice of encoder moves the result by about $0.01$; at $r{=}2\%$ and $10\%$ every other encoder is above the default.
\end{revblock}

\begin{table}[!htbp]
\centering
\small
\caption{\textbf{Swapping the encoder lowers VAW by $0.011$--$0.012$ at $r{=}5\%$.} Macro Spearman $\rho$ (mean over categories) on held-out models, mean over the $20$ random $15/15$ splits, on the four categories that every encoder covers (Comprehensive, Document Understanding, Text Recognition, Science Knowledge). Each encoder supplies both $\delta^{\mathrm{VL}}$ and the cells; $\Delta\rho$ is against the default encoder at $r{=}5\%$. Variance (item-index tie-breaking) is a reference on the same splits and categories. Best in bold, second best underlined.}
\label{tab:app_encoder}
\setlength{\tabcolsep}{6pt}
\begin{tabular}{lcccc}
\toprule
Encoder & $r{=}2\%$ & $r{=}5\%$ & $r{=}10\%$ & $\Delta\rho$ \\
\midrule
Qwen3-VL-Embedding-8B (default) & 0.907 & \textbf{0.943} & 0.949 & -- \\
Qwen3-VL-Embedding-2B & \underline{0.911} & 0.931 & 0.950 & $-0.012$ \\
VLM2Vec-V2 & \textbf{0.923} & \underline{0.932} & \textbf{0.956} & $-0.011$ \\
Gemini-Embedding-2 & 0.909 & \underline{0.932} & \underline{0.954} & $-0.011$ \\
\midrule
Variance & 0.897 & 0.914 & 0.940 & $-0.029$ \\
\bottomrule
\end{tabular}
\end{table}

\FloatBarrier
\section{Stage 3: Category-Count Pruning}
\label{app:ksweep}\label{app:stage3}

\begin{revblock}
Stage~3 enumerates all $\binom{\rev{18}}{k}$ category subsets (\cref{eq:stage3}) on the Stage~2 output at $r{=}5\%$; for $k{=}5$ it enumerates \rev{$8{,}568$} subsets. Exact ties in $\rho$ are broken by taking the first subset in alphabetical category order. We report two correlations for each $k$ (\cref{tab:ksweep_exhaustive}). The \emph{Stage~3 convention} compares the ranking by the $k$ selected pruned categories with the ranking by all \rev{$18$} pruned categories, which is the quantity the search maximises. The \emph{end-to-end} correlation compares it with the ranking by \rev{the category means over all Stage~0-cleaned benchmarks}, and is the one reported for the released suite. The released suite uses $k{=}5$, chosen from the spectrum of the category score matrix (\cref{fig:low_rank_spectrum}); Stage~3 maximises ranking fidelity at that $k$ and does not minimise the number of items. The selected subsets are not nested across $k$, and the category pools differ widely in size (Visual Reasoning alone keeps $7{,}033$ items), so the item count is not monotone in $k$: the $k{=}8$ subset omits Visual Reasoning and keeps $1{,}136$ items at an end-to-end $\rho$ of $0.962$ on the full panel. A user who needs fewer items can take this subset from \cref{tab:ksweep_exhaustive}. The released composition is in \cref{tab:release_composition}.
\end{revblock}

\begin{table}[!htbp]
\centering
\small
\caption{\rev{\textbf{Exhaustive Stage~3 $k$-sweep on the Stage~2-pruned ($r{=}5\%$, VAW) category matrix.} Top-$1$ of the $\binom{18}{k}$ subsets by Spearman $\rho$ against the $18$-category ranking on the pruned matrix (Stage~3 convention), also evaluated against the category means over all Stage~0-cleaned benchmarks (end-to-end).}}
\label{tab:ksweep_exhaustive}
\setlength{\tabcolsep}{4pt}
\begin{tabular}{c c c r l}
\toprule
 & \multicolumn{2}{c}{$\rho$} & & \\
\cmidrule(lr){2-3}
$k$ & Stage~3 & End-to-end & Items & Selected categories \\
\midrule
$3$ & 0.976 & 0.919 & 7{,}391 & \parbox[t]{0.62\linewidth}{\raggedright Academic~Knowledge, Chart~Understanding, Visual~Reasoning} \\\addlinespace[2pt]
$5$ & 0.988 & 0.959 & 9{,}377 & \parbox[t]{0.62\linewidth}{\raggedright Academic~Knowledge, Chart~Understanding, Hallucination~Detection, Visual~Reasoning, Web/UI~Understanding} \\\addlinespace[2pt]
$8$ & 0.996 & 0.962 & 1{,}136 & \parbox[t]{0.62\linewidth}{\raggedright Academic~Knowledge, Chart~Understanding, Commonsense, Document~Understanding, Math~Reasoning, Object~Recognition, Science~Knowledge, Spatial~Reasoning} \\\addlinespace[2pt]
$10$ & 0.995 & 0.958 & 9{,}882 & \parbox[t]{0.62\linewidth}{\raggedright Academic~Knowledge, Comprehensive, Hallucination~Detection, Math~Reasoning, Object~Grounding, Object~Recognition, Science~Knowledge, Text~Recognition, Visual~Reasoning, Web/UI~Understanding} \\
\bottomrule
\end{tabular}
\end{table}

\begin{table}[!htbp]
\centering
\small
\caption{\rev{\textbf{Composition of the released suite.} The $k{=}5$ categories of \cref{tab:ksweep_exhaustive} and the items each keeps after Stage~2 at $r{=}5\%$.}}
\label{tab:release_composition}
\setlength{\tabcolsep}{6pt}
\begin{revblock}
\begin{adjustbox}{max width=\linewidth}
\begin{tabular}{l r}
\toprule
Category & Items kept \\
\midrule
Academic Knowledge & 100 \\
Chart Understanding & 258 \\
Hallucination Detection & 100 \\
Visual Reasoning & 7{,}033 \\
Web/UI Understanding & 1{,}886 \\
\midrule
Total & 9{,}377 \\
\bottomrule
\end{tabular}
\end{adjustbox}
\end{revblock}
\end{table}

\paragraph{\rev{Cross-family agreement.}}
\rev{The item-level agreement between model families on the released suite, the Spearman correlation between two families' mean scores over its $9{,}377$ items for the families with at least four models, is \rev{between $0.115$ and $0.463$ over the 6 family pairs}, so the families differ in which items they find hard.}

\FloatBarrier
\section{\rev{Theoretical Analysis}}
\label{app:theory}

\begin{revblock}
\Cref{app:theory_notation} fixes the notation and the standing assumptions, \cref{app:scaling_law} states the results for the VAW score of \cref{eq:vaw} on a single item pool, in the ranking-equivalent form $\widehat v_i+\alpha_M\delta_i^{\mathrm{VL}}$, and its paragraph on cells carries them into the cells of Stage~2; \cref{app:proofs} proves them, and \cref{app:spectral_intuition} gives the matrix reading of the low-rank observation behind Stage~3. In words, the plug-in variance of an item is estimated with noise of order $M^{-1/2}$ (\cref{thm:concentration}); a tilt held fixed as the panel grows converges to a different selection whenever it changes the population top-$K$ (\cref{thm:vaw-fixed}); a decaying tilt such as $\alpha_M$ approaches the population variance selector at an explicit rate when the boundary-atom mass of A4 vanishes (\cref{thm:vaw-asymptotic}), and recovers it exactly with high probability under the boundary-gap condition A3 (\cref{prop:vaw-exact}); and for a fixed pool and fixed cells each statement holds inside every cell (\cref{app:cells_theory}).
\end{revblock}

\subsection{\rev{Notation and standing assumptions}}
\label{app:theory_notation}

\begin{revblock}
$M$ is the panel size and $N$ the size of the item pool. For item $i$, $x_{1,i},\ldots,x_{M,i}$ are the per-model correctness indicators, $\widehat p_i = \tfrac{1}{M}\sum_{m=1}^{M} x_{m,i}$ is the empirical correctness rate, $p_i=\mathbb{E}[x_{m,i}]$ is the correctness probability, $v_i = p_i(1-p_i)$ is the population Bernoulli variance, and $\widehat v_i = \widehat p_i(1-\widehat p_i)$ is the plug-in variance estimator; for binary scoring the variance score of \cref{eq:vardisc} equals $\sqrt{M}\,\widehat v_i$. $\delta_i^{\mathrm{VL}}\in[0,D_\delta]$ is the vision-dependence score, $\alpha$ is the weight of \cref{eq:vaw}, and $\alpha_M$ the corresponding tilt on $\widehat v_i$, related by $\alpha_M=\alpha/\sqrt M$ (\cref{eq:vaw-normalized}); the pipeline uses $\alpha=\alpha_0\sqrt{14}\approx 0.187$ with $\alpha_0=0.05$, that is $\alpha_M=\alpha_0\sqrt{14/M}$. $T_K(\cdot)$ returns the indices of the $K$ largest scores (ties broken by a fixed order that does not depend on the responses), $S_V^* = T_K(v_i)$ is the population variance top-$K$ set, \rev{$S_{\rm VAW}(M;a) = T_K(\widehat v_i + a\,\delta_i^{\mathrm{VL}})$ is the top-$K$ set on one pool for a tilt $a\ge 0$ on $\widehat v_i$ (the pipeline uses $a=\alpha_M$)}, $K/N$ is the effective retention after the $100$-item floor and $\triangle$ the symmetric difference.
\end{revblock}

\paragraph{Assumptions used by the theorems.}
\textit{(A1) i.i.d.\ Bernoulli within item.} For each item $i$ the per-model correctness indicators $x_{1,i},\ldots,x_{M,i}$ are i.i.d.\ Bernoulli($p_i$).
\rev{\textit{Scores in $[0,1]$.} Stage~2 uses item scores scaled to $[0,1]$. For scores that are i.i.d.\ in $[0,1]$ within an item (A1 with the Bernoulli law replaced by a law on $[0,1]$), let $v_i$ be the population variance of the item score and $\widehat v_i=\frac{1}{M}\sum_m(x_{m,i}-\bar x_i)^2$ its plug-in estimator, which equals $\widehat p_i(1-\widehat p_i)$ for binary scores. Changing one model's score changes $\widehat v_i$ by at most $3/M$, and $\mathbb{E}[\widehat v_i]=(1-\frac{1}{M})v_i$ with $v_i\le 1/4$, so McDiarmid's inequality gives $\Pr(|\widehat v_i-v_i|\ge\epsilon+\frac{1}{4M})\le 2\exp(-2M\epsilon^2/9)$. The results below are stated under (A1); with this bound in place of \cref{thm:concentration} they hold with $\sqrt{\log(2N/\eta)/(2M)}$ replaced by $3\sqrt{\log(2N/\eta)/(2M)}+\frac{1}{4M}$, so the generic concentration and the downstream selection rates are unchanged, while the $\Theta(1/M)$ refinement at $p_i=1/2$ is specific to binary scores. When items have different numbers $M_i$ of valid records, $M_i$ replaces $M$ item by item, the ranked score is $\widehat v_i+\alpha\,\delta_i^{\mathrm{VL}}/\sqrt{M_i}$, the uniform bounds hold with $M=\min_i M_i$, and recovery requires $\min_i M_i$ to grow.}
\textit{(A2) Bounded vision-dependence score.} $\delta_i^{\mathrm{VL}}\in[0,D_\delta]$ for some constant $D_\delta<\infty$; for the cosine distance used here $D_\delta=2$.
\textit{(A3) Boundary-gap condition (used by \cref{thm:vaw-fixed,prop:vaw-exact}).} The population top-$K$ boundary gap $\Delta_V := \min_{i\in S_V^*} v_i - \max_{j\notin S_V^*} v_j$ is strictly positive.
\textit{(A4) Margin condition with boundary-atom mass (used by \cref{thm:vaw-asymptotic}).}
There exist a boundary threshold $\tau_K \in [0,1]$ separating the population top-$K$ from the rest, a margin constant $L_V<\infty$, and a (possibly $N$-dependent) boundary-atom mass $\gamma_N \ge 0$ such that
\[
  \frac{1}{N}\#\{i : |v_i - \tau_K| \le h\} \;\le\; \gamma_N + 2 L_V\, h
  \qquad \text{for every } h > 0,
\]
where $\gamma_N$ absorbs any mass at the boundary. This is the standard nonparametric margin condition~\citep{tsybakov2004} adapted to a finite item pool with explicit tie accounting.

\subsection{\rev{Behaviour of VAW as the number of models grows}}
\label{app:scaling_law}

\paragraph{Bridging \cref{eq:vaw} to the theorems below.}
The released VAW score in \cref{eq:vaw} is $q_i^{\mathrm{VAW}} = \widehat\sigma_i^2 + \alpha\,\delta_i^{\mathrm{VL}}$ with $\widehat\sigma_i^2 = \tfrac{1}{\sqrt M}\sum_m(x_{m,i}-\bar x_i)^2$, which for binary scoring equals $\sqrt M\,\widehat v_i$ where $\widehat v_i = \widehat p_i(1-\widehat p_i)$. For a common $M$ across the compared items, multiplying by the positive constant $1/\sqrt M$ does not change the top-$K$ selector \rev{or the per-cell maximiser of Stage~2} and yields the ranking-equivalent score
\begin{equation}
\frac{q_i^{\mathrm{VAW}}}{\sqrt M}
\;=\;
\widehat v_i + \alpha_M\,\delta_i^{\mathrm{VL}},
\qquad
\alpha_M \;=\; \frac{\alpha}{\sqrt M}.
\label{eq:vaw-normalized}
\end{equation}
A fixed weight $\alpha$ in \cref{eq:vaw} therefore induces an annealed tilt $\alpha_M = O(M^{-1/2})$ on the standard plug-in variance $\widehat v_i$. \rev{The results below are stated for the tilted top-$K$ selector on one pool of $N$ items, and the paragraph on cells below applies them to the cells.} The theorems below are stated in this normalised parametrisation: \cref{thm:vaw-asymptotic} governs the implementation schedule realised by \cref{eq:vaw} via $\alpha_M = \alpha/\sqrt M$, while \cref{thm:vaw-fixed} concerns the (different) regime in which a fixed \rev{tilt $a$} is held constant against $\widehat v_i$ rather than against $\widehat\sigma_i^2$.

\begin{theorem}[Variance-estimator concentration]\label{thm:concentration}
Let $x_{1,i},\ldots,x_{M,i} \stackrel{\mathrm{iid}}{\sim} \mathrm{Bernoulli}(p_i)$ with $v_i = p_i(1-p_i)$.
The plug-in variance estimator $\widehat v_i = \widehat p_i (1-\widehat p_i)$ satisfies
\[
  \mathbb{E}[\widehat v_i] = \bigl(1 - \tfrac{1}{M}\bigr) v_i,
  \qquad
  \mathrm{Var}(\widehat v_i) \le \tfrac{v_i}{M} \le \tfrac{1}{4M},
\]
and for every $\epsilon > 0$, $\Pr(|\widehat v_i - v_i| \ge \epsilon) \le 2\exp(-2 M \epsilon^2)$.
At $p_i = 1/2$ the rate sharpens to $\mathrm{sd}(\widehat v_i) = \Theta(1/M)$.
\end{theorem}

\begin{theorem}[Fixed normalised tilt is inconsistent when it changes the population top-$K$]\label{thm:vaw-fixed}
Assume A1, A2. Either of the following regimes is admissible:
(i) the item count $N$ is fixed as $M\to\infty$;
(ii) $N=N_M$ grows but the perturbed boundary gap $\Delta_{\rm Add,N}(a)$, the population top-$K$ gap of $v_i+a\delta_i^{\mathrm{VL}}$ (written $\Delta_{\rm Add}(a)$ in the proof), satisfies
$\Delta_{\rm Add,N}(a) \gg \sqrt{\log N / M}$
so that uniform concentration drives the empirical perturbed top-$K$ set to its population counterpart.
For any fixed $a\!>\!0$ independent of $M$, the additive VAW score $q_i^{\rm Add}(M;a)\!=\!\widehat v_i + a\delta_i^{\mathrm{VL}}$ converges in probability to the perturbed population score $v_i + a\delta_i^{\mathrm{VL}}$.
If the perturbed population top-$K$ set $S_{\rm Add}^*(a):=T_K(v_i+a\delta_i^{\mathrm{VL}})$ differs from the population variance top-$K$ set $S_V^*\!=\!T_K(v_i)$ and both have positive boundary gaps, then
\[
\Pr\bigl(S_{\rm VAW}(M;a)=S_V^*\bigr)\;\to\;0
\quad\text{as }M\to\infty.
\]
\textit{Remark on \cref{eq:vaw}.} The \rev{fixed-tilt} impossibility above is stated for the \rev{normalised} additive score $\widehat v_i+\rev{a}\,\delta_i^{\mathrm{VL}}$. The released score \cref{eq:vaw} uses $\widehat\sigma_i^2 = \sqrt{M}\,\widehat v_i$ in place of $\widehat v_i$, so a fixed weight $\alpha$ in \cref{eq:vaw} is ranking-equivalent to the annealed schedule $\alpha_M=\alpha/\sqrt M$ on $\widehat v_i$ (cf.\ \cref{eq:vaw-normalized}); this case is governed by \cref{thm:vaw-asymptotic}, not by \cref{thm:vaw-fixed}.
\end{theorem}

\begin{proposition}[Fixed-pool exact recovery under the boundary-gap condition]\label{prop:vaw-exact}
Assume A1, A2, A3 with positive top-$K$ variance gap $\Delta_V > 0$, and a non-negative annealing schedule $\alpha_M\ge 0$.
Then for every \rev{$\eta\in(0,1)$}, with probability at least \rev{$1-\eta$},
\[
\max_i |\widehat v_i - v_i| \;\le\; \rev{\sqrt{\frac{\log(2N/\eta)}{2M}}}.
\]
Consequently, whenever the panel size $M$ and schedule $\alpha_M$ satisfy
\[
\rev{D_\delta \alpha_M + \sqrt{\frac{\log(2N/\eta)}{2M}} \;<\; \Delta_V/2,}
\]
the VAW selector recovers $S_V^*$ exactly with probability at least \rev{$1-\eta$}:
\rev{$\Pr(S_{\rm VAW}(M;\alpha_M) = S_V^*) \ge 1-\eta$}.
\end{proposition}

\begin{theorem}[Annealed tilt approaches the population variance selector --- density-margin rate]\label{thm:vaw-asymptotic}
Assume A1, A2, A4 with constant $L_V$ and boundary-atom mass $\gamma_N$, and a non-negative annealing schedule $\alpha_M\ge 0$ with $\alpha_M\to 0$.
For any confidence parameter $\eta\in(0,1)$, the VAW selector with annealed schedule $\alpha_M\!\to\!0$ satisfies, with probability at least $1\!-\!\eta$,
\[
\frac{|S_{\rm VAW}(M;\alpha_M)\,\triangle\,S_V^*|}{K}
\;=\;O\!\left(\frac{N}{K}\Bigl[\gamma_N + \alpha_M + \sqrt{\tfrac{\log(N/\eta)}{M}}\Bigr]\right),
\]
where the $\gamma_N$ term carries any boundary-atom mass that A4 cannot continuously bound.
For a fixed retention $r = K/N > 0$, $N/K = 1/r$ is a constant and the rate simplifies to $O\bigl(\gamma_N + \alpha_M + \sqrt{\log(N/\eta)/M}\bigr)$; the schedule realised by \cref{eq:vaw} via the $1/\sqrt M$ variance scaling is $\alpha_M=\alpha/\sqrt M$ on $\widehat v_i$ (cf.\ \cref{eq:vaw-normalized}), so the $\alpha_M$ and $\sqrt{\log N/M}$ terms scale as $O(M^{-1/2})$ up to log factors and the symmetric-difference fraction vanishes whenever the displayed right-hand side tends to zero.
\end{theorem}

\paragraph{What these theorems do not claim.}
\begin{revblock}
\Cref{thm:vaw-asymptotic} shows that the annealed tilt approaches the population variance selector at large $M$ when all terms of its bound vanish; for a fixed pool under the boundary-gap condition (A3), \cref{prop:vaw-exact} gives exact recovery. At the panel sizes used here ($M\le 30$ with pools of $90$ items or more in our panel, and up to $265$ models with pools of $218$ items or more on OpenVLM), the union-bound radius over all items used in \cref{prop:vaw-exact} at confidence $0.95$ exceeds $1/8$, half the largest possible gap between two variances, so the bounds describe the large-$M$ regime and give no finite-sample guarantee at these panel sizes. The analysis does not show that variance selection is the strongest selector at large $M$, which would need a minimax lower bound, nor that the tilt improves on variance at small $M$, which would need a finite-sample risk bound; the properties above bound the effect of the tilt in the limit under these conditions and do not show that it helps. Whether it helps is an empirical question, answered by the scan of \cref{app:finite_m_v2}. The results also differ in kind from the IRT-based selection of tinyBenchmarks~\citep{polo2024tinybenchmarks} and metabench~\citep{kipnis2025metabench}, which fit a parametric response model and select items with it. Our statements assume no ability model and bound the estimation error of a model-free item statistic. Four limits apply. The pool is chosen with model responses: Stage~0 removes items by their responses, Stage~1 picks benchmarks by response correlations and adds supplements in descending variance of the same scores (\cref{app:multisource}), so A1 is assumed for the fixed pool rather than inherited from the pipeline. A1 also treats the models as independent draws, while models of one family share architecture and data, so the effective panel size is smaller than $M$; extending the bounds to dependent families would need a dependence model and a matching concentration bound, which we do not provide. The margin constant $L_V$ of A4 is a population quantity that we do not estimate. The bounds hold within one pool, and transfer across pools requires the distribution of $(v_i,\delta_i^{\mathrm{VL}})$ to be stable.
\end{revblock}

\paragraph{\rev{Inside the cells.}}\label{app:cells_theory}
\rev{Stage~2 keeps one item per cell (\cref{sec:stage2}). A cell is itself an item pool, so for a fixed pool and fixed cells the results above apply inside every cell with $K=1$, and one union bound over all $N$ items covers every cell at once. The population choice of a cell is its variance maximiser, which the annealed tilt recovers as $M$ grows when the cell's best two items are separated (\cref{prop:vaw-exact} with $K=1$, which needs the deviation radius below half the smallest such gap over all cells). As for the whole pool, the statements treat the pool and its cells as fixed; the pool itself is chosen with model responses in Stages~0 and~1, which is the first of the limits listed above.}

\paragraph{Scope of the design.}\label{app:vaw_design}
The in-cell score $q^{\mathrm{VAW}}_i = \widehat{\sigma}^2_i + \alpha\,\delta_i^{\mathrm{VL}}$ of \cref{eq:vaw} is constrained in four ways, the first three by construction and the fourth under the assumptions of \cref{app:theory_notation}. First, what it adds to \rev{the variance score} \emph{uses no response information}, since the cells and $\delta_i^{\mathrm{VL}}$ come from frozen vision-text embeddings (Qwen3-VL-Embedding-8B), not from model responses; the pool they act on is fixed by Stages~0 and~1, which do use responses. Second, it is \emph{VLM-targeted}, since $\delta_i^{\mathrm{VL}}$ measures how far the image moves an item's representation, which a text-only criterion cannot see. Third, it is \emph{annealed}, since the ranked score $q^{\mathrm{VAW}}_i/\sqrt{M}$ equals the plug-in variance plus $(\alpha/\sqrt{M})\,\delta_i^{\mathrm{VL}}$, so the tilt on the plug-in variance shrinks as $M^{-1/2}$, the worst-case rate at which that variance concentrates (\cref{thm:concentration}). Fourth, it \emph{\rev{recovers variance-only selection}}. Setting $\alpha{=}0$ gives \rev{variance-only selection in the same cells (VAW$-\delta$)}, and for a fixed pool and fixed cells \cref{prop:vaw-exact} holds within each cell with $K=1$, so under a boundary-gap condition (A3) on the cells the VAW choice recovers \rev{the choice by the items' true variances} as $M$ grows. \Cref{app:finite_m_v2} reports its measured effect at each panel size.

\FloatBarrier
\subsection{\rev{Formal proofs}}
\label{app:proofs}

\rev{This subsection proves \cref{thm:concentration,thm:vaw-fixed,thm:vaw-asymptotic} and \cref{prop:vaw-exact}, with the notation and assumptions A1--A4 of \cref{app:theory_notation}; the spectral low-rank structure (\cref{fig:low_rank_spectrum}) is an empirical observation, discussed in \cref{app:spectral_intuition}.}

\subsubsection{\rev{Proof of \cref{thm:concentration} (variance-estimator concentration)}}

\rev{The proof computes the mean from binomial moments and transfers Hoeffding concentration from $\widehat p_i$ to $\widehat v_i$ through the $1$-Lipschitz map $x\mapsto x(1-x)$; the sharper rate at $p_i=1/2$ follows because the quadratic expansion of that map is exact.}

Let $S_M = \sum_{m=1}^M x_{m,i} = M\widehat p_i$, so $S_M\sim\mathrm{Binomial}(M,p_i)$.
\paragraph{Mean.} $\mathbb{E}[\widehat v_i] = \mathbb{E}[\widehat p_i] - \mathbb{E}[\widehat p_i^2]$. Since $\mathbb{E}[\widehat p_i]=p_i$ and $\mathrm{Var}(\widehat p_i) = p_i(1-p_i)/M = v_i/M$, we have $\mathbb{E}[\widehat p_i^2] = p_i^2 + v_i/M$. Therefore
\[
  \mathbb{E}[\widehat v_i] = p_i - p_i^2 - v_i/M = v_i(1 - 1/M).
\]
\paragraph{Variance.} Write $\widehat v_i = g(\widehat p_i)$ for $g(x)=x(1-x)$. The function $g$ is $1$-Lipschitz on $[0,1]$ (since $|g'(x)|=|1-2x|\le 1$), so for any pair of i.i.d.\ copies $X,Y$ of $\widehat p_i$ we have $(g(X)-g(Y))^2 \le (X-Y)^2$ deterministically; taking expectations and using the identity $\mathrm{Var}(g(\widehat p_i)) = \tfrac{1}{2}\mathbb{E}[(g(X)-g(Y))^2]$ yields the finite-sample bound
\[
  \mathrm{Var}(\widehat v_i) \;\le\; \mathrm{Var}(\widehat p_i) \;=\; v_i/M \;\le\; 1/(4M).
\]
\paragraph{Concentration.} Apply Hoeffding's inequality to $\widehat p_i$: $\Pr(|\widehat p_i - p_i|\ge \epsilon)\le 2\exp(-2M\epsilon^2)$. The 1-Lipschitz property of $g$ gives the deterministic bound $|\widehat v_i - v_i| = |g(\widehat p_i) - g(p_i)| \le |\widehat p_i - p_i|$. Chaining the two bounds:
\[
  \Pr\!\bigl(|\widehat v_i - v_i|\ge\epsilon\bigr)
  \;\le\; \Pr\!\bigl(|\widehat p_i - p_i|\ge\epsilon\bigr)
  \;\le\; 2\exp(-2M\epsilon^2),
\]
which is exactly the bound stated in the theorem.
\paragraph{Refined rate at $p_i = 1/2$.} At $p_i = 1/2$, $g'(p_i)=0$, and because $g$ is quadratic the second-order Taylor expansion is \emph{exact} with no remainder:
\[
  \widehat v_i - v_i \;=\; g(\widehat p_i) - g(p_i) \;=\; -(\widehat p_i - 1/2)^2.
\]
For $S_M\sim\mathrm{Binomial}(M,1/2)$, $\mathrm{Var}((\widehat p_i - 1/2)^2)=(M-1)/(8M^3)=\Theta(1/M^2)$, so $\mathrm{sd}(\widehat v_i) = \Theta(1/M)$, sharper than the generic $\Theta(M^{-1/2})$ rate.
\hfill$\square$

\subsubsection{\rev{Proof of \cref{thm:vaw-fixed} (fixed normalised tilt is inconsistent when it changes the population top-$K$)}}

\rev{The strategy is uniform convergence of the perturbed empirical score to its population counterpart $v_i+a\delta_i^{\mathrm{VL}}$: the perturbed population top-$K$ set has its own positive boundary gap, so the empirical selector locks onto that perturbed set, which differs from $S_V^*$ by assumption.}

Fix $a>0$ independent of $M$. Define $u_i^{(M)} := \widehat v_i + a\delta_i^{\mathrm{VL}}$ and the population perturbed score $u_i := v_i + a\delta_i^{\mathrm{VL}}$. By \cref{thm:concentration}, $\widehat v_i \xrightarrow{p} v_i$ at rate $O(M^{-1/2})$; since $\delta_i^{\mathrm{VL}}$ is fixed (label-free, not estimated from the panel), $u_i^{(M)} \xrightarrow{p} u_i$ uniformly in $i\in\{1,\ldots,N\}$ by union bound: for any $\epsilon>0$,
\[
  \Pr\bigl(\max_i |u_i^{(M)} - u_i|\ge\epsilon\bigr) \le 2N\exp(-2M\epsilon^2)\to 0 \text{ as }M\to\infty.
\]
For the growing-$N$ regime $N=N_M$, set $\epsilon_M = \Delta_{\rm Add,N_M}(a)/2$ so that
\[
  \Pr\bigl(\max_i |u_i^{(M)} - u_i|\ge \Delta_{\rm Add,N_M}(a)/2\bigr) \le 2N_M\exp\!\bigl(-M\Delta_{\rm Add,N_M}(a)^2/2\bigr) \to 0,
\]
which holds under the gap rate $\Delta_{\rm Add,N}(a) \gg \sqrt{\log N/M}$ stated in the theorem.
Let $S_{\rm Add}^*(a) := T_K(u_i)$. Suppose $S_{\rm Add}^*(a)\neq S_V^* := T_K(v_i)$, and let
\[
  \Delta_{\rm Add}(a) := \min_{i\in S_{\rm Add}^*(a)} u_i - \max_{j\notin S_{\rm Add}^*(a)} u_j > 0
\]
denote the population top-$K$ boundary gap of the perturbed score (positive by assumption). On the event $\{\max_i|u_i^{(M)}-u_i|<\Delta_{\rm Add}(a)/2\}$, the empirical top-$K$ set $T_K(u_i^{(M)}) = S_{\rm VAW}(M;a)$ equals $S_{\rm Add}^*(a)$ exactly (the gap is wider than the maximum perturbation, so no item flips across the boundary). The probability of this event tends to $1$ as $M\to\infty$, so
\[
  \Pr\bigl(S_{\rm VAW}(M;a) = S_{\rm Add}^*(a)\bigr) \to 1.
\]
Since $S_{\rm Add}^*(a)\neq S_V^*$, $\Pr(S_{\rm VAW}(M;a) = S_V^*)\to 0$.
\hfill$\square$

\subsubsection{\rev{Proof of \cref{thm:vaw-asymptotic} (annealed tilt approaches the population variance selector)}}

\rev{We bound the uniform deviation of the annealed score by an estimation-noise term plus the vanishing tilt, show that every misclassified item lies in a near-boundary band twice as wide as that deviation, and let the margin condition A4 convert the band width into an item count.}

Let $\alpha_M\to 0$ as $M\to\infty$ and fix $\eta\in(0,1)$. Write $u_i^{(M)} := \widehat v_i + \alpha_M\delta_i^{\mathrm{VL}}$ and decompose
\[
  u_i^{(M)} - v_i = (\widehat v_i - v_i) + \alpha_M\delta_i^{\mathrm{VL}}.
\]
The first term is bounded uniformly in $i$ by Hoeffding + union bound: with probability at least $1 - \eta$,
\[
  \max_i |\widehat v_i - v_i| \le \sqrt{\frac{\log(2N/\eta)}{2M}} = O\!\left(\sqrt{\frac{\log(N/\eta)}{M}}\right).
\]
The second term is bounded deterministically: $|\alpha_M\delta_i^{\mathrm{VL}}|\le \alpha_M D_\delta = O(\alpha_M)$. Therefore $\max_i|u_i^{(M)} - v_i| = O(\alpha_M + \sqrt{\log(N/\eta) / M})$ with probability at least $1-\eta$.

Let $\tau_K$ denote the population boundary threshold separating top-$K$ from the rest under A4, and write $\xi_M = \max_\ell |u_\ell^{(M)} - v_\ell|$. We show by a pairwise gap argument that $S_{\rm VAW}(M;\alpha_M)\,\triangle\,S_V^*$ is contained in the near-boundary set $\{i:|v_i-\tau_K|\le 2\xi_M\}$.
For any $i \in S_V^*\!\setminus\!S_{\rm VAW}(M;\alpha_M)$ and $j \in S_{\rm VAW}(M;\alpha_M)\!\setminus\!S_V^*$, the empirical top-$K$ rule gives $u_j^{(M)} \ge u_i^{(M)}$, hence
$
  v_i \le u_i^{(M)} + \xi_M \le u_j^{(M)} + \xi_M \le v_j + 2\xi_M.
$
Since $v_i \ge \tau_K$ (top side) and $v_j \le \tau_K$ (bottom side), both lie within $2\xi_M$ of $\tau_K$, so $|v_i - \tau_K|\le 2\xi_M$ and $|v_j - \tau_K|\le 2\xi_M$. Applying A4 with $h = 2\xi_M$, whose $\gamma_N$ term absorbs boundary ties (including the boundary item itself in finite pools), the number of items that can be misclassified is at most
\begin{align*}
  N \cdot \Bigl[ \gamma_N + 2 L_V \cdot 2\max_i|u_i^{(M)}-v_i| \Bigr]
  &= N \cdot \Bigl[ \gamma_N + 4 L_V \max_i|u_i^{(M)}-v_i| \Bigr]\\
  &= O\!\left(N\Bigl[\gamma_N + \alpha_M + \sqrt{\tfrac{\log(N/\eta)}{M}}\Bigr]\right).
\end{align*}
This bound is density-based with explicit tie accounting: we do not assume the boundary-gap condition (A3), and we do not condition on an event in which exactly zero items cross the boundary; the $\gamma_N$ term \rev{handles} both the \rev{$K$-th} order-statistic atom and any other boundary ties.
Dividing by $K$ gives the symmetric-difference bound
\[
  \frac{|S_{\rm VAW}(M;\alpha_M)\,\triangle\,S_V^*|}{K} = O\!\left(\frac{N}{K}\Bigl[\gamma_N + \alpha_M + \sqrt{\tfrac{\log(N/\eta)}{M}}\Bigr]\right).
\]
For the implementation schedule $\alpha_M = \alpha/\sqrt M$ realised by \cref{eq:vaw} (cf.\ \cref{eq:vaw-normalized}), $\alpha_M = O(M^{-1/2})$ matches the estimator-noise term up to constants and log factors, so the two $M$-dependent terms vanish at rate $O(M^{-1/2})$ up to logarithmic factors; the $\gamma_N$ term must vanish separately under the assumed asymptotic regime (e.g., $\gamma_N \to 0$ under A4 in the large-$N$ limit, or $\gamma_N \le c/N$ for a constant $c$ independent of $N$, which vanishes as $N\to\infty$, if the only boundary atom is the order-statistic itself).
\hfill$\square$

\subsubsection{\rev{Proof of \cref{prop:vaw-exact} (fixed-pool exact recovery)}}

\rev{Hoeffding's inequality and a union bound give a uniform deviation radius $\zeta_M$ for the tilted score; once $\zeta_M$ is below half the population boundary gap $\Delta_V$, no item can cross the top-$K$ boundary, which forces exact recovery.}

By Hoeffding's inequality applied to each $\widehat p_i$ followed by the $1$-Lipschitz bound $|\widehat v_i - v_i| \le |\widehat p_i - p_i|$ from \cref{thm:concentration}, $\Pr(|\widehat v_i - v_i|\ge\epsilon)\le 2\exp(-2M\epsilon^2)$ for each $i$. By a union bound over $N$ items,
\[
  \Pr\!\Bigl(\max_i |\widehat v_i - v_i| \ge \epsilon\Bigr) \;\le\; 2N\exp(-2M\epsilon^2).
\]
\begin{revblock}
Setting $2N\exp(-2M\epsilon^2) = \eta$ yields $\epsilon = \sqrt{\log(2N/\eta)/(2M)}$. Combining with the deterministic bound $|\alpha_M\delta_i^{\mathrm{VL}}| \le D_\delta\alpha_M$, the maximum perturbation between the empirical score $\widehat v_i + \alpha_M\delta_i^{\mathrm{VL}}$ and the population variance $v_i$ is, with probability at least $1-\eta$,
\[
  \max_i |(\widehat v_i + \alpha_M\delta_i^{\mathrm{VL}}) - v_i| \;\le\; D_\delta\alpha_M + \sqrt{\frac{\log(2N/\eta)}{2M}}.
\]
Define $\zeta_M = D_\delta\alpha_M + \sqrt{\log(2N/\eta)/(2M)}$. With probability at least $1-\eta$, $\max_i |(\widehat v_i + \alpha_M\delta_i^{\mathrm{VL}}) - v_i| \le \zeta_M$. If $\zeta_M < \Delta_V/2$, then for every $i \in S_V^*$ and $j \notin S_V^*$,
\[
  (\widehat v_i + \alpha_M\delta_i^{\mathrm{VL}}) - (\widehat v_j + \alpha_M\delta_j^{\mathrm{VL}})
  \;\ge\; (v_i - v_j) - 2\zeta_M
  \;\ge\; \Delta_V - 2\zeta_M
  \;>\; 0,
\]
so the empirical top-$K$ set equals $S_V^*$ exactly.
\end{revblock}
\hfill$\square$

\subsection{\rev{Spectral intuition for the low-rank structure of VLM score matrices}}
\label{app:spectral_intuition}

\rev{The low-rank structure enters the pipeline as an empirical observation. The leading singular values of the category matrix carry most of its energy (\cref{fig:low_rank_spectrum}), which motivates the small number of categories in Stage~3 (\cref{sec:stage3}). }

\noindent\textit{Intuition.}
\rev{Eckart--Young is a matrix-level statement: it bounds the squared Frobenius residual of the rank-$k$ SVD approximation $\mathbf{S}_k^{\mathrm{svd}}$ of the $z$-normalised model-by-category matrix $\mathbf{S}$, $\|\mathbf{S} - \mathbf{S}_k^{\mathrm{svd}}\|_F^2 = V - V_k = V R_k$ (with $V$ the total and $V_k$ the top-$k$ squared Frobenius energy, and $R_k$ the relative residual energy), not the projection residual of an arbitrary single vector. Under a unit-Frobenius normalisation $V=1$ (the $z$-normalised matrix of \cref{fig:low_rank_spectrum} has $V=\rev{540}$ instead), the matrix-level translation into an entrywise bound on the mean score vector follows from a Cauchy--Schwarz argument: $\|\mathbf{s}_{\mathrm{full}} - \mathbf{s}_k^{\mathrm{svd}}\|_\infty \le \sqrt{R_k/B_{\rm cat}}$ for the column-mean score vectors of $\mathbf{S}$ and $\mathbf{S}_k^{\mathrm{svd}}$ over its $B_{\rm cat}$ columns; without the $V=1$ normalisation the bound reads $\sqrt{V R_k / B_{\rm cat}}$.}
\begin{revblock}
Before item pruning $95\%$ of the energy needs \rev{4} components (participation-ratio effective rank \rev{$(\sum_j \lambda_j)^2/\sum_j \lambda_j^2$ over the singular values $\lambda_j$,} \rev{4.53}), \rev{within} the number of categories Stage~3 keeps; after Stage~2 keeps $5\%$ of the items it needs \rev{7} (effective rank \rev{7.73}; top-$5$ \rev{$92.2\%$}), more than Stage~3 keeps, because sampling noise and the variance criterion, which keeps the items that separate models, spread energy over more components. Stage~3 only has to reproduce the ranking by the mean score, which is one linear functional of the matrix, so a number of categories below the effective rank can preserve it; the exhaustive $k$-sweep measures how far (\cref{tab:ksweep_exhaustive}).
\end{revblock}

\FloatBarrier
\section{\rev{Finite-$M$ Scan of the Vision Tilt inside Cells}}
\label{app:finite_m_v2}\label{app:finite_m}

\begin{figure}[!t]
\centering
  \begin{subfigure}{0.49\linewidth}
    \centering
    \includegraphics[width=\linewidth]{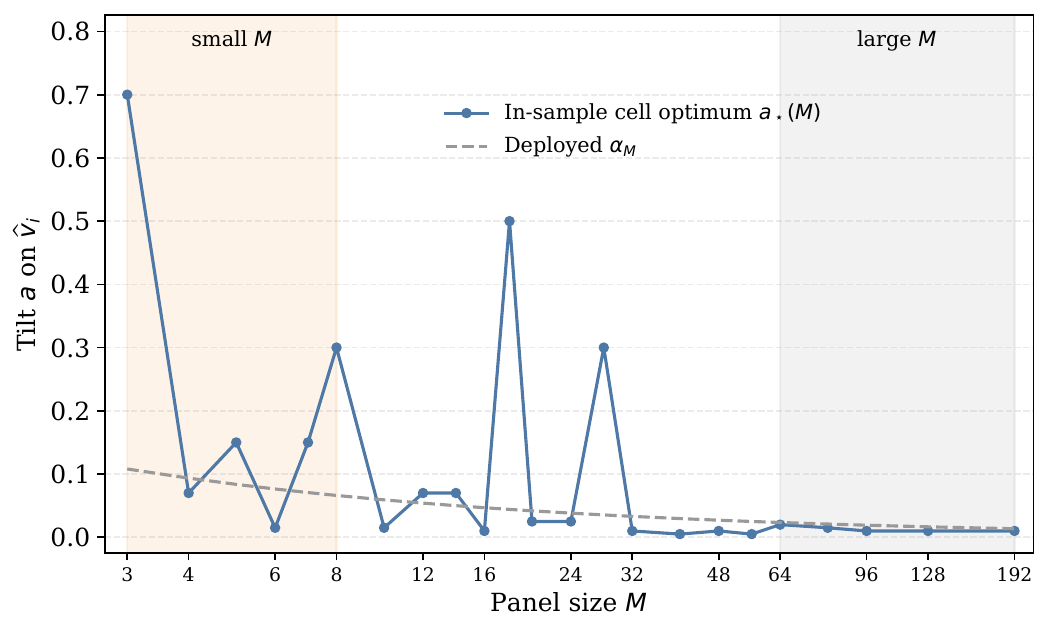}
    \subcaption{\rev{In-sample best tilt $a^\star(M)$ inside cells and the deployed $\alpha_M$.}}\label{fig:finite_m_scaling-a}
  \end{subfigure}\hfill
  \begin{subfigure}{0.49\linewidth}
    \centering
    \includegraphics[width=\linewidth]{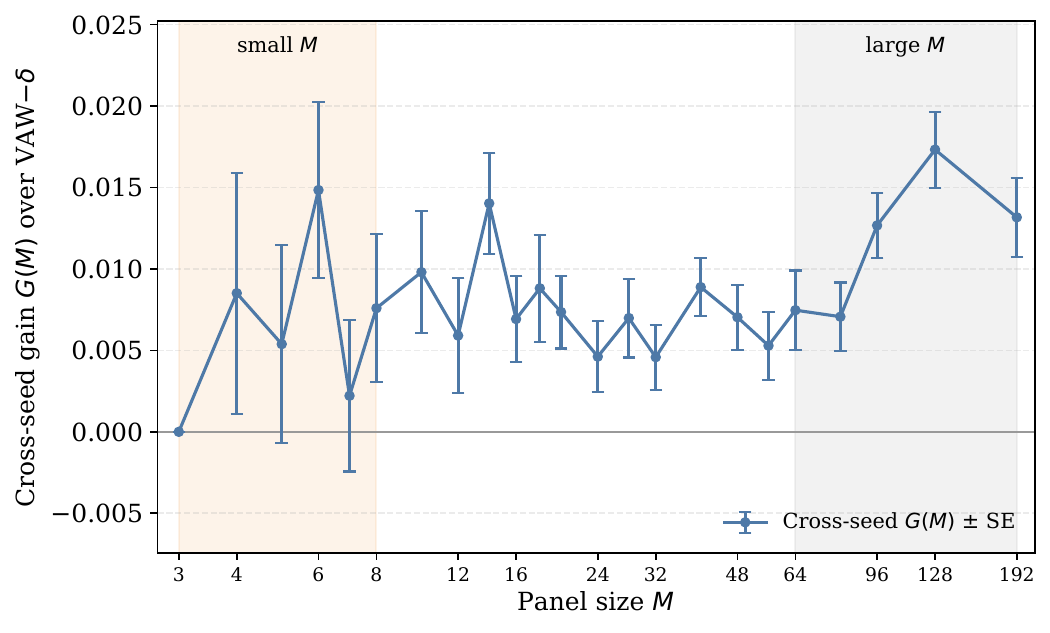}
    \subcaption{\rev{Cross-seed gain $G(M)$ over VAW$-\delta$ in the same cells, $\pm$ one SE.}}\label{fig:finite_m_scaling-b}
  \end{subfigure}
\caption{\rev{\textbf{Inside cells the best tilt tends to decrease with the panel size, and cross-seed tilt selection improves ranking fidelity.} In-cell scan on the $10$ OpenVLM benchmarks with aligned $\delta^{\mathrm{VL}}$ at $r{=}2\%$: $23$ panel sizes $M\in[3,192]$, $30$ model subsamples per $M$, $5$ item splits $\times$ $2$ folds, and one cell per selected item. The tilt is chosen on two split seeds and its gain over VAW$-\delta$ ($a{=}0$ in the same cells) is measured on the other three, with seeded random tie-breaking. Shaded: small ($M\le 8$) and large ($M\ge 64$) panels. The best tilt falls with $M$ (Spearman $-0.664$), and the cross-seed gain is positive at every $M\ge 4$ and exceeds two standard errors at every $M\ge 14$.}}
\label{fig:finite_m_scaling}
\end{figure}

\begin{revblock}
This section asks whether the tilt $\alpha_M\delta_i^{\mathrm{VL}}$ improves ranking fidelity inside the cells, and how the best tilt depends on the panel size. Without cells the in-sample optimum of the normalised tilt is at most $0.02$ at every $M$, with a gain over Variance of at most $+0.007$ under seeded random tie-breaking. \Cref{fig:finite_m_scaling} shows the scan inside cells.

The scan uses the $10$ OpenVLM benchmarks whose items align with our embeddings (AI2D, DocVQA, InfoVQA, MMT-Bench, MM-Vet, MathVista, OCRBench, RealWorldQA, SEEDBench, TextVQA) and the models that cover at least $95\%$ of a benchmark; at the largest panel sizes only the benchmarks with enough such models remain (down to seven). For each of $23$ panel sizes $M$ from $3$ to $192$ it draws $30$ model subsamples. Five seeds split each benchmark's items into halves, and for each seed and fold the training half selects $K=\max(\lfloor 0.02\,n_{\rm train}\rfloor,1)$ items ($r{=}2\%$), with $n_{\rm train}$ the number of items in the training half, while all items are evaluated; $\rho$ is the Spearman correlation over the $M$ models between the kept-item mean and the all-item mean. The cells are drawn by $k$-means from the training-half items' projected vision-text embeddings, fixed across $M$, subsamples and \rev{$a$}, and for each of $19$ \rev{tilts $a$} from $0$ to $2$ and the pipeline's $\alpha_M$, each cell contributes its item with the largest $\widehat v_i+\rev{a}\,\delta_i^{\mathrm{VL}}$. The baseline is \rev{VAW$-\delta$} in the same cells (\rev{$a=0$}).

The in-sample $a^\star(M)$ maximises the mean gain over the baseline. Because a maximum over a grid is optimistic, the reported gain is a cross-seed gain: the tilt is chosen on two split seeds, and the gain $G(M)$ and its standard error are measured on the other three. We report the trend of $a^\star(M)$ over the $23$ panel sizes and, at each size, $G(M)$ with $t=G(M)/\mathrm{SE}$. Ties follow the seeded random order.

The best tilt shrinks as the panel grows: the Spearman correlation between $M$ and $a^\star(M)$ is $-0.664$, with $a^\star$ between $0.005$ and $0.02$ for every $M\ge 32$, of the same order as the deployed $\alpha_M$ ($0.014$--$0.023$ for $M\ge 64$). The tilt chosen on the selection seeds gives a positive cross-seed gain at every $M$ from $4$ to $192$. From $M{=}14$ on the gain exceeds two standard errors at every size ($G=0.005$--$0.017$), and at every $M\ge 64$ it is $G=0.007$--$0.017$ with $t=3.1$--$7.4$; among $M\le 8$ it exceeds two standard errors only at $M{=}6$ ($G=0.015$, $t=2.8$). These results support smaller tilts on larger panels, in the direction of the annealing of \cref{app:scaling_law}, and inside the cells the cross-seed-selected tilt adds a consistent gain on moderate and large panels.
\end{revblock}

\FloatBarrier
\section{External Validation on OpenVLM}
\label{app:external_validation}

\subsection{OpenVLM data}
\label{app:openvlm_data}

OpenVLM publishes per-item predictions of up to $265$ models on the $14$ benchmarks that overlap with our catalogue: AI2D, ChartQA, DocVQA, InfoVQA, MathVista, MME, MMMU, MMT-Bench, MM-Vet, OCRBench, RealWorldQA, ScienceQA, SEEDBench and TextVQA. Per-item correctness uses each benchmark's own metric, with rule-based matching as an offline stand-in for the GPT grading of MM-Vet. All predictions are produced with the image, so the blind-solvability filter of Stage~0, which needs an image-free run, cannot be applied to these records; the all-correct filter, Stage~2 and Stage~3 can. The full pipeline, including the released suite, is validated end to end only on our own panel.

\subsection{Stage~2 on OpenVLM: selector comparison}
\label{app:openvlm_stage2}

Every selector keeps $\mathrm{round}(rN)$ items of each benchmark, and $\rho$ compares the ranking on the kept items with the full-benchmark ranking on the same models, so this comparison is same-model.
With hundreds of models IRT 2PL leads at $r{\le}2\%$ ($0.922$ and $0.938$) and AnchorPoints from $r{=}5\%$ to $20\%$ (\cref{tab:openvlm_stage2}), as expected for response-pattern fits scored on the models they were fitted to. On the $\rev{8}$ benchmarks with a closed answer space, each scored by $\rev{200}$ or more models, DISCO's selection score reaches \rev{0.910} against \rev{0.907} for Variance at $r{=}5\%$.

\begin{table}[!htbp]
\centering
\small
\caption{\rev{\textbf{With up to $263$ models per benchmark, IRT~2PL leads at $r{\le}2\%$ and AnchorPoints at $r{=}5$--$20\%$.} Mean Spearman $\rho$ over the $14$ OpenVLM benchmarks between the full-benchmark ranking and the ranking on the kept items; selection and scoring use the same OpenVLM models (same-model), each benchmark selected on its own; $200$--$263$ models score ten benchmarks, and ChartQA, DocVQA, InfoVQA and TextVQA have $33$, $61$, $53$ and $126$. Best in bold, second best underlined.}}
\label{tab:openvlm_stage2}
\setlength{\tabcolsep}{5pt}
\begin{revblock}
\begin{adjustbox}{max width=\linewidth}
\begin{tabular}{lcccccc}
\toprule
Selector & $1\%$ & $2\%$ & $5\%$ & $10\%$ & $20\%$ & $50\%$ \\
\midrule
Variance & 0.818 & 0.889 & 0.932 & 0.953 & \underline{0.973} & \textbf{0.993} \\
DatBench $r_{pb}$ & 0.805 & 0.847 & 0.903 & 0.929 & 0.959 & 0.986 \\
AnchorPoints & \underline{0.869} & \underline{0.931} & \textbf{0.955} & \textbf{0.966} & \textbf{0.974} & \underline{0.992} \\
IRT 2PL & \textbf{0.922} & \textbf{0.938} & \underline{0.952} & \underline{0.962} & 0.971 & 0.988 \\
\bottomrule
\end{tabular}
\end{adjustbox}
\end{revblock}
\end{table}

\subsection{Stage~3 on OpenVLM: benchmark selection}
\label{app:openvlm_stage3}

The reference ranking is the mean of per-benchmark $z$-scores over each model's available benchmarks, without imputation, on the $231$ models that cover at least $7$ of the $14$ benchmarks. Exhaustive selection at $k{=}5$ reaches $\rho{=}0.988$ (\cref{tab:openvlm_stage3}) with AI2D, MMMU, OCRBench, RealWorldQA and ScienceQA, and greedy selection is within $0.002$ of exhaustive at every $k$. The selection is re-fitted on OpenVLM, so this checks the Stage~3 mechanism on a larger and more diverse panel; whether a frozen selection holds for newer models is tested by the shelf-life analysis of \cref{app:robustness}.

\begin{table}[!htbp]
\centering
\small
\caption{\rev{\textbf{Five of the $14$ OpenVLM benchmarks recover the full ranking at $\rho{=}0.988$.} $\rho$ of exhaustive Stage~3 selection against the $14$-benchmark ranking on the $231$ models that cover at least $7$ benchmarks, with $95\%$ CIs from $10{,}000$ paired bootstrap resamples stratified over $95$ model families inferred from model names ($53$ of them singletons); the CI is computed on the selected subset and omits selection uncertainty. The last column fills missing scores with the benchmark mean over all $265$ models and re-selects the subset.}}
\label{tab:openvlm_stage3}
\setlength{\tabcolsep}{8pt}
\rev{\begin{tabular}{rccc}
\toprule
$k$ & $\rho$ & $95\%$ CI & $\rho$, mean imputation \\
\midrule
\rev{3} & \rev{0.978} & \rev{$[0.973,\,0.982]$} & \rev{0.951} \\
\rev{5} & \rev{0.988} & \rev{$[0.984,\,0.990]$} & \rev{0.978} \\
\rev{8} & \rev{0.994} & \rev{$[0.993,\,0.995]$} & \rev{0.990} \\
\bottomrule
\end{tabular}}
\end{table}

\subsection{Stage~3 on OpenVLM: low rank across release eras}
\label{app:openvlm_eras}

We split the $265$ OpenVLM models into three release eras and recompute the spectrum of each era's $z$-normalised matrix over the $14$ benchmarks (\cref{tab:openvlm_eras}). In every era the top five components hold $86$--$89\%$ of the energy and the participation-ratio effective rank is $7.2$--$8.1$, about half the number of benchmarks; this concentration is what allows a small $k$ in Stage~3.

\begin{table}[!htbp]
\centering
\small
\caption{\rev{\textbf{The top five components hold $86$--$89\%$ of the OpenVLM spectral energy in every release era.} Spectrum of each era's $z$-normalised model-by-benchmark matrix; the last era has $13$ benchmarks because one has no scores, and each model's era is inferred from its name. This concentration is what lets a few benchmarks stand in for all of them.}}
\label{tab:openvlm_eras}
\setlength{\tabcolsep}{4pt}
\rev{\begin{tabular}{lccccc}
\toprule
Release era & Models & Benchmarks & Components for $95\%$ & Participation ratio & Top-$5$ energy (\%) \\
\midrule
Before Jun 2024 & \rev{73} & \rev{14} & \rev{8} & \rev{7.2} & \rev{88.8} \\
Jun--Dec 2024 & \rev{114} & \rev{14} & \rev{9} & \rev{7.7} & \rev{88.1} \\
From Jan 2025 & \rev{78} & \rev{13} & \rev{8} & \rev{8.1} & \rev{86.0} \\
\bottomrule
\end{tabular}}
\end{table}

\FloatBarrier


\section{Additional Analyses}
\label{app:extended_discussion}

\subsection{Shelf life of a frozen benchmark selection}
\label{app:robustness}

\begin{revblock}
We resolve Hugging Face Hub release dates for $203$ of the $265$ OpenVLM models, sort them by date and append the $62$ undated models last. On each cumulative panel the models with at least $7$ of the $14$ benchmarks are kept first, and the per-benchmark $z$-normalisation is fitted on them afterwards. The exhaustive $k{=}5$ selection on the $132$ oldest dated models is $\{$AI2D, MathVista, MME, MMMU, OCRBench$\}$. Frozen, it reaches $\rho{=}0.988$, $0.990$, $0.989$ and $0.989$ at $M{=}132$, $166$, $200$ and $203$ (\cref{tab:shelflife_cumulative}), at or above the $99.9$th percentile of all $2{,}002$ five-benchmark subsets on each panel, at most $0.0004$ below the best subset. On the $55$ newer dated models alone, with the normalisation of the $M{=}203$ panel kept, it reaches $0.989$, at the $88$th percentile. As a separate stress test, appending the $62$ undated models lowers it to $0.971$ ($65$th percentile), where the best subset reaches $0.988$. On the dated panels, a greedy selection frozen the same way reaches $0.986$--$0.988$. Because a random five-benchmark subset already reaches $\rho{\approx}0.97$ on average on this low-rank matrix, we rank the frozen selection among all subsets rather than against a fixed threshold. The comparison is relative to other subsets and does not test the item-level release, and Hub release dates are an imperfect proxy for model generation, so we treat the trace as a stress test, not a causal claim.
\end{revblock}

\begin{table}[!htbp]
\centering
\small
\caption{\rev{\textbf{A five-benchmark selection frozen on the $132$ oldest dated OpenVLM models keeps $\rho\approx 0.99$ as newer dated models join, at or above the $99.9$th percentile of all $2{,}002$ five-benchmark subsets; appending the $62$ undated models lowers it to $0.971$.} Cumulative panels in release order; $M$ counts dated models, and the last row appends the $62$ undated models. Scored = models with at least $7$ of the $14$ benchmarks. Gap = best subset $\rho$ minus frozen $\rho$.}}
\label{tab:shelflife_cumulative}
\setlength{\tabcolsep}{5pt}
\begin{revblock}
\begin{adjustbox}{max width=\linewidth}
\begin{tabular}{r r c c c c}
\toprule
$M$ & Scored & Frozen $\rho$ & Random-five mean & Percentile & Gap \\
\midrule
132 & 126 & 0.988 & 0.965 & 100.0 & 0.0000 \\
166 & 150 & 0.990 & 0.970 & 100.0 & 0.0000 \\
200 & 178 & 0.989 & 0.971 & 99.9 & 0.0001 \\
203 & 181 & 0.989 & 0.971 & 99.9 & 0.0003 \\
$203$ dated $+$ $62$ undated & 231 & 0.971 & 0.962 & 64.8 & 0.0168 \\
\bottomrule
\end{tabular}
\end{adjustbox}
\end{revblock}
\end{table}

\subsection{Capability Frontier: \rev{a descriptive stress set}}
\label{app:capability_frontier}

\begin{revblock}
The Capability Frontier collects the items that most current VLMs fail. An item is a candidate when every family's mean score on it is at most $0.5$ (the ten families of \cref{tab:models}) and at least two models are scored on it; \rev{33{,}262} items qualify. The Frontier selects only from the \rev{16{,}692} of them with a panel mean score in $[0.10,0.50]$, which excludes items that almost every model fails, and keeps those with the largest difficulty-weighted variance $\widehat v_i(1-\bar x_i)$. As a heuristic budget, its size applies the largest per-step accuracy gain along the panel's model axes, the Qwen per-generation gain on the full pool (\cref{tab:cf_sizing}), to this in-band pool, so that \rev{16{,}692} $\times$ \rev{0.09902} gives \rev{1{,}653} items.

We use the Frontier as a descriptive stress set for the current panel. On its items the Qwen-VL-Chat$\to$Qwen3-VL-8B generation gap is \rev{0.62}$\times$ its full-pool value, so the Frontier still separates the Qwen generations. Because the Qwen family's own mean score helps define the pool, this ratio partly reflects the selection itself. VQAv2 supplies \rev{33\%} of the items, but model scores weight benchmarks equally, so VQAv2 does not dominate them. We have not tested whether Frontier items become discriminative as newer generations arrive; the shelf-life check of \cref{app:robustness} is benchmark-level on OpenVLM and does not bear on these items. The item list is produced by the same pipeline and panel as the released suite and released with it.
\end{revblock}

\begin{table}[!htbp]
\centering
\small
\caption{\rev{\textbf{The Qwen generation axis gives the largest per-step gain, \rev{0.099}, and sets the Frontier size.} Accuracy gain on the full pool between the endpoint models of each axis, divided by the number of steps. The Frontier keeps \rev{16{,}692}$\times$\rev{0.09902}$\approx$\rev{1{,}653} items.}}
\label{tab:cf_sizing}
\setlength{\tabcolsep}{4pt}
\begin{revblock}
\begin{adjustbox}{max width=\linewidth}
\begin{tabular}{l l c c c}
\toprule
Axis & Endpoints & Gain & Steps & Per step \\
\midrule
Qwen generations & Qwen-VL-Chat $\to$ Qwen3-VL-8B & +0.297 & $3$ & 0.099 \\
InternVL generations & InternVL2-8B $\to$ InternVL3.5-8B & +0.051 & $3$ & 0.017 \\
Qwen3 parameter scale & Qwen3-VL-2B $\to$ Qwen3-VL-8B & +0.088 & $2$ & 0.044 \\
\bottomrule
\end{tabular}
\end{adjustbox}
\end{revblock}
\end{table}

\subsection{\rev{Left-out benchmark prediction across panel sizes}}
\label{app:cross_benchmark}

\begin{revblock}
We evaluate left-out benchmark prediction on four OpenVLM targets: MMT-Bench, ScienceQA, TextVQA, and BLINK. For each target, a Ridge regressor takes the model's scores on the remaining aggregate benchmarks as input ($13$ features, or $14$ for BLINK) and predicts its score on the target benchmark. Depending on the target, $126$--$214$ models have a score on it. We vary the number of training models from $M{=}1$ to $128$ ($64$ for TextVQA) and report MAE on the remaining models over $30$ random splits.

\Cref{tab:app_d1_mae} lists the MAE of each target at each $M$, and \cref{fig:d1_mae_scaling} plots it. The MAE decreases from $0.11$--$0.21$ at $M{=}1$ to $0.04$--$0.10$ at the largest available number of training models. The improvement becomes smaller at larger $M$: between $M{=}64$ and $128$, MMT-Bench remains at $0.039$ MAE, while ScienceQA and BLINK improve by at most $0.003$. TextVQA is available for $126$ models, so $M{=}64$ is its largest tested training size. The predictors are full benchmark scores, not the released suite, so the analysis probes cross-benchmark redundancy rather than the compressed suite.
\end{revblock}

\begin{table}[!htbp]
\centering
\small
\caption{\rev{\textbf{Left-out benchmark error falls with the number of training models and flattens beyond $M{=}64$.} Mean MAE over $30$ random splits of a Ridge regressor that predicts each target benchmark from the other OpenVLM aggregate benchmarks, by the number $M$ of training models. Models = models scored on the target, with a missing feature score replaced by its mean over the training models; TextVQA has no tested training size above $M{=}64$.}}
\label{tab:app_d1_mae}
\setlength{\tabcolsep}{4pt}
\begin{revblock}
\begin{tabular}{lrrrrrrrrr}
\toprule
Target & Models & $M{=}1$ & $2$ & $4$ & $8$ & $16$ & $32$ & $64$ & $128$ \\
\midrule
MMT-Bench & $200$ & $0.114$ & $0.089$ & $0.074$ & $0.061$ & $0.051$ & $0.044$ & $0.039$ & $0.039$ \\
ScienceQA & $214$ & $0.168$ & $0.126$ & $0.117$ & $0.103$ & $0.086$ & $0.071$ & $0.065$ & $0.062$ \\
TextVQA & $126$ & $0.211$ & $0.194$ & $0.169$ & $0.139$ & $0.123$ & $0.111$ & $0.102$ & -- \\
BLINK & $177$ & $0.119$ & $0.092$ & $0.079$ & $0.065$ & $0.058$ & $0.054$ & $0.049$ & $0.047$ \\
\bottomrule
\end{tabular}
\end{revblock}
\end{table}

\subsection{\rev{Checkpoint and hyper-parameter ranking}}
\label{app:ckpt_ranking}

\begin{revblock}
A compressed suite is also used to rank variants of one model. The variant sets are LoRA fine-tunes (rank $16$, two epochs) of Qwen2-VL-2B on ScienceQA: learning-rate grid A (eight rates, final adapter of each), a narrower grid B, eight checkpoints spread over one training run, eight adapters from the converged end of the middle-rate run, and a denser $15$-checkpoint trajectory. Grid A is also run with twelve rates, grid B with twelve rates at three training seeds, and grid A with twelve rates on Qwen2.5-VL-3B-Instruct. Items are selected on the ScienceQA records of $\rev{30}$ panel models, and no variant takes part in selection. Fidelity is the Spearman $\rho$ between the ranking of the variants on the selected items and on the full $4{,}241$-item test set. A variant set supports a conclusion only if its full-test ranking is itself stable, with bootstrap $\rho\ge 0.90$, a conservative ceiling that the converged tail and the dense trajectory fail.

On the \rev{eight} ceiling-passing variant sets at $r{=}5\%$ (\cref{tab:app_ckpt_ranking}), VAW beats Variance by more than one adjacent swap in \rev{six} variant sets and loses in \rev{one}, for a mean gain of \rev{0.082}. The mean gain comes mostly from the twelve-rate grid~B at seeds $43$ and $44$, where Variance falls to \rev{0.592} and \rev{0.808}, while at seed $42$ \rev{VAW is also ahead} (\rev{0.908} against \rev{0.870}); across the three seeds the VAW lead ranges from \rev{+0.04} to \rev{+0.28}. \rev{Because the variants of one training run are not independent draws, we report this comparison descriptively.}
\end{revblock}

\begin{table}[!htbp]
\centering
\small
\caption{\rev{\textbf{VAW raises the checkpoint-ranking fidelity over Variance \rev{in six of the eight ceiling-passing variant sets}.} Spearman $\rho$ (average ranks for ties) between the ranking of LoRA variants on the selected ScienceQA items and on the full $4{,}241$-item test set; items are selected on $\rev{30}$ panel models, which never include the variants. $n$ = variants; ceiling = bootstrap $\rho$ of the full-test ranking, and variant sets below $0.90$ support no conclusion. Without ties, one adjacent swap moves $\rho$ by $0.024$ ($n{=}8$), $0.007$ ($n{=}12$) or $0.004$ ($n{=}15$). Means over the \rev{eight} ceiling-passing variant sets at $r{=}5\%$: Variance \rev{0.857}, VAW \rev{0.939}. \rev{With $920$ pool items, both retentions select the $100$-item floor, so their columns coincide.} The higher of the two selectors is in bold.}}
\label{tab:app_ckpt_ranking}
\footnotesize
\setlength{\tabcolsep}{3pt}
\begin{revblock}
\begin{adjustbox}{max width=\linewidth}
\begin{tabular}{l c c cc cc}
\toprule
 & & & \multicolumn{2}{c}{$r{=}5\%$} & \multicolumn{2}{c}{$r{=}10\%$} \\
\cmidrule(lr){4-5}\cmidrule(lr){6-7}
Variant set & $n$ & Ceiling & Variance & VAW & Variance & VAW \\
\midrule
LR grid A & 8 & 1.00 & 0.874 & \textbf{1.000} & 0.874 & \textbf{1.000} \\
LR grid B & 8 & 0.99 & \textbf{0.976} & 0.868 & \textbf{0.976} & 0.868 \\
Trajectory & 8 & 0.98 & 0.861 & \textbf{0.970} & 0.861 & \textbf{0.970} \\
Converged tail & 8 & 0.62 & -0.252 & \textbf{0.540} & -0.252 & \textbf{0.540} \\
Trajectory, dense & 15 & 0.87 & 0.585 & \textbf{0.882} & 0.585 & \textbf{0.882} \\
LR grid A, 12 rates & 12 & 1.00 & 0.881 & \textbf{0.982} & 0.881 & \textbf{0.982} \\
LR grid B, 12 rates, seed 42 & 12 & 0.99 & 0.870 & \textbf{0.908} & 0.870 & \textbf{0.908} \\
LR grid B, 12 rates, seed 43 & 12 & 0.99 & 0.592 & \textbf{0.868} & 0.592 & \textbf{0.868} \\
LR grid B, 12 rates, seed 44 & 12 & 0.99 & 0.808 & \textbf{0.924} & 0.808 & \textbf{0.924} \\
LR grid A, 12 rates, 3B & 12 & 1.00 & \textbf{0.991} & 0.989 & \textbf{0.991} & 0.989 \\
\bottomrule
\end{tabular}
\end{adjustbox}
\end{revblock}
\end{table}

\subsection{\rev{Evaluation-compression selectors do not transfer to training-data pruning}}
\label{app:d6_training_pruning}

\begin{revblock}
To test whether evaluation-compression selectors carry over to training-data selection, we fine-tune a SigLIP-base vision encoder with a BERT-base text encoder and a multiple-choice scalar head on ScienceQA, with no ScienceQA-specific training before this experiment. For each ratio $r\in\{0.05,0.1,0.2,0.3,0.5,0.7,0.9\}$ and each of three seeds ($42$, $1337$, $2024$), we train on the selected fraction of the $11{,}453$-item training pool (a stratified $90\%$ of the $12{,}726$ training items; the other $10\%$ is the early-stopping set) and report accuracy on the full $4{,}241$-item test split. We compare six selectors: Random, Variance, VAW (the released selector, which keeps one item from each of $K$ coverage cells of the vision-text embeddings), DatBench~$r_{pb}$~\citep{joshi2026datbench}, AnchorPoints~\citep{vivek2024anchor} (\rev{here} $k$-means on SigLIP image embeddings, with no response signal\rev{, unlike the response-based AnchorPoints used for evaluation}) and IRT~2PL~\citep{kipnis2025metabench}. The response-based selectors score the training items on the correctness matrix of the $30$ panel models. Variance, VAW and DatBench select the same items for every seed; AnchorPoints and IRT vary with it.

No selector beats Random by more than the seed spread (\cref{tab:d6_training_pruning}). At $r\le 0.3$ VAW is within $3.1$ points of test accuracy of Random and ahead of it only at $r{=}0.3$, by $0.8$ points against seed standard deviations of $2$--$3$ points; Variance trails Random by $10.5$--$20.2$ points, and the best of DatBench, AnchorPoints and IRT trails it by $1.8$--$9.5$ points. At $r{=}0.5$ and $0.7$ Random leads every selector by at least $1.3$ points, and at $r{=}0.9$ DatBench and IRT edge past it by $1.7$ and $1.1$ points. The result indicates that training selection and evaluation compression are different problems, since the first rewards informative training examples and the second rewards items that separate models. We therefore do not recommend evaluation-side item scores, VAW included, for training-data pruning; in this setting uniform random sampling is the safer baseline.
\end{revblock}

\begin{table}[!htbp]
\centering
\small
\caption{\textbf{No evaluation-compression selector reliably beats random sampling for training-data selection.} ScienceQA test accuracy, mean over $3$ seeds, of a SigLIP+BERT model fine-tuned on the selected fraction $r$ of the training pool. VAW is ahead of Random only at $r{=}0.3$, and DatBench~$r_{pb}$ and IRT 2PL only at $r{=}0.9$, each within the seed spread. Best in bold, second best underlined.}
\label{tab:d6_training_pruning}
\setlength{\tabcolsep}{4pt}
\begin{revblock}
\begin{adjustbox}{max width=\linewidth}
\begin{tabular}{lccccccc}
\toprule
Selector & $0.05$ & $0.1$ & $0.2$ & $0.3$ & $0.5$ & $0.7$ & $0.9$ \\
\midrule
Random & $\mathbf{0.542}$ & $\mathbf{0.619}$ & $\mathbf{0.707}$ & $\underline{0.735}$ & $\mathbf{0.782}$ & $\mathbf{0.805}$ & $0.804$ \\
Variance & $0.437$ & $0.473$ & $0.505$ & $0.565$ & $0.676$ & $0.748$ & $0.796$ \\
VAW & $\underline{0.541}$ & $\underline{0.607}$ & $\underline{0.676}$ & $\mathbf{0.743}$ & $\underline{0.751}$ & $0.773$ & $0.803$ \\
DatBench~$r_{pb}$~\citep{joshi2026datbench} & $0.512$ & $0.549$ & $0.612$ & $0.670$ & $0.723$ & $0.769$ & $\mathbf{0.821}$ \\
AnchorPoints~\citep{vivek2024anchor} & $0.524$ & $0.545$ & $0.572$ & $0.633$ & $0.739$ & $\underline{0.792}$ & $0.782$ \\
IRT 2PL~\citep{kipnis2025metabench} & $0.452$ & $0.477$ & $0.526$ & $0.569$ & $0.704$ & $0.761$ & $\underline{0.815}$ \\
\bottomrule
\end{tabular}
\end{adjustbox}
\end{revblock}
\end{table}


\FloatBarrier
\section{Limitations}
\label{app:limitations}

\begin{revblock}
\paragraph{Panel size and evaluation noise.}
With $10$ to $20$ held-out models per split, the split-to-split variation of $\rho$ is comparable to the gaps between the stronger selectors, so we report means over $20$ splits, with the number of splits each selector wins, and make no claim that depends on a single split.

\paragraph{External validation.}
On OpenVLM we validate Stage~3 at the benchmark level and compare Stage~2 selectors under same-model scoring; the blind-solvability filter cannot be rerun there, because the records carry no image-free predictions. The released suite is validated end to end only on our panel.

\paragraph{Scope of the theory.}
The theoretical results show that the effect of the annealed tilt on the plug-in variance shrinks at the rate of the estimation noise and that, for a fixed pool and fixed cells with a boundary-gap condition (A3), the cell selector recovers the per-cell variance choice as $M$ grows; at the panel sizes used here they give no finite-sample guarantee, and they do not show that the tilt or the cells improve ranking fidelity. Evidence on those questions comes from the experiments of \cref{app:ablation_cells,app:finite_m_v2}.

\paragraph{Size of the vision contribution.}
Embedding cells beat random partitions by about \rev{$0.011$} in $\rho$, and text-only cells do as well as vision-text cells, so the gain comes from grouping similar items and does not require image inputs to build the cells.

\paragraph{Cross-benchmark prediction.}
Scores on the other OpenVLM benchmarks predict a left-out benchmark with an error that falls as the number of training models grows (\cref{fig:d1_mae_scaling}), but the error is still $0.04$--$0.10$ MAE at the largest panels available, and we have not tested the released suite itself as an estimator.

\paragraph{Judges and modality scope.}
Scores on the judge-scored benchmarks inherit the errors of their judge. The panel contains only open-weight models on image--text benchmarks; proprietary API models and video or audio benchmarks are not covered. Stages~1 and~3 consume only the score matrix and are modality-agnostic, and Stage~0 extends to another modality by rerunning inference with that modality removed; Stage~2 would need a dependence score and an embedding for the new modality.
\end{revblock}

\FloatBarrier
\section{Broader Impact}
\label{app:broader_impact}
PRIMEBench lowers the cost of VLM evaluation by replacing repeated full-suite evaluation with a compact, ranking-preserving subset. This can make systematic VLM evaluation more accessible to small-budget research groups, reduce duplicated GPU expenditure across model releases, and lower the environmental cost of large-scale benchmarking. The released suite also provides a common reference point for comparing new models without requiring every group to rerun hundreds of thousands of benchmark items.

The main risk is that aggressive compression may narrow the community's attention towards capabilities that are already well represented in existing benchmarks. A compact suite should therefore not be treated as a substitute for comprehensive evaluation in high-stakes deployment settings or for auditing newly emerging capabilities. To mitigate this, PRIMEBench is released as a hierarchical pipeline rather than only as a fixed \rev{item set}: users can stop at less compressed stages when broader coverage is required, and future versions can refresh the item pool as new model families and new capability axes appear.

\end{appendices}

\end{document}